\documentclass{article} 
\usepackage{iclr2025_conference,times}

\usepackage{amsmath,amsfonts,bm}

\def\eqref#1{equation~\ref{#1}}

\def\1{\bm{1}}

\DeclareMathAlphabet{\mathsfit}{\encodingdefault}{\sfdefault}{m}{sl}
\SetMathAlphabet{\mathsfit}{bold}{\encodingdefault}{\sfdefault}{bx}{n}

\usepackage[utf8]{inputenc} 
\usepackage[T1]{fontenc}    
\usepackage{hyperref}
\usepackage{amssymb}
\usepackage{url}

\usepackage{booktabs}       
\usepackage{amsfonts}       
\usepackage{nicefrac}       
\usepackage{microtype}      
\usepackage{xcolor}         
\usepackage{graphicx}
\usepackage{float}
\usepackage{wrapfig}
\usepackage{caption}
\usepackage{enumitem}
\usepackage{subcaption}

\usepackage{lscape}

\title{(Mis)Fitting: A Survey of Scaling Laws}

\author{Margaret Li*, Sneha Kudugunta*, Luke Zettlemoyer \\
\thanks{Equal contribution}
\texttt{\{margsli,snehark\}@cs.washington.edu} 
}

\iclrfinalcopy 
\begin{document}

\maketitle

\begin{abstract}


Modern foundation models rely heavily on using scaling laws to guide crucial training decisions. Researchers often extrapolate the optimal architecture and hyper parameters settings from smaller training runs by describing the relationship between, loss, or task performance, and scale. All components of this process vary, from the specific equation being fit, to the training setup, to the optimization method. Each of these factors may affect the fitted law, and therefore, the conclusions of a given study. We discuss discrepancies in the conclusions that several prior works reach, on questions such as the optimal token to parameter ratio. We augment this discussion with our own analysis of the critical impact that changes in specific details may effect in a scaling study, and the resulting altered conclusions. Additionally, we survey over 50 papers that study scaling trends: while 45 of these papers quantify these trends using a power law, most under-report crucial details needed to reproduce their findings. To mitigate this, we we propose a checklist for authors to consider while contributing to scaling law research.


\end{abstract}

\section{Introduction}


Training at the scale seen in recent large foundation models \citep{dubey2024llama,openai2023gpt,reid2024gemini} is an expensive and uncertain process. Given the infeasibility of hyperparameter tuning multi-billion parameter models, researchers extrapolate the optimal training setup from smaller training runs.
More precisely, scaling laws \citep{kaplan2020scaling} are used to study many different aspects of model scaling. Scaling laws can guide targets for increasing dataset size and model size in pursuit of desired accuracy and latency for a specific deployment scenario, study architectural improvements, determine optimal hyperparameters and assist in model debugging. 
\vspace{-2mm}
\begin{wrapfigure}{r}{0.5\columnwidth}
 \centering
 \captionof{table}{We provide a summary of the papers surveyed, highlighting the reproducibility challenges endemic to scaling law papers. }
\begin{tabular}{lc}
\toprule
 $\#$ Papers... & \\ \midrule
Surveyed & 51 \\ 
With a quantified scaling law & 45 \\
With a description of training setup & 40 \\
With a definition of equation variables & 36 \\
With a description of evaluation & 29 \\
With a description of curve fitting & 28 \\
With analysis code & 19 \\
With metric scores or checkpoints provided & 17 \\\bottomrule
\end{tabular}%
\label{tab:summary} 
  \vspace{-3mm}
\end{wrapfigure}

 Scaling laws are often characterized as power laws between the loss and size of the model and dataset, and are seen in several variations (Section \ref{sec:related-work}). These laws are found empirically by training models across a few orders of magnitude in model size and dataset size, and fitting the loss of these models to a proposed scaling law. Each component of this process varies in the reported literature, from the specific equation being fit, to the training setup, and the optimization method, as well as specific details for selecting checkpoints, counting parameters and the objective loss optimized during fitting. 

Changes to this setup can lead to significant changes to the results, and therefore completely different conclusion to the study. For example, \citet{kaplan2020scaling} studied the optimal allocation of compute budget, and found that dataset size should be scaled more slowly than model size ($D \propto N^{0.74}$, $D$ is dataset size, $N$ is model size). Later, \citet{hoffmann2022training} contradicted this finding, showing that model size and dataset size should be scaled roughly equally for optimal scaling. They highlight the differences in setup which lead to them showing that large models should be trained for significantly more tokens: particularly, they point to using later checkpoints, training larger models, a different learning rate schedule and changing the number of training tokens used across runs. Multiple followup works have focused on either reproducing or explaining the differences between these two papers \citep{besiroglu2024chinchilla,porian2024resolving, pearce2024reconcilingkaplanchinchillascaling}. The authors find it challenging to reproduce results of previous papers - we refer the reader to Section \ref{sec:related-work} for a further discussion on these replication efforts.

Motivated by this, we survey over 50 papers on scaling laws across a variety of modalities, tasks and architectures, and find that essential details needed to reproduce scaling law studies are often underreported. We broadly categorize these details as follows:

\paragraph{Section \ref{sec:power-law-form}: What \textit{form} are we fitting?} Researchers may choose any number of power law forms relating any set of variables, to which they fit the data extracted from training runs. Even seemingly minor differences in form, may imply critical changes in assumptions -- for example, about certain interactions between variables which are excluded, the definitions of these variables or error terms which are deemed significant enough to include.
\vspace{-5mm}
\paragraph{Section \ref{sec:model_training}: How do we \textit{train models}?} In order to fit a scaling law, one needs to train a range of models spanning orders of magnitude in parameter count and/or dataset size. Each model requires a multitude of hyperparameter and parameter choices, such as the specific model/dataset sizes to use, the architecture shape, batch size or learning rate schedule.
\vspace{-5mm}
\paragraph{Section \ref{sec:data}: How do we \textit{extract data} after training?} Once these models are trained, downstream metrics like perplexity must be obtained from the intermediate or final checkpoints. This data may be also scaled, interpolated or bootstrapped to create more datapoints to fit the power law parameters.

\vspace{-5mm}
\paragraph{Section \ref{sec:opt}: How are we \textit{optimizing} the fit?} Finally, the variable must be fit with an objective and optimization method, which may in turn have their own initialization and hyperparameters to choose.

To aid scaling laws researchers in reporting details necessary to reproduce their work, we propose a checklist  (Figure~\ref{sec:checklist} - an expanded version may be found in Appendix \ref{sec:app_checklist} ). Based on this checklist, we summarize these details for all 51 papers in tabular form in Appendix \ref{app:full-details}. 
We find that important details are frequently underreported, significantly impacting reproducibility, especially in cases where there is no code - only 19 of 42 papers surveyed have analysis code/code snippets available. Additionally, 23 (a little over half) of surveyed papers do not describe the optimization process, and 15 do not describe how training FLOPs or number of parameters are counted, which has been found to significantly change results \citep{porian2024resolving}. 
In addition, we fit our own power laws to further demonstrate how these choices critically impact the final scaling law~(Section \S\ref{sec:own-repl}). 


\begin{figure}
\fbox{\begin{minipage}{38em}

\subsubsection*{Scaling Law Reproducilibility Checklist}

\small

\begin{minipage}[t]{0.18\textwidth}
\raggedright
\paragraph{Scaling Law Hypothesis (\S\ref{sec:power-law-form})}

\begin{itemize}[leftmargin=*]
    \item Form
    \item Variables (Input)
    \item Parameters
    \item Derivation and Motivation
    \item Assumptions
\end{itemize}

\end{minipage}
\begin{minipage}[t]{0.28\textwidth}
\raggedright

\paragraph{Training Setup (\S\ref{sec:model_training})}
\begin{itemize}[leftmargin=*]
    \item \# of models
    \item Model size range
    \item Dataset source \& size
    \item Parameter/ FLOP Count Calculation
    \item Hyperparameter Choice
    \item Other Settings
    \item Code Open Sourcing
\end{itemize}

\end{minipage}
\begin{minipage}[t]{0.23\textwidth}
\raggedright

\paragraph{Data Collection(\S\ref{sec:data})}
\begin{itemize}[leftmargin=*]
    \item Checkpoint Open Sourcing
    \item \# Checkpoints per Power Law
    \item Evaluation Dataset \& Metric
    \item Metric Modification \& Code 
\end{itemize}

\end{minipage}
\begin{minipage}[t]{0.28\textwidth}
\raggedright

\paragraph{Fitting Algorithm (\S\ref{sec:opt})}
\begin{itemize}[leftmargin=*]
    \item Objective (Loss)
    \item Algorithm
    \item Optimization Hyperparameters
    \item Optimization Initialization
    \item Data Usage Coverage
    \item Validation of Law(s)
\end{itemize}

\end{minipage}


\end{minipage}
}
\caption{We introduce a checklist for researcher to use for scaling laws research. In Appendix \ref{sec:app_checklist}, we include an expanded version of the checklist that may be used as a template. }
\end{figure}\label{sec:checklist}

\section{Papers on Scaling Laws}\label{sec:related-work}


Researchers have proposed scaling laws to study the scaling of deep learning across multiple domains and for several tasks. Studies of the scaling properties of generalization error with training data size and model capacity predate modern deep learning. \citet{banko2001scaling} observed\ a power law scaling of average validation error on a confusion set disambiguation task with increasing dataset size. The authors also claimed that the model size required to fit a given dataset grows log linearly. As for larger scale models, \cite{amodei2016deep} observe a power-law WER improvement on increasing training data for a 38M parameter Deep Speech 2 model. \cite{hestness2017deep} show similar power law relationships across several domains such as machine translation, language modeling, image processing and speech recognition. Moreover, they find that these exponential relationships found hold across model improvements. 

\cite{kaplan2020scaling} push the scale of these studies further, studying power laws for models up to 1.5B parameters trained on 23B tokens to determine the optimal allocation of a fixed compute budget. Later studies \citep{hoffmann2022training,hu2024minicpm} revisit this and find that \cite{kaplan2020scaling} greatly underestimate the amount of data needed to train models optimally, though major procedural differences render it challenging to attribute the source of this discrepancy. Since then, researchers have studied various aspects of scaling up language models. \cite{wei2022emergent} examine the emergence of abilities with scale that are not present in smaller models, while \cite{hernandez2021scaling} study the scaling laws for transfer between distributions in a finetuning setting. \cite{henighan2020scaling} consider possible interactions between different modalities while recently, \cite{aghajanyan2023scaling} study scaling in multimodal foundation models. \cite{tay2022scaling} show that not all architectures scale equally well, highlighting the importance of using scaling studies to guide architecture development. \cite{poli2024mechanistic} scale hybrid architectures like Mamba \citep{gu2023mamba}, showing the efficacy of this new model family. Other researchers also formulate specific scaling laws to study other Transformer based architectures. For example, \cite{clark2022unified} and \cite{frantar2023scaling} introduce new scaling laws to study mixture of expert models \citep{fedus2022switch,shazeer2017outrageously} and sparse models \citep{zhu2017prune} respectively. Researchers have also used scaling laws to study encoder-decoder models for neural machine translation \citep{ghorbani2021scaling,gordon2021data}, and the effect of data quality and language on scaling coefficients \citep{bansal2022data,zhang2022examining}. While language models form the majority of the papers surveyed, we also consider papers that study VLMs \citep{cherti2023reproducible,henighan2020scaling}, vision \citep{alabdulmohsin2022revisiting,zhai2022scaling}, reinforcement learning \citep{hilton2023scaling,jones2021scaling, gao2023scaling} and recommendation systems \citep{ardalani2022understanding}. We further discuss different forms of scaling laws researchers introduce for the specific research questions they wish to answer in Section \ref{sec:power-law-form}. 

A majority of the surveyed papers study Transformer \citep{vaswani2017attention} based models, but a few consider different architectures. For example, \cite{sorscher2022beyond} investigate data pruning laws in ResNets, and some smaller scale studies use MLPs or SVMs \citep{hashimoto2021model}. This overrepresentation is perhaps partially a result of Transformer-based models achieving higher scale than other architectures; a ResNet101 has 44M parameters, while the largest Llama 3 model has 405B.

\paragraph{Replication Efforts} \citet{besiroglu2024chinchilla} seek to reproduce the parameter fitting approach used by \citet{hoffmann2022training}. They are unable to recover the scaling law from \citet{hoffmann2022training}, and demonstrate that the claims of the original paper are inconsistent with descriptions of the setup. They then seek to improve the fit of the scaling law by initializing from the parameters found in \citet{hoffmann2022training} and modifying parts of the power law fitting process.


\citet{porian2024resolving} isolate several decisions as primarily responsible for the discrepancy between the recommendations of \citet{kaplan2020scaling} and \citet{hoffmann2022training}: (1) learning rate scheduler warmup, (2) learning rate decay, (3) inclusion of certain parameters in total parameter count, and (4) specific training hyperparameters. By adjusting these factors, they are able to reproduce the results of \citet{kaplan2020scaling} and \citet{hoffmann2022training}. However, they only use 16 training runs to fit their scaling laws, each designed to match one targeted setting (e.g., replicating \citet{kaplan2020scaling}). 
Instead of using raw loss values, they fit to loss values found by interpolating between checkpoints. Like \citet{pearce2024reconcilingkaplanchinchillascaling}, they apply a log transform and linear regression to fit their law.

\section{What \emph{form} are we fitting?}\label{sec:power-law-form}


A majority of papers we study fit some kind of power law ($f(x)=ax^{-k}$). That is, they specify an equation defining the relationship between multiple factors, such that a proportional change in one results in the proportional change of at least one other. 
They then optimize this power law to find some parameters. 
A few efforts do not seem to fit a power law, but may show a line of best fit, obtained through unspecified methods \citep{rae2021scaling,dettmers2022llm,tay2022scaling,shin2023scaling,schaeffer2023emergent,poli2024mechanistic}.



 
 The specific form may be motivated by researcher intuition, previous empirical results, prior work, code implementation, or data availability. More importantly, the form is often determined by the specific question(s) a paper investigates. For example, one may attempt to predict the performance achieved by scaling up different model architectures, or the optimal ratio for model scaling vs data scaling when increasing training compute \citep{kaplan2020scaling,hoffmann2022training}. Based on this, we loosely classify scaling laws by their form as \textit{performance prediction} and \textit{ratio optimization} approaches. We indicate this classification for all surveyed papers in Appendix \ref{app:full-details}.

\subsection{Ratio Optimization} \label{sec:ratio_opt}
The simplest scaling law forms usually predict the relation between two variables in an optimal setting. For example, approaches 1 and 2 from \citet{hoffmann2022training}
fit to the optimal (i.e., lowest loss) $D$ and $N$ values for a particular compute budget $C$. 
\citet{porian2024resolving}, aiming to resolve these inconsistencies, defines\ $\rho^* = \frac{D^*}{N^*}$ and writes this relationship as: 
\begin{equation}
    N^*(C) = N^*_0 \cdot C^{\alpha} ; D^*(C) = D^*_0 \cdot C^{\alpha}; \rho^*(C) = \rho^*_0 \cdot C^{\alpha}
\end{equation}
They assume $C \approx 6ND$, and thus only need to fit the first equation; the other power laws can be inferred. This simplicity is deceptive in some cases, as collecting $(N^*(C), C)$ pairs may not be trivial. It is possible to fix $C$ and follow a binary search approach to train a multitude of models, then bisect to approximate the performance-optimal $N,D$ pair. However, this quickly grows prohibitively costly. In practice, it is common to interpolate between a set of fixed results to estimate the true $N^*(C)$ {\S\ref{sec:data}}. This adds to the complexity of this approach, and introduces a hidden dependency on the performance evaluation, yet it does not actually predict the performance of the optimal points. 
If only the performance of the optimal-ratio model is of interest, it is possible to fit a second power law $L(N^*(C),  D^*(C)) = a \cdot C^\alpha$. Most papers we survey choose to fit a power law which directly predicts performance.

\subsection{Performance prediction}\label{sec:perf-pred}
\citet{kaplan2020scaling} proposes a power law between Loss $L$, number of model Parameters $N$, and number of Dataset tokens $D$: 
\begin{equation}
L(N, D) = \left[ \left( \frac{N}{N_c}\right)^{\frac{\alpha_N}{\alpha_D}} + \frac{D}{D_c} \right]^{\alpha_D}     
\end{equation}

On the other hand, Approach 3 of \citet{hoffmann2022training} proposes
\begin{equation}
    L(N, D) = E + \frac{A}{N^\alpha} + \frac{B}{D^\beta} 
\end{equation}

In both of the above, all variables other than $L$, $N$, and $D$ are parameters to be found in the power law fitting process. Though these two forms are quite similar, they differ in some assumptions. \citet{kaplan2020scaling} constructs their form on the basis of 3 expected scaling law behaviors, and \citet{hoffmann2022training} explains in their Appendix D that their form is based on risk decomposition. The resulting \citet{kaplan2020scaling} form includes an interaction between $N$ and $D$ in order to satisfy a constraint requiring assymmetry introduced by one of their expected behaviors. The \citet{hoffmann2022training} form, on the other hand, consists of 3 additive sources of error, $E$ representing the irreducible error that would exist even with infinite data and compute budget, as well as two terms representing the error introduced by limited parameters and limited data, respectively.

Power laws for performance prediction can sometimes yield closed form solutions for optimal ratios as well. However, the additional parameters and input variables, introduced by the need to incorporate the performance metric term, add random noise and dimensionality. This increases the difficulty of optimization convergence, so when prediction performance is not the aim, a ratio optimization approach is frequently a better choice.


Many papers directly adopt one of these forms, but some adapt these forms to study relationships with other input variables. \citet{clark2022unified}, for example, study routed Mixture-of-Expert models, and propose a scaling law that relates dense model size (effective parameters) $N$ and number of experts $E$ with a biquadratic interaction ($\log L(N, E) \triangleq a \log N+b \log E+c \log N \log E+d$). \citet{frantar2023scaling} study sparsified models, and propose a scaling law with an additional parameter sparsity $S$, the optimal value of which increases with $N$ ($L(S, N, D)=\left(a_S(1-S)^{b_S}+c_S\right) \cdot\left(\frac{1}{N}\right)^{b_N}+\left(\frac{a_D}{D}\right)^{b_D}+c$). Other papers change the form to model variables in the data setup. \citet{aghajanyan2023scaling} consider interference and synergy between multiple data modalities ($L(N, D_j)=E_j + \frac{A_j}{N^{\alpha_j}} + \frac{B_j}{|D_j|^{\beta_j}}$, $L(N, D_i, D_j) = [\frac{L(N, D_i) + L(N, D_j)}{2}] - C_{i,j} + \frac{A_{i,j}}{N^{\alpha_{i,j}}} + \frac{B_{i,j}}{|D_i|+|D_j|^{\beta_{i,j}}}$), while \cite{goyal2024scaling}, \citet{fernandes2023scaling} and \cite{muennighoff2024scaling} add terms to their scaling law formulations which represent mixing data sources and/or repeated data, using notions such as diminishing utility. A comprehensive list of the power law forms in the surveyed papers may be found in Table \ref{tab:full-powerlaw}.




\section{How do we \emph{train models}?}\label{sec:model_training}


In order to fit a scaling law, one needs to train a range of models across multiple orders of magnitude in model size and/or dataset size. Researchers must first decide the range and distribution of $N$ and $D$ values for their training runs, in order to achieve stable convergence to a solution with high confidence, while limiting the total compute budget of all experiments. Many papers did not specify the number of data points used to fit each scaling law; those that did range from 4 to several hundred, but most used fewer than 50 data points. The specific $N$ and $D$ values also skew the optimization process towards a certain range of $N/D$ ratios, which may be too narrow to include the true optimum. Some approaches, such as using IsoFLOPs \citep{hoffmann2022training}, additionally dictate rules for choosing $N$ and $D$ values. Moreover, using a minimum $N$ or $D$ value may result in outlier values that may need to be dropped \citep{porian2024resolving,shin2023scaling,henighan2020scaling}. We investigate this choice in Section \S\ref{sec:repl-model_training}

The definition of $N$, $D$, or compute cost $C$ can affect the results of a scaling study. For example, if a study studies variation in tokenizers, a definition of training data size based on character count may be more appropriate than one based on token count \citep{tao2024scaling}. The inclusion or exclusion of embedding layer compute and parameters, may also skew the results of a study - a major factor in the different in optimal ratios determined by \cite{kaplan2020scaling} and \cite{hoffmann2022training} has been attributed to not factoring embedding FLOPs into the final compute cost \citep{pearce2024reconcilingkaplanchinchillascaling, porian2024resolving}. Given the increase in extremely long context models (128k-1M) \cite{reid2024gemini}, the commonly used training FLOPs approximation $C = 6 ND$ (see Appendix \ref{app:full-details}) may not hold for such models, given the additional cost proportional to the context length and model dimension - \citet{bi2024deepseek} introduce a new terms non-embedding FLOPs/token to account for this.









Scaling law fit depends on the performance of each individual checkpoint, which is highly dependent on factors such as training data source, architecture and hyperparameter choice. \citet{bansal2022data} and \citet{goyal2024scaling}, for instance, discuss the effect of data quality and composition on power law exponents and constants. Repeating data has also been found to yield different scaling patterns in large language models \citep{muennighoff2024scaling,goyal2024scaling}. 

Researchers have also studied the effect of architecture choice on scaling - \citet{hestness2017deep} find that architectural improvements only shift the irreducible loss, while \citet{poli2024mechanistic} suggest that these improvements may be more significant. The way in which a model is scaled can also affect results. Within the same architecture family, \citet{clark2022unified} show that increasing the number of experts in a routed language model has diminishing returns beyond a point, while \citet{ghorbani2021scaling} find that scaling the encoder and decoder have different effects on model performance. Scaling embedding size can also drastically change scaling trends \citep{tao2024scaling}.

The optimal hyperparameters to train a model changes with scale. Changing batch size, for example, can change model performance \cite{mccandlish2018empirical, kaplan2020scaling}. Optimal learning rate is another hyperparameter shown to change with scale, though techniques such as those proposed in Tensor Programs series of papers \citep{yang2022tensor} can keep this factor constant with simple changes to initialization. More specifically, changing the learning rate schedule from a cosine decay to a constant learning rate with a cooldown (or even changing the learning rate hyperparameters) has been found to greatly affect the results of scaling laws studies \citep{hu2024minicpm,porian2024resolving,hagele2024scaling, hoffmann2022training}.





One common motivation for fitting a scaling laws is extrapolation to higher compute budgets. However, there is no consensus on the orders of magnitude up that one can project a scaling law and still find it accurate, nor on the breadth of compute budgets that should be covered by the data. We find that the range of model size $N$ and dataset size $D$ greatly varies, with the maximum value of $N$ in each paper ranging from 10M parameters to around 7B and that of $D$ being as large as 400B tokens. 
For most papers we survey, the scales are relatively modest: 13 of 51 papers train models beyond 2B parameters; most only train models smaller than 1B parameters.
It has been shown, with some controversy \cite{schaeffer2023emergent}, that scaling to significantly larger scales can result in new abilities that did not appear in smaller models \citep{wei2022emergent}. Forecasting limits to extrapolation and the appearance of new abilities at new scales is an open question.

%



\section{How do we \emph{collect data} from model training?}\label{sec:data}

To evaluate the range of models trained to fit a scaling law, train or validation loss are most commonly used, but some works consider other metrics, such as ELO score \citep{jones2021scaling,neumann2022scaling}, reward model score \citep{gao2023scaling}, or downstream task metrics like accuracy or classification error rate \citep{henighan2020scaling,zhai2022scaling,cherti2023reproducible,goyal2024scaling,gao2023scaling}. This choice is non-trivial - while some papers show that there is a power law relation between the predicted loss found by using validation loss and a different downstream task \citep{dubey2024llama}, it is possible for the results of a study to change completely depending on the metric used. \citet{schaeffer2023emergent}, for example, find that using linear metrics such as Token-edit distance instead of non-linear metrics such as accuracy produces smooth, continuous predictable changes in model performance, contrary to an earlier study by \citet{wei2022emergent}. Moreover, \citet{neumann2022scaling} find that they are unable to use test loss instead of Elo scores  to fit a power law.

While it is most straightforward to evaluate only the final checkpoint on the target metric, some studies may use the median score of the last several checkpoints of each training run \citet{ghorbani2021scaling}, or multiple intermediate checkpoints throughout each training run for various reasons. One common reason is that this is the only computationally feasible way to obtain a fit with sufficient confidence intervals \citep{besiroglu2024chinchilla}. For instance, the ISOFLop approach to finding the optimal $D/N$ ratio in \citet{hoffmann2022training} requires training multiple models for each targeted FLOP budget - this would be computationally prohibitive to do without using intermediate checkpoints. \citet{hoffmann2022training}, in particular, use the last $15\%$ checkpoints. Some papers also report bootstrapping values \citep{ivgi2022scaling}. This detail is often not specified in scaling law papers, with only 29 of 51 papers reporting this information - we point the reader to Appendix \ref{app:full-details} for an overview.

A related technique is performance interpolation. \citet{porian2024resolving} do not aim to exactly match the desired FLOP counts when evaluating model checkpoints mid-training. They instead interpolate between multiple model checkpoints to estimate the performance of a model with the target number of FLOPs. \citet{hoffmann2022training} and \citet{tao2024scaling} also interpolate intermediate checkpoints. \citet{hilton2023scaling}, relatedly, smooth the learning curve before extracting metric scores. 

As discussed in Section \ref{sec:model_training}, training models with too little data or too few parameters can skew the results. To prevent this issue, several works report filtering out data points before fitting their power law. \citet{henighan2020scaling} drop their smallest models, while \citet{hilton2023scaling} and \citet{hoffmann2022training} exclude early checkpoints. \citet{muennighoff2024scaling} remove outlier datapoints that perform badly due to excess parameters or excess epochs. Similarly, \citet{ivgi2022scaling} remove outlier solutions after bootstrapping.

\section{How are we \emph{optimizing} the fit?}\label{sec:opt}
\begin{table}[!h]
\centering
\resizebox{\textwidth}{!}{%
\begin{tabular}{llllll}
\hline
Paper & Curve-fitting Method & Loss Objective & Hyperparameters & Initialization & Are scaling laws  \\
&  &  & Reported? &  &  validated? \\ \hline  \hline
\cite{rosenfeld2019constructive} & Least Squares Regression & Custom error term & N/A & Random & Y \\
\cite{mikamiscaling} & Non-linear Least Squares in log-log space &  & N/A & N/A & Y \\
\cite{schaeffer2023emergent} & NA & NA & NA & NA & NA \\
\cite{sardana2023beyond} & L-BFGS & Huber Loss & Y & Grid Search & N \\
\cite{sorscher2022beyond} & NA & NA & NA & NA & NA \\
\cite{caballero2022broken} & Least Squares Regression & MSLE & N/A & Grid Search, optimize one & Y \\
\cite{besiroglu2024chinchilla} & L-BFGS & Huber Loss & Y & Grid Search & Y \\
\cite{gordon2021data} & Least Squares Regression &  & N/A & N.S. & N \\
\cite{bansal2022data} & NS & NS & N & NS & N \\
\cite{hestness2017deep} & NS & RMSE & N & NS & Y \\
\cite{bi2024deepseek} & NS & NS & N & NS & Y \\
\cite{bahri2021explaining} & NS & NS & N & NS & N \\
\cite{geiping2022much} & Non-linear Least Squares &  & NA & Non-augmented parameters & Y \\
\cite{poli2024mechanistic} & NS & NS & N & NS & N \\
\cite{hu2024minicpm} & scipy curvefit & NS & N & NS & N \\
\cite{hashimoto2021model} & Adagrad & Custom Loss & Y & Xavier & Y \\
\cite{ruan2024observational} & Linear Least Squares & Various & N/A & N/A & Y \\
\cite{anil2023palm} & Polynomial Regression (Quadratic) & N.S. & N & N.S. & Y \\
\cite{pearce2024reconciling} & Polynomial Least Squares & MSE on Log-loss & N/A & N/A & N \\
\cite{cherti2023reproducible} & Linear Least Squares & MSE & N/A & N/A & N \\
\cite{porian2024resolving} & Weighted Linear Regression & weighted SE on Log-loss & N/A & N/A & Y \\
\cite{alabdulmohsin2022revisiting} & Least Squares Regression & MSE & Y & N.S. & Y \\
\cite{gao2024scalingevaluatingsparseautoencoders} & N.S & N.S & N.S & N.S & N.S \\
\cite{muennighoff2024scaling} & L-BFGS & Huber on Log-loss & Y & Grid Search, optimize all & Y \\
\cite{rae2021scaling} & None & None & N/A & N/A & N \\
\cite{shin2023scaling} & NA & NA & NA & NA & NA \\
\cite{hernandez2022scaling} & NS & NS & NS & NS & NS \\
\cite{filipovich2022scaling} & NS & NS & NS & NS & NS \\
\cite{neumann2022scaling} & NS & NS & NS & NS & NS \\
\cite{droppo2021scaling} & NS & NS & NS & NS & NS \\
\cite{henighan2020scaling} & NS & NS & NS & NS & NS \\
\cite{goyal2024scaling} & Grid Search & L2 error & Y & NA & Y \\
\cite{aghajanyan2023scaling} & L-BFGS & Huber on Log-loss & Y & Grid Search, optimize all & Y \\
\cite{kaplan2020scaling} & NS & NS & NS & NS & N \\
\cite{ghorbani2021scaling} & Trust Region Reflective algorithm, Least Squares & Soft-L1 Loss & Y & Fixed & Y \\
\cite{gao2023scaling} & NS & NS & NS & NS & Y \\
\cite{hilton2023scaling} & CMA-ES+Linear Regression & L2 log loss & Y & Fixed & Y \\
\cite{frantar2023scaling} & BFGS & Huber on Log-loss & Y & N Random Trials & Y \\
\cite{prato2021scaling} & NS & NS & NS & NS & NS \\
\cite{covert2024scaling} & Adam & Custom Loss & Y & NS & Y \\
\cite{hernandez2021scaling} & NS & NS & NS & NS & Y \\
\cite{ivgi2022scaling} & Linear Least Squares in Log-Log space & MSE & NA & NS & Y \\
\cite{tay2022scaling} & NA & NA & NA & NA & NA \\
\cite{tao2024scaling} & L-BFGS, Least Squares & Huber on Log-loss & Y & N Random Trials from Grid & Y \\
\cite{jones2021scaling} & L-BFGS & NS & NS & NS & NS \\
\cite{zhai2022scaling} & NS & NS & NS & NS & NS \\
\cite{dettmers2023case} & NA & NA & NA & NA & NA \\
\cite{dubey2024llama} & NS & NS & NS & NS & Y \\
\cite{hoffmann2022training} & L-BFGS & Huber on Log-loss & Y & Grid Search, optimize all & Y \\
\cite{ardalani2022understanding} & NS & NS & NS & NS & NS \\
\cite{clark2022unified} & L-BFGS-B & L2 Loss & Y & Fixed & NS \\ \hline
\end{tabular}%
}
\caption{We provide an overview of which papers provide specific details required to reproduce how they fit their scaling law equation.}
\label{tab:opt-details}
\end{table}

The optimization of a power law requires several design decisions, including optimizer, loss, initialization values, and bootstrapping. We discuss each in this section. Over half of the papers we analyze do not provide any information about their power law fitting process, or provide limited information only and fail to detail crucial aspects. Specifically, many papers fail to describe their choice of optimizer or loss function. In Table \ref{tab:opt-details}, we provide an overview of the optimization details (if specified) for each paper considered.



\paragraph{Optimizer}

Power laws are most commonly fit with a variety of algorithms designed to optimize non-linear functions. One of the most common is the BFGS (Broyden-Fletcher-Goldfarb-Shanno) algorithm, or a variation L-BFGS \citep{liu1989limited}.
Some papers \citep{hashimoto2021model,covert2024scaling} use Adam, Adagrad, or other optimizers common in machine learning, such as AdamW, RMSProp, and SGD. Though effective for LLM training, these are sometimes ill-suited for the purpose of fitting a scaling law, due to various factors limiting their practicality, such as data-hungriness. \citet{goyal2024scaling} forgo the use of an optimizer altogether due to instability of solutions (see initialization) and rely exclusively on grid search to fit their scaling law parameters.

Some scaling law works \citep{rosenfeld2019constructive} opt to use a linear method, such as linear regression, which is generally much simpler. To do this, they typically convert the hypothesized power law to a linear form by taking the log. For example, for a power law $y^b = c \cdot x^a + d$, use the form $\beta \cdot log(y) = \gamma + \alpha \cdot log(x)$ instead. For loss prediction, this results in a form similar to $log(L(N, D)) = \alpha \cdot logN + \beta \cdot logD + E$. This trick is employed even when using an optimizer capable of operating on non-linear functions \citep{hashimoto2021model}. Though this conversion may sometimes work in practice, it is generally not advised because the log transformation also changes the distribution of errors, exaggerating the effects of errors at small values. This mismatch increases the likelihood of a poor fit \citep{goldstein2004problems}. We found this approach to be very common among the papers we study.

\paragraph{Loss}
Various loss functions have been chosen for power law optimization, including variants on MAE (mean absolute error), MSE (mean squared error), and the Huber loss \citep{huber1992robust}, which is identical to MSE for errors less than some value $\delta$ (a hyperparameter), but grows linearly, like MAE, for larger errors, effectively balancing the weighting of small errors with robustness to outliers. Of the papers which specify their loss function, most use a variant of MAE \citep{ghorbani2021scaling}, MSE \citep{goyal2024scaling,hilton2023scaling}, Huber loss \citep{hoffmann2022training,aghajanyan2023scaling,frantar2023scaling,tao2024scaling,muennighoff2024scaling}, or a custom loss \citep{covert2024scaling}. 

\paragraph{Initialization}

Initialization can have a substantial impact on final optimization fit (\S\ref{sec:own-repl}). One approach is to iteratively train with different initializations, selecting the best fit at the termination of the search. This is typically a grid search over choices for each parameter \citep{aghajanyan2023scaling,muennighoff2024scaling}, or a random sample from that grid \citep{frantar2023scaling,tao2024scaling}. Alternatively, the full grid of potential initializations can be evaluated on the loss function without training, and the most optimal $k$ used for initialization and optimization \citep{caballero2022broken}. Finally, if a hypothesis exists, either from prior work or expert knowledge about the function, this hypothesis may be used instead of a search, or to guide the search \citep{besiroglu2024chinchilla}.

\paragraph{Validating the Scaling Law} A majority of the papers surveyed do not report validating the scaling law in any meaningful way. Knowing this is critical to understanding whether the results of the scaling laws study are valid, given the examples given throughout the paper of scaling laws study conclusions changing depending on the process details. \citet{porian2024resolving} and \citet{alabdulmohsin2022revisiting} use confidence intervals and goodness of fit measures to validate their scaling laws. \citet{ghorbani2021scaling} and \citet{bansal2022data} also do this. Otherwise, a majority of the papers that we report as validating their scaling laws mainly extrapolate to models a few orders of magnitude larger and observe the adherence to the scaling law obtained. 






\section{Our Replications and Analyses}\label{sec:own-repl}

Each of the choices discussed above in Sections~\ref{sec:power-law-form}~-~\ref{sec:opt} may have a crucial impact on the result, yet it remains common to critically underspecify the setup for fitting a power law. Scaling law works often fail to open source their model and code, making reproduction infeasible, and likely contributing to contradictory conclusions as discussed in Section \ref{sec:related-work}. Though some efforts have been made  \citep{porian2024resolving,besiroglu2024chinchilla} to reconcile such discrepancies, there is still only sparse understanding of the impact of each of the decisions we discuss.

To investigate the significance of these scaling law optimization decisions, we vary these choices to fit our own scaling laws. We fit both the Chinchilla-scraped data from \citet{besiroglu2024chinchilla}, and data from our own models.

\paragraph{Reconstructed Chinchilla data \citep{besiroglu2024chinchilla}}
This data is extracted from a vector-based figure in the pdf of \citet{hoffmann2022training}, who claim that this includes all models trained for the paper. It consists of 245 datapoints, each corresponding to a final checkpoint collected at the end of model training. There is a potential risk of errors in this recovery process. 


\paragraph{Data from \citet{porian2024resolving}} This data includes training losses for Transformer LMs ranging in size from 5 to 901 million parameters, each trained on 
OpenWebText2 \citep{gao2020pile800gbdatasetdiverse} and RefinedWeb data \citep{penedo2023refinedwebdatasetfalconllm}, across a variety of data and compute budgets, from 5 million to 14 billion tokens (depending on parameter count). Each model is trained with a different peak learning rate and batch size setting, found by fitting a separate set of scaling laws.

\paragraph{Our models} We train a variety of Transformer LMs, ranging in size from 12 million to 1 billion parameters, on varied data and compute budgets and hyperparameter settings. Details about our setup, including hyperparameters, are listed in Appendix~\ref{app:our_models}. We open source all of our models, evaluation results, code, and FLOP calculator at \url{https://github.com/hadasah/scaling_laws}

We fit a multitude of power laws and study the effects of: (1) power law form; 
(2) model learning rate; 
(3) compute budget, model size, and data budget range and coverage; 
(4) definition of $N$ and $C$;
(5) inclusion of mid-training checkpoints;
(6) power law parameter initialization; 
(7) choice of loss and (8) optimizer.

Based on our observations, we also make some more concrete recommendations in Appendix \S\ref{sec:app_recs}, with the caveat that following the recommendations cannot guarantee a good scaling law fit.

\begin{figure}[]
\centering

\begin{subfigure}{\textwidth}
\begin{subfigure}{0.49\textwidth}
    \centering
    \includegraphics[width=\textwidth]{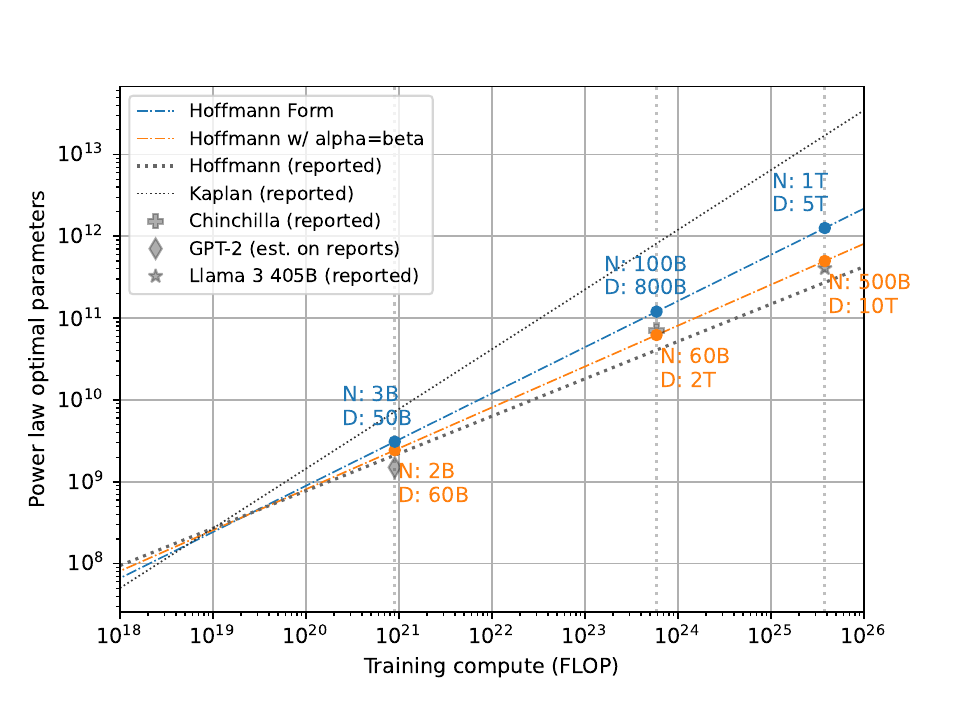}
    \footnotesize{\citet{hoffmann2022training,besiroglu2024chinchilla}}
    \label{fig:analysis_form_epoch}
\end{subfigure}
\hfill
\begin{subfigure}{0.49\textwidth}
    \centering
    \includegraphics[width=\textwidth]{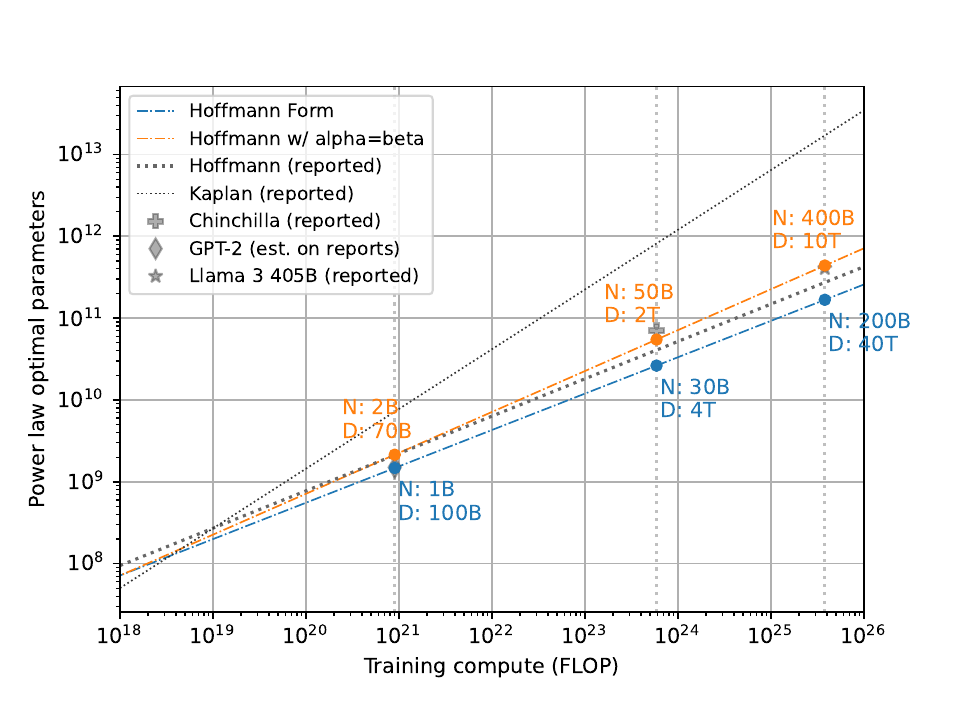}
    \footnotesize{\citet{porian2024resolving}}
\end{subfigure}
\vspace{1em}
\centering

\begin{subfigure}{0.49\textwidth}
    \centering
    \includegraphics[width=\textwidth]{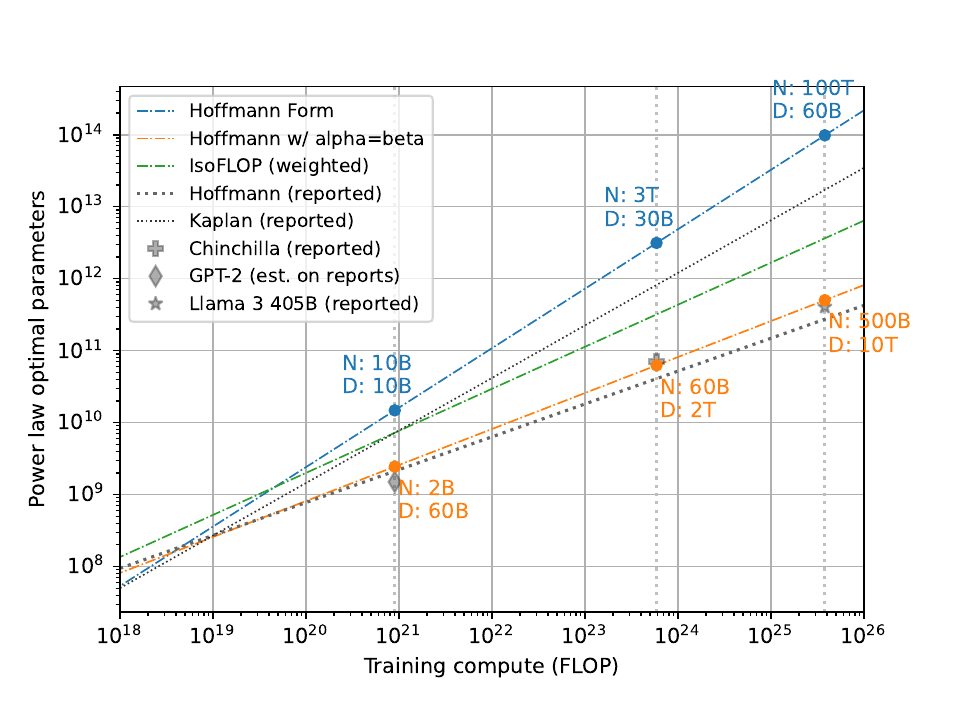}
    \footnotesize{Ours}
    \label{fig:analysis_form_ours}
\end{subfigure}
\vspace{1em}

\caption{\textbf{\S\ref{sec:power-law-form}, \S\ref{sec:repl-power-law-form}} Using data from all 3 datasets, we compare the effects of fitting to the power law form used in Approach 3 of \citet{hoffmann2022training} with the variant used by \citet{muennighoff2024scaling}, which assumes that the exponents $\alpha, \beta$ are equal -- equivalently, that $N^*(C)$ and $D^*(C)$ scale about linearly with each other.  When using only the performance of final checkpoints from both \citet{besiroglu2024chinchilla} and our own experiments, taking this assertion results in a law much closer to the one reported by \citet{hoffmann2022training}. On our own models, we also show results when using the IsoFLOP approach from \citet{hoffmann2022training}. As we are using only results from final model checkpoints, the size of the data input to the IsoFLOP approach in this case is reduced.}
\label{fig:analysis_form}
\end{subfigure}

\end{figure}
\begin{figure}[]
\ContinuedFloat 
\begin{subfigure}{\textwidth}
    \centering
    
    \includegraphics[width=0.7\textwidth]{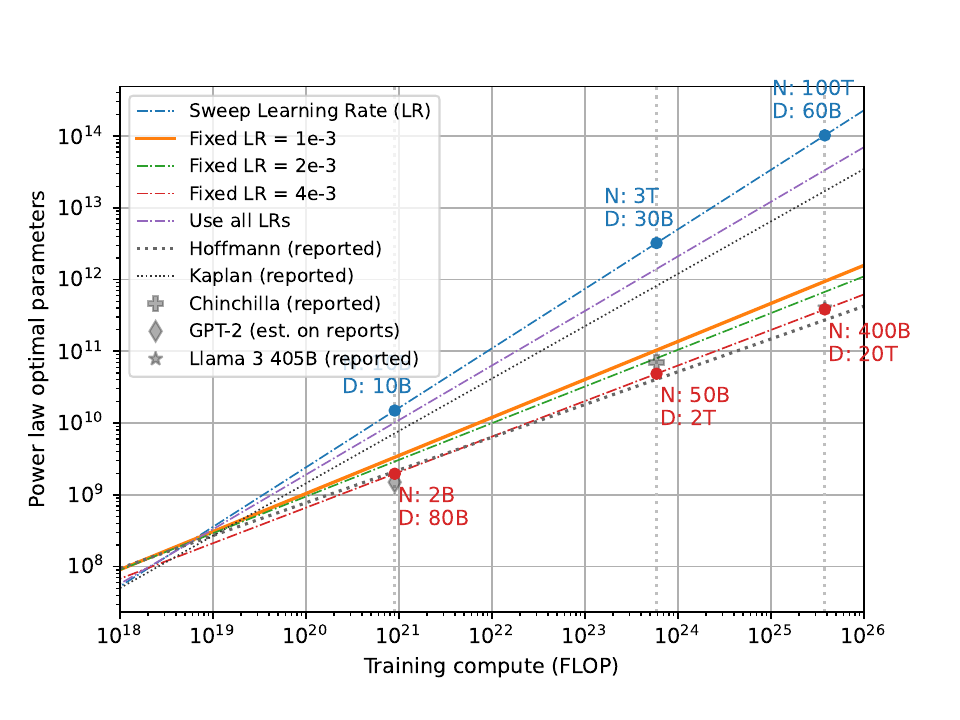}

    \caption{\textbf{\S\ref{sec:model_training}, \S\ref{sec:repl-model_training}} With our models, we simulate the effects of not sweeping the learning rate. As a baseline, (1) we sweep at each ($N$, $D$) pair for the optimal learning rate over a range of values at most a multiple of 2 apart. Next, (2) we use a learning rate of 1e-3 for all $N$, the optimal for our 1 billion parameter models, and do the same for (3) 2e-3 and (4) 4e-3, which is optimal for our 12 million parameter models. Lastly, we use all models across all learning rates at the same $N$ and $D$. Results vary dramatically across these settings. Somewhat surprisingly, using all learning rates results in a very similar power law to sweeping the learning rate, whereas using a fixed learning rate of 1e-3 or 4e-3 yields the lowest optimization loss or closest match to the \citet{hoffmann2022training} power laws, respectively.}
    \label{fig:analysis_lr_ours}
\end{subfigure}

\end{figure}
\begin{figure}[]
\ContinuedFloat
\centering 


\begin{subfigure}{\textwidth}
\begin{subfigure}{0.49\textwidth}
    \centering
    \includegraphics[width=\textwidth]{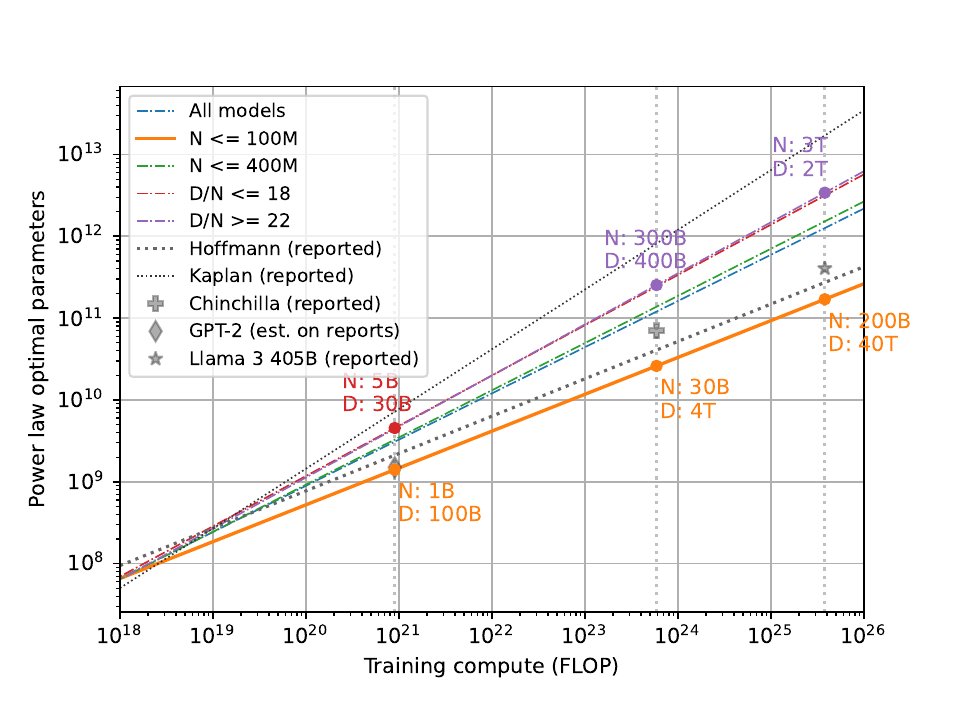}
    \footnotesize{\citet{hoffmann2022training,besiroglu2024chinchilla}}
    \label{fig:analysis_form_epoch}
\end{subfigure}
\hfill
\begin{subfigure}{0.49\textwidth}
    \centering
    \includegraphics[width=\textwidth]{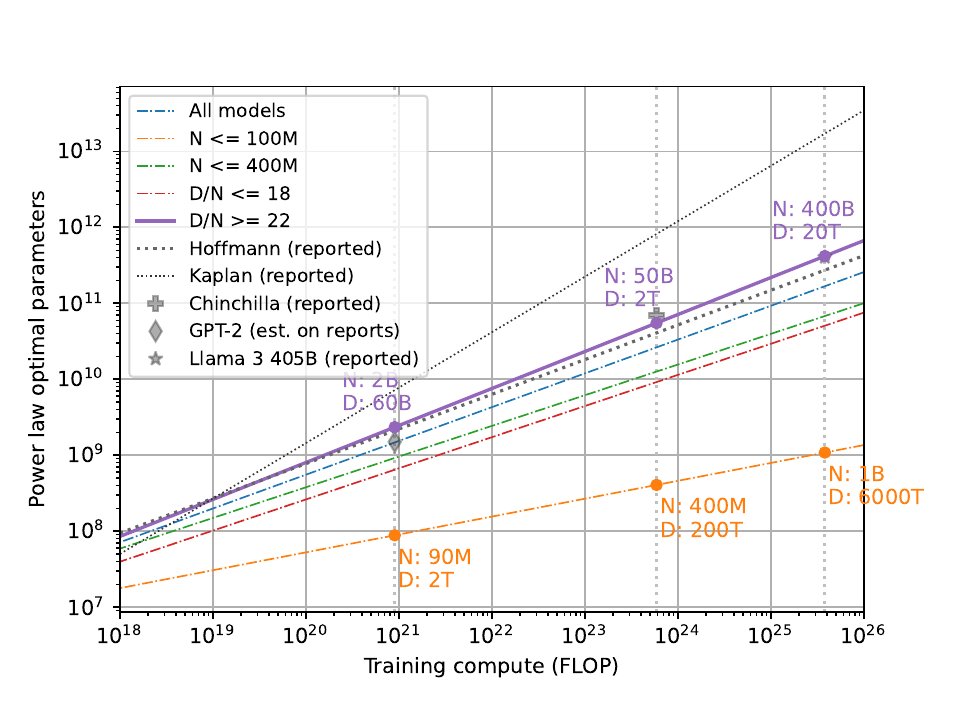}
    \footnotesize{\citet{porian2024resolving}}

    \label{fig:analysis_form_ours}
\end{subfigure}
\vspace{1em}
    \centering
\begin{subfigure}{0.49\textwidth}
    \centering
    \includegraphics[width=\textwidth]{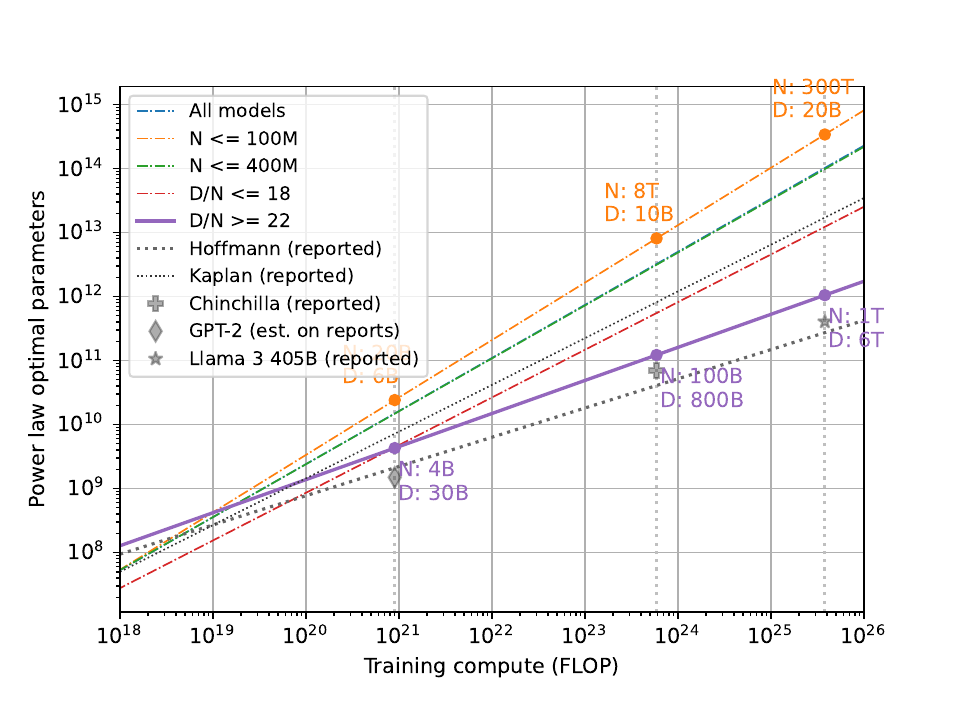}
    \footnotesize{Ours}
    \label{fig:analysis_form_ours}
\end{subfigure}
\vspace{1em}

    \caption{\textbf{\S\ref{sec:model_training}, \S\ref{sec:repl-model_training}} From all 3 datasets, we choose subsets with ($N$, $D$) values which fit a particular method one might have of setting up training. We fit with (1) all models, which for our dataset, ranges from 12 million to 1 billion parameters, then with (2) only models of up to about 100 million parameters or (3) up to 400 million parameters. We also compare the effects of a higher or lower hypothesis about the optimal $D/N$ ratio, including (4) only models with $D/N \leq 18$ or (5) $D/N \geq 22$. These ranges are designed to exclude $D/N = 20$, the rule of thumb based on \citet{hoffmann2022training}. The minimum or maximum $D/N$ ratio tested does skew results; above $10^{22}$ FLOPs, (4) and (5) fit to optimal ratios $D/N < 18$ and $D/N > 22$, respectively. Removing our largest models in (2) also creates a major shift in the predicted optimal $D/N$.}
    \label{fig:analysis_nd_ours}
\end{subfigure}


\end{figure}
\begin{figure}[]
\ContinuedFloat
\centering 

    
\begin{subfigure}{\textwidth}
\begin{subfigure}{0.49\textwidth}
    \centering
    \includegraphics[width=\textwidth]{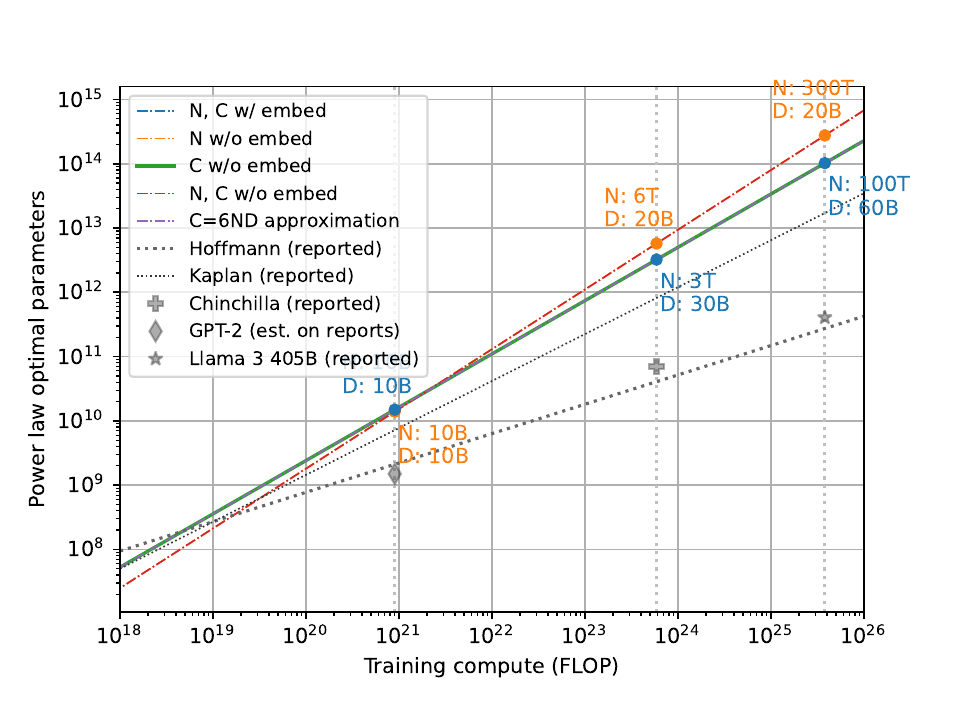}
    \footnotesize{Ours}
    \label{fig:analysis_form_epoch}
\end{subfigure}
\hfill
\begin{subfigure}{0.49\textwidth}
    \centering
    \includegraphics[width=\textwidth]{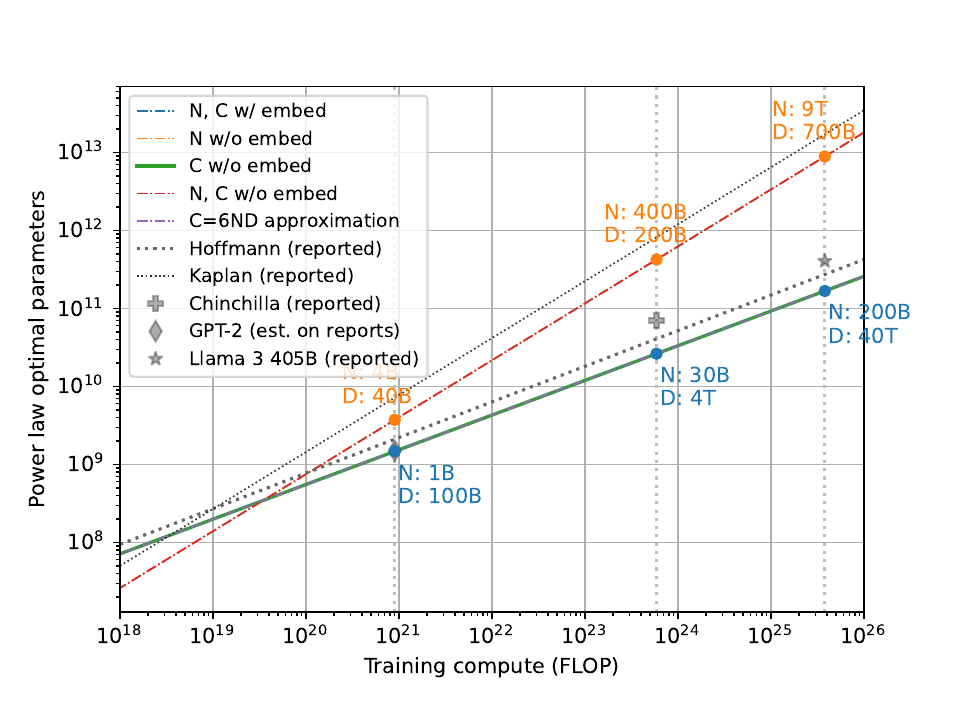}
    \footnotesize{\citet{porian2024resolving}}
    \label{fig:analysis_form_ours}
\end{subfigure}
\vspace{1em}

    \caption{\textbf{\S\ref{sec:model_training}, \S\ref{sec:repl-model_training}} With our own data and those from \citet{porian2024resolving}, we fit power laws to the same sets of models, while varying the ways we count $N$ and $C$. We compare (1) including embeddings, which is our baseline, with (2) excluding embeddings only in $N$, (3) excluding embeddings only in $C$, (4) excluding embeddings in both $N$ and $C$. We also compare to using the $C=6ND$ approximation, including embedding parameters. Throughout this work, we calculate the FLOPs in a manner similar to \citet{hoffmann2022training}, and we open source the code for these calculations.
    With both datasets, the exclusion of embeddings in FLOPs has very little impact on the final fit. Similarly, using the $C=6ND$ approximation has no visible impact. For the \citet{porian2024resolving} models, the exclusion of embedding parameters in the calculation of $N$ results in scaling laws which differ substantially, and with increasing divergences at large scales. 
    }
    \label{fig:analysis_counting_ours}
\end{subfigure}

\end{figure}
\begin{figure}[]
\ContinuedFloat
\centering 

\begin{subfigure}{\textwidth}
\centering
\begin{subfigure}{0.7\textwidth}
    \centering
    \includegraphics[width=\textwidth]{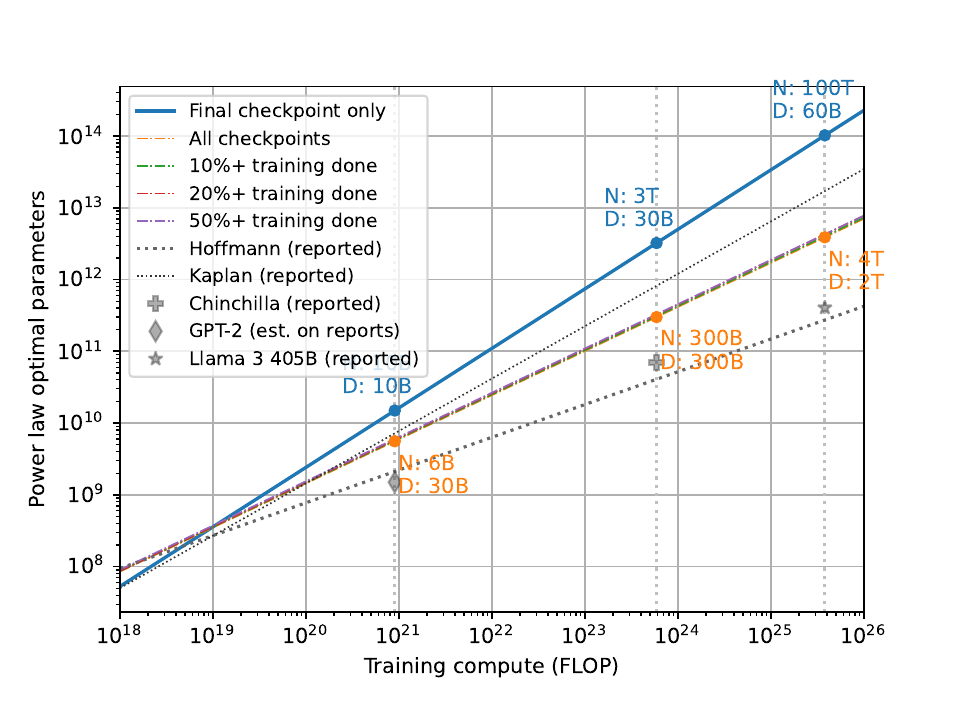}
\end{subfigure}
\vspace{1em}
\begin{subfigure}{0.49\textwidth}
    \centering
    \includegraphics[width=\textwidth]{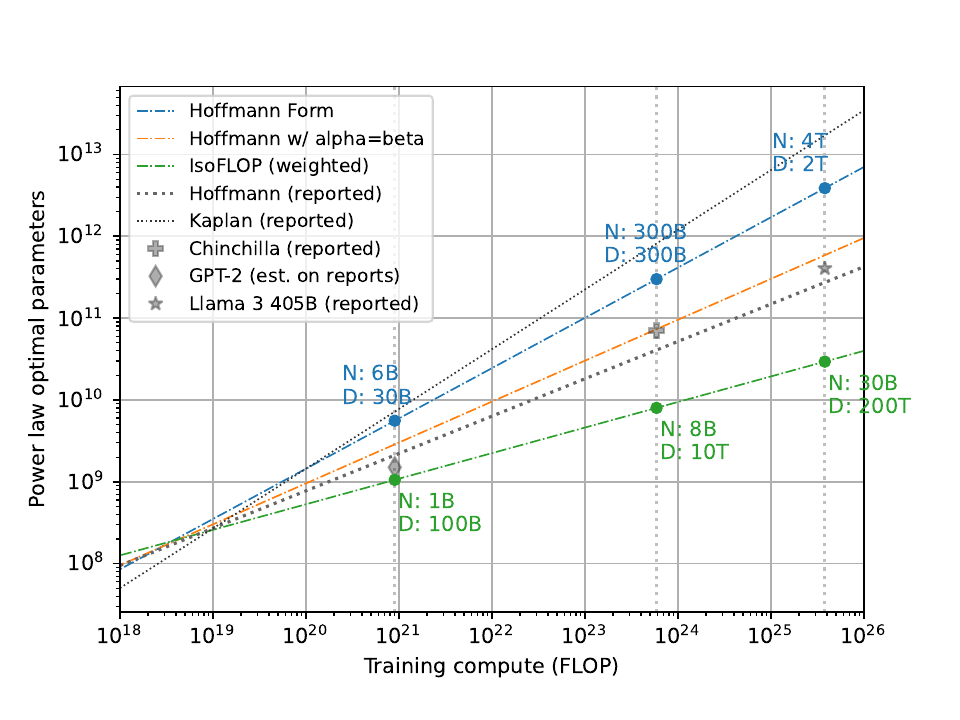}
\end{subfigure}
\hfill
\begin{subfigure}{0.49\textwidth}
    \centering
    \includegraphics[width=\textwidth]{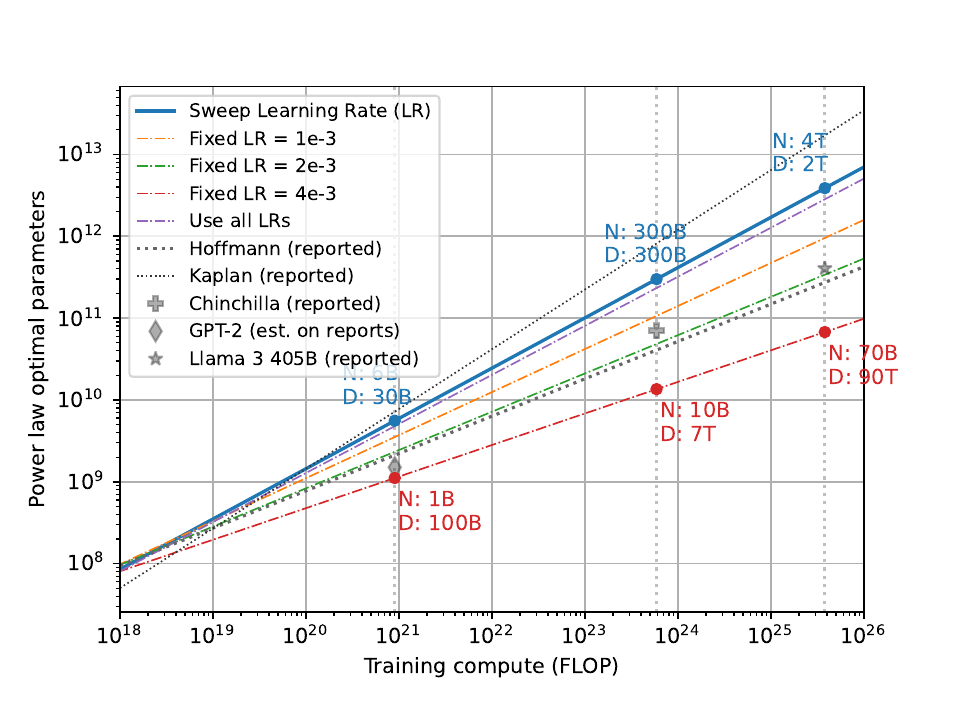}
\end{subfigure}
\vspace{1em}
\begin{subfigure}{0.49\textwidth}
    \centering
    \includegraphics[width=\textwidth]{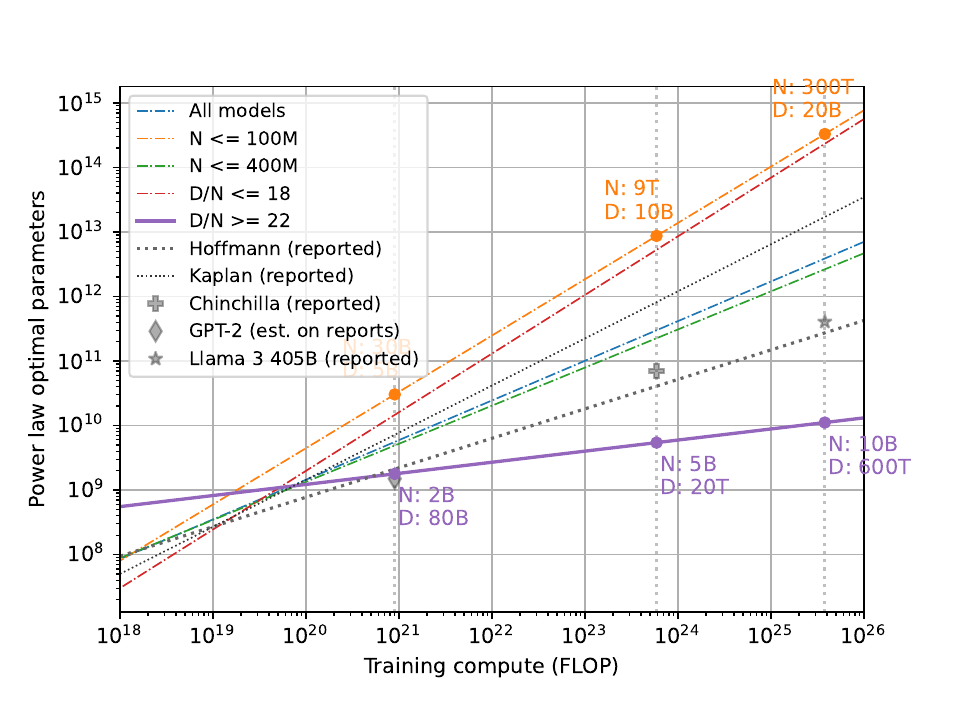}
\end{subfigure}
\hfill
\begin{subfigure}{0.49\textwidth}
    \centering
    \includegraphics[width=\textwidth]{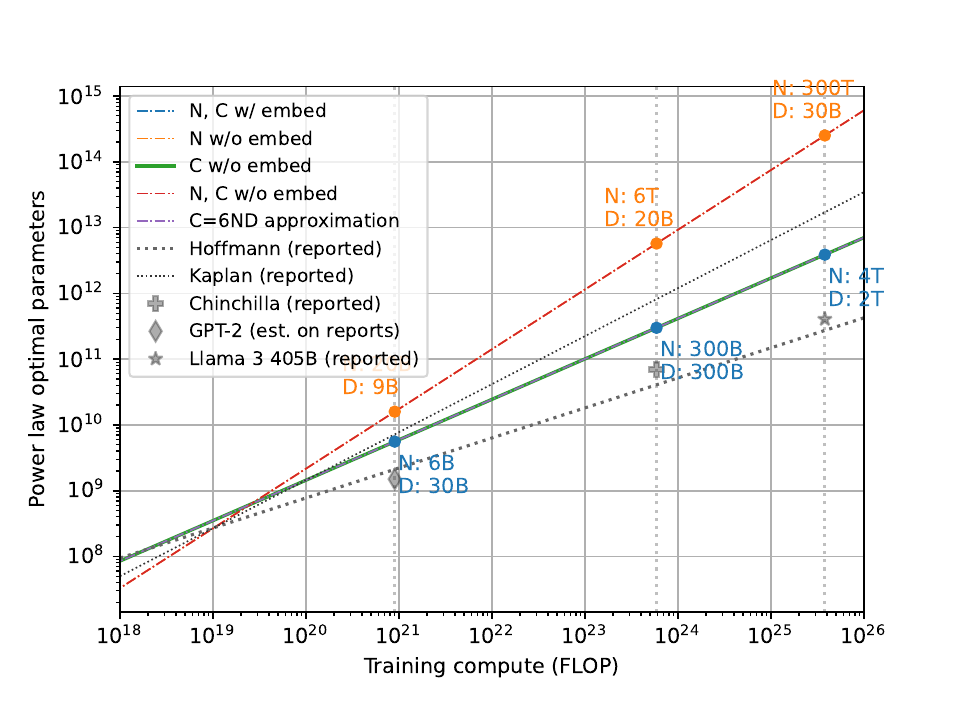}
\end{subfigure}

\caption{\textbf{\S\ref{sec:data}, \S\ref{sec:repl-data}} With our own dataset we compare (1) fitting a power law using only the final checkpoint with (2) using all mid-training checkpoints (3) using all checkpoints, starting 10\% through training, (4) the same, starting 20\% through training, and (5) the same again, starting 50\% through training. We observe that (2)-(5) consistently
yields power laws more similar to that reported by \citet{hoffmann2022training}, so we also repeat all analyses in Figures~\ref{fig:analysis_form}-\ref{fig:analysis_counting_ours}. Using mid-training checkpoints sometimes results in more stable fits which are similar to the \citet{hoffmann2022training} scaling laws, but the effect is unreliable and is dependent on other decisions. 
}
\label{fig:analysis_checkpoint_ours}        
\end{subfigure}

\end{figure}
\begin{figure}[]
\ContinuedFloat
\centering 

\begin{subfigure}{\textwidth}
\centering 
\begin{subfigure}{0.7\textwidth}
    \centering
    \includegraphics[width=\textwidth]{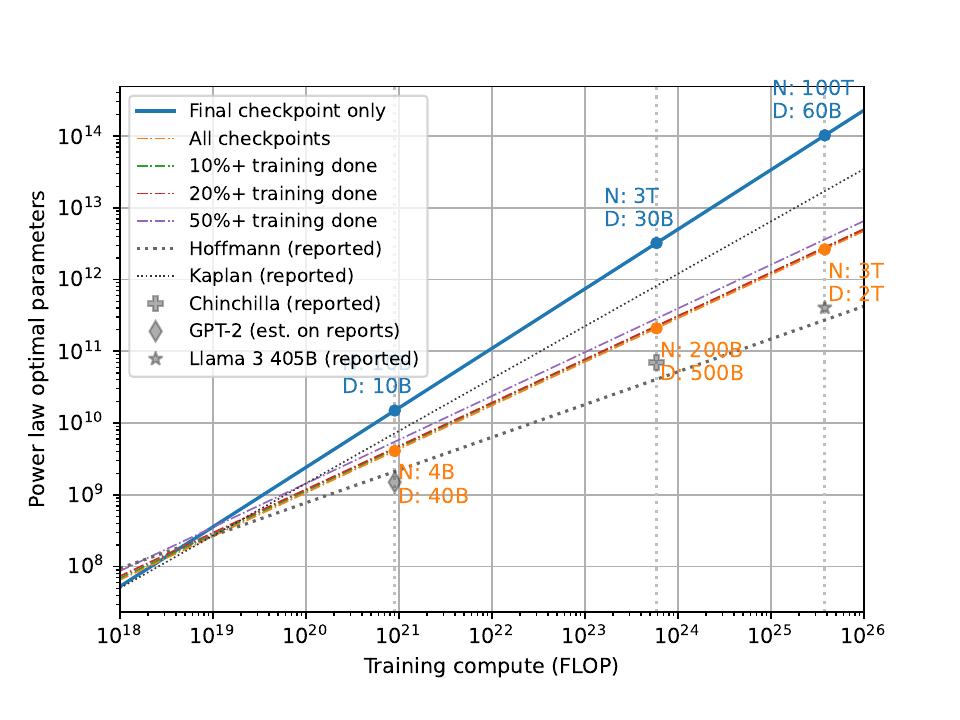}
\end{subfigure}
\\ \vspace{1em}
\begin{subfigure}{0.49\textwidth}
    \centering
    \includegraphics[width=\textwidth]{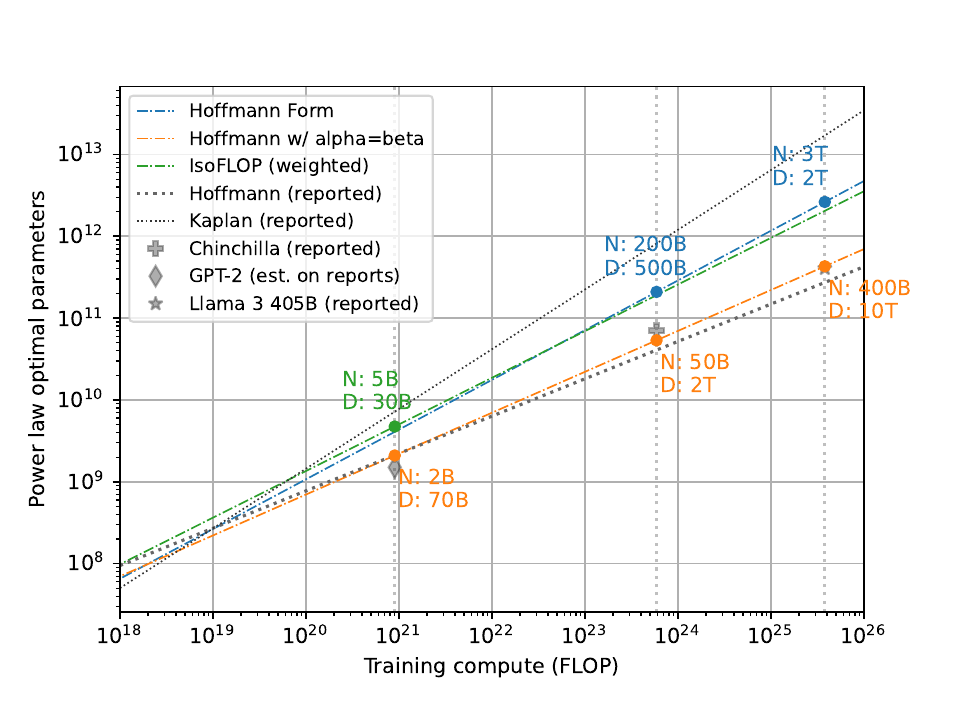}
\end{subfigure}
\hfill
\begin{subfigure}{0.49\textwidth}
    \centering
\end{subfigure}
\\ \vspace{1em}
\begin{subfigure}{0.49\textwidth}
    \centering
    \includegraphics[width=\textwidth]{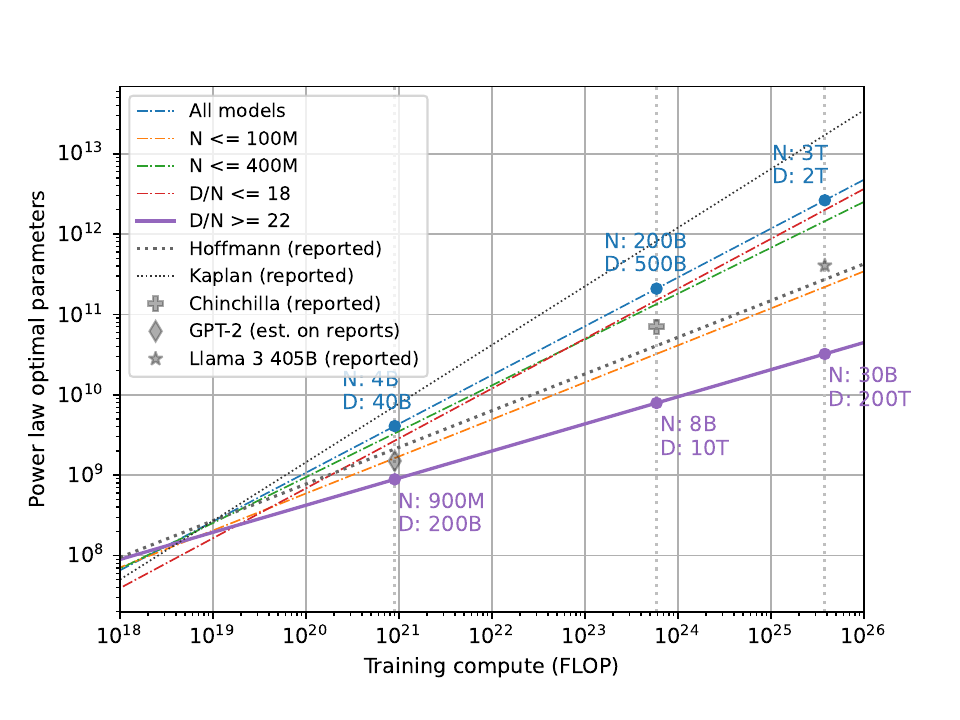}
\end{subfigure}
\hfill
\begin{subfigure}{0.49\textwidth}
    \centering
    \includegraphics[width=\textwidth]{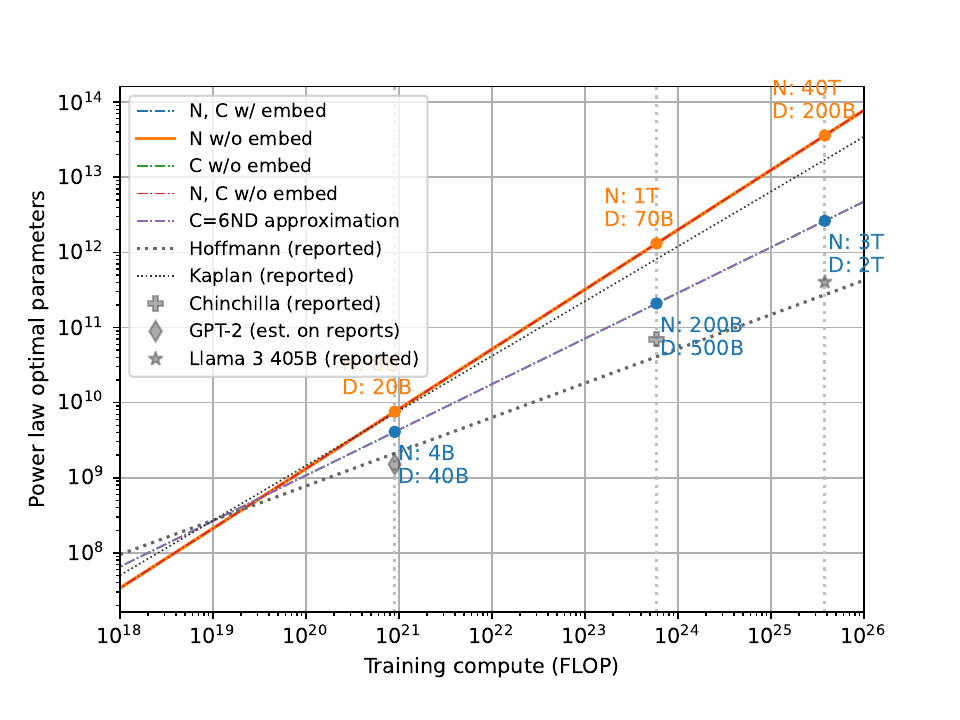}
\end{subfigure}

\caption{
\textbf{\S\ref{sec:data}, \S\ref{sec:repl-data}} With the \citet{porian2024resolving} dataset, we compare (1) fitting a power law using only the final checkpoint with (2) using all mid-training checkpoints (3) using all checkpoints, starting 10\% through training, (4) the same, starting 20\% through training, and (5) the same again, starting 50\% through training. We observe that (2)-(5) consistently yields power laws more similar to that reported by \citet{hoffmann2022training}, so we also repeat analyses in Figures~\ref{fig:analysis_form}~and~\ref{fig:analysis_counting_ours}. Using mid-training checkpoints sometimes results in more stable fits which are similar to the \citet{hoffmann2022training} scaling laws, but the effect is unreliable and is dependent on other decisions. 
}
\label{fig:analysis_checkpoint_rsld}        
\end{subfigure}

\end{figure}
\begin{figure}[]
\ContinuedFloat
\centering 



\begin{subfigure}{\textwidth}
\begin{subfigure}{0.49\textwidth}
    \centering
    \includegraphics[width=\textwidth]{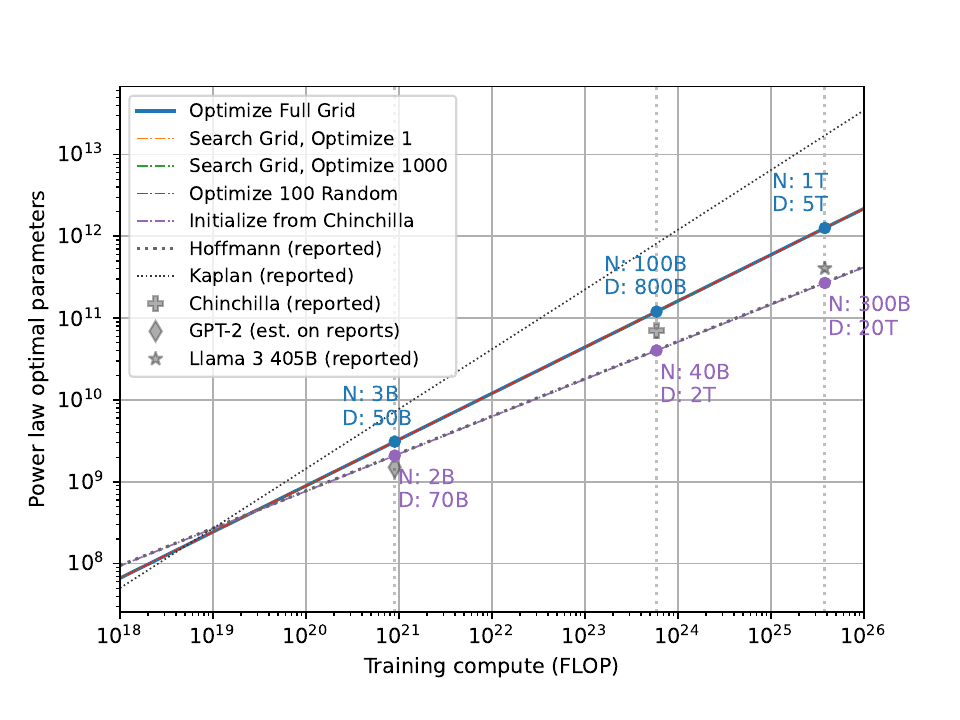}
    \footnotesize{\citet{hoffmann2022training,besiroglu2024chinchilla}}
\end{subfigure}
\hfill
\begin{subfigure}{0.49\textwidth}
    \centering
    \includegraphics[width=\textwidth]{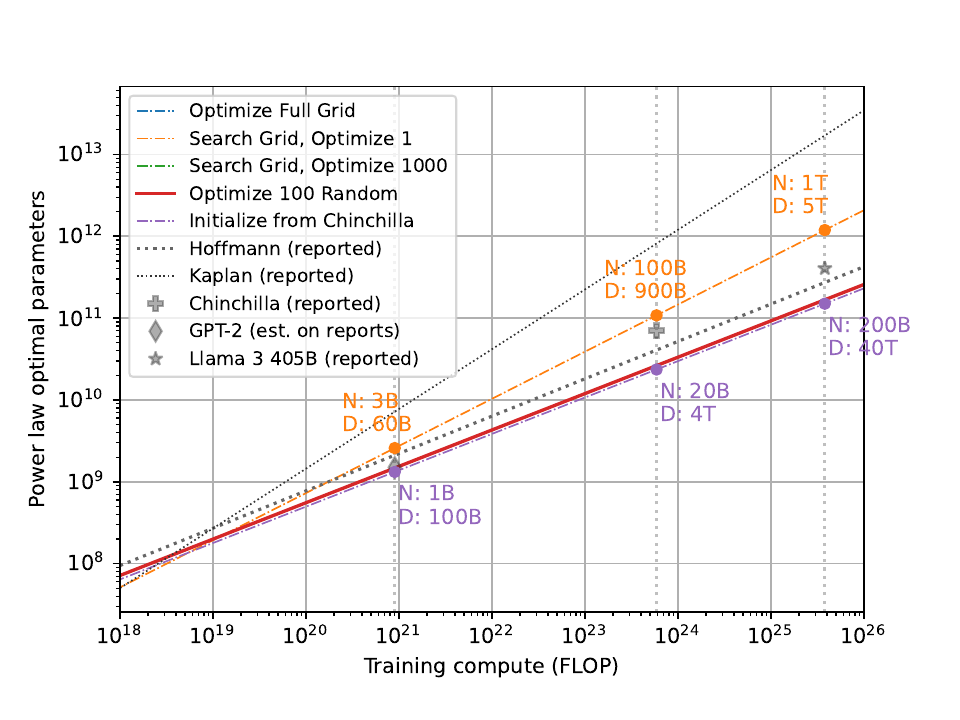}
    \footnotesize{\citet{porian2024resolving}}
\end{subfigure}
\vspace{1em}
    \centering
\begin{subfigure}{0.49\textwidth}
    \centering
    \includegraphics[width=\textwidth]{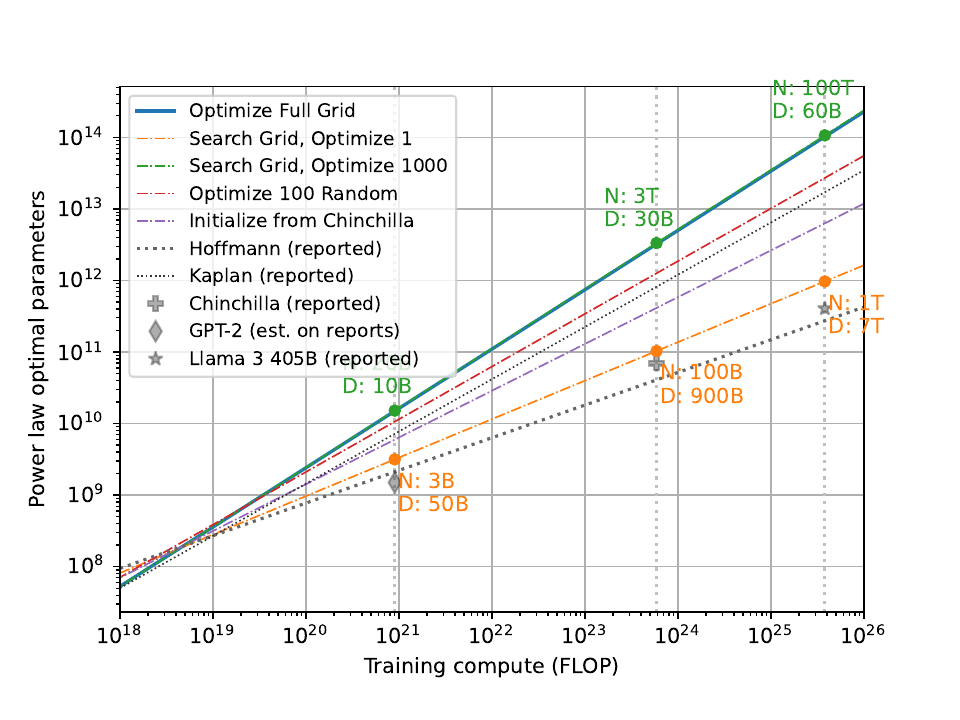}
    \footnotesize{Ours}
\end{subfigure}
\vspace{1em}
\caption{\textbf{\S\ref{sec:opt}, \S\ref{sec:repl-opt}} We fit to data from all 3 datasets to experiment with the initialization of parameters in the power law. We start with (1) optimizing every point in a grid search of 6x6x5x5x5=4500 initializations \citep{hoffmann2022training}, (2) searching for the lowest loss initialization point \citep{caballero2022broken} and optimizing only from that initialization, (3) optimizing from the 1000 lowest loss initializations, (4) randomly sampling 100 points, and (5) initializing with the coefficients found in \citet{hoffmann2022training}, as \citet{besiroglu2024chinchilla} does. With the \citet{besiroglu2024chinchilla} data, (5) yields a fit nearly identical to that reported by \citet{hoffmann2022training}, although (1) results in the lowest fitting loss. With the \citet{porian2024resolving} data, all approaches except (2) yield a very similar fit, which gives a recommended $D/N$ ratio similar to that of\citet{hoffmann2022training}. However, using our data, (2) optimizing over only the most optimal initialization yields the best match to the \citet{hoffmann2022training} power laws, followed by (5) initialization from the reported \citet{hoffmann2022training} scaling law parameters. Optimizing over the full grid frequently yields the power law which diverges most from the \citet{hoffmann2022training} law, suggesting the difficulty of optimizing over this space, and the presence of many local minima.
}
\label{fig:analysis_init}
\end{subfigure}

\end{figure}
\begin{figure}[]
\ContinuedFloat
\centering 

\begin{subfigure}{\textwidth}
\begin{subfigure}{0.49\textwidth}
    \centering
    \includegraphics[width=\textwidth]{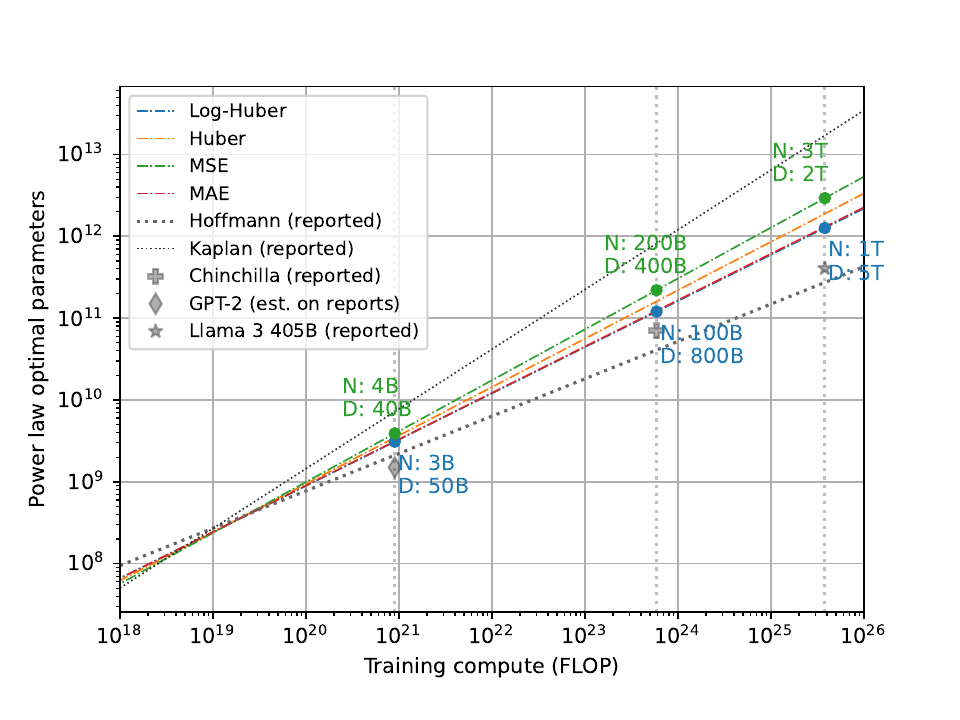}
    \footnotesize{\citet{hoffmann2022training,besiroglu2024chinchilla}}
\end{subfigure}
\hfill
\begin{subfigure}{0.49\textwidth}
    \centering
    \includegraphics[width=\textwidth]{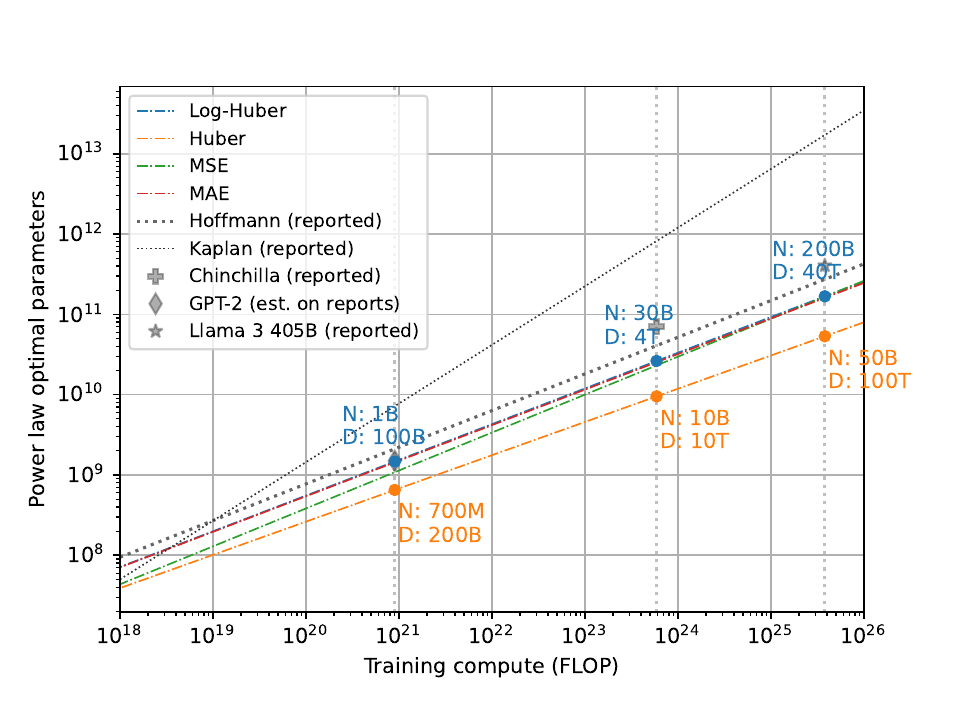}
    \footnotesize{\citet{porian2024resolving}}
\end{subfigure}
\vspace{1em}
    \centering
\begin{subfigure}{0.49\textwidth}
    \centering
    \includegraphics[width=\textwidth]{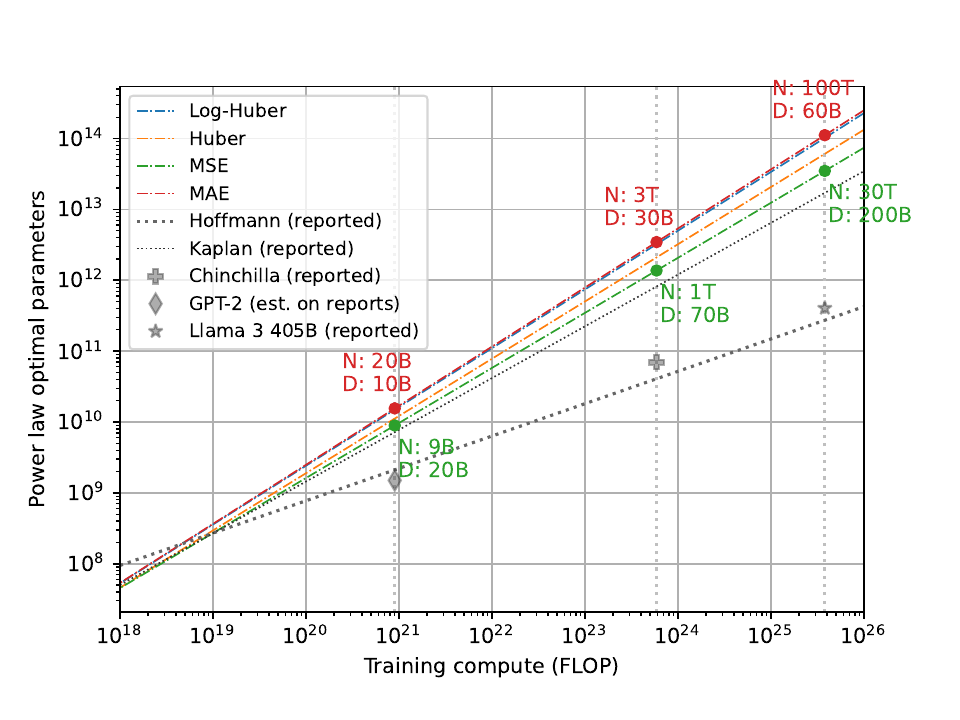}
    \footnotesize{Ours}
\end{subfigure}
\vspace{1em}
\caption{\textbf{\S\ref{sec:opt}, \S\ref{sec:repl-opt}} We fit a power law to data from all 3 datasets, minimizing different objective functions: (1) the baseline log-Huber loss, (2) MSE, (3) MAE, and (4) the Huber loss. 
The resulting power laws are less disparate than when varying many of the other factors discussed above and generally fall near the power law parameters reported by \citet{kaplan2020scaling} and \citet{hoffmann2022training}, but this is still a wide range of recommended optimal parameter counts for each compute budget. Overall, the loss function behavior is not predictable, given the differences between loss functions when looking at the power laws resulting from these three sources of data.}
\label{fig:analysis_loss}
\end{subfigure}


\end{figure}
\begin{figure}[]
\ContinuedFloat
\centering

\begin{subfigure}{\textwidth}
\begin{subfigure}{0.49\textwidth}
    \centering
    \includegraphics[width=\textwidth]{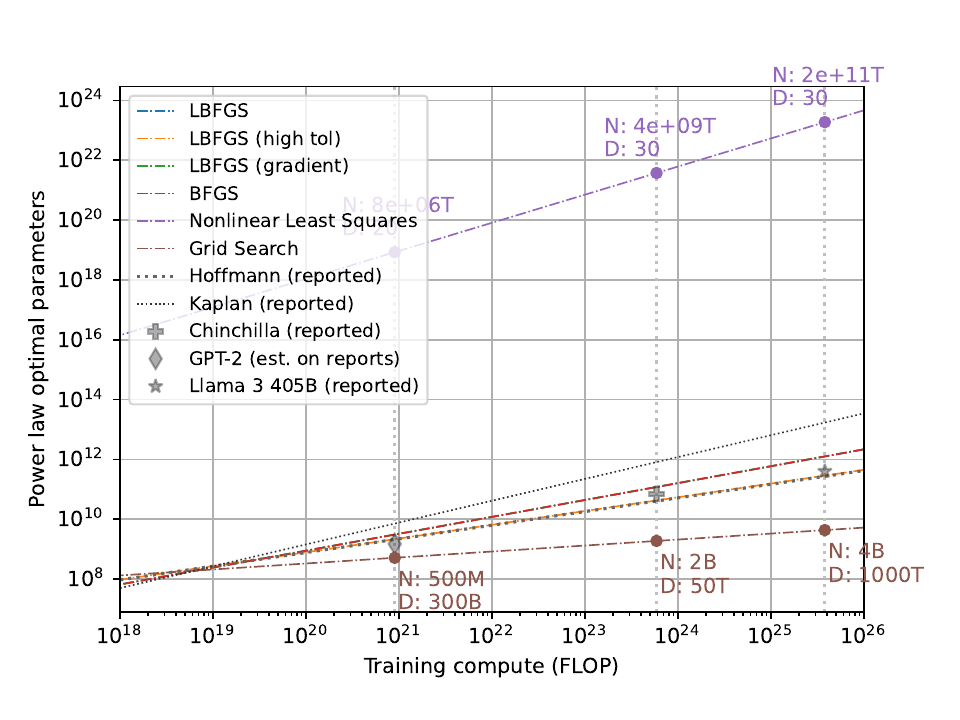}
    \footnotesize{\citet{hoffmann2022training,besiroglu2024chinchilla}}
    \label{fig:analysis_opt_epoch}
\end{subfigure}
\hfill
\begin{subfigure}{0.49\textwidth}
    \centering
    \includegraphics[width=\textwidth]{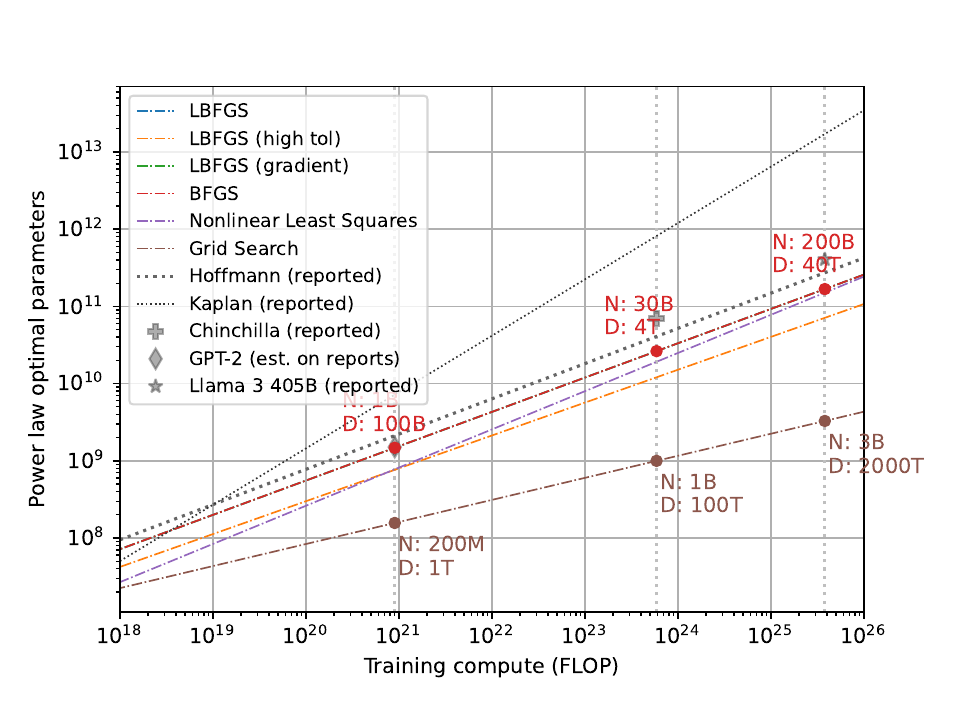}
    \footnotesize{\citet{porian2024resolving}}
    \label{fig:analysis_opt_ours}
\end{subfigure}
\vspace{1em}
    \centering
\begin{subfigure}{0.49\textwidth}
    \centering
    \includegraphics[width=\textwidth]{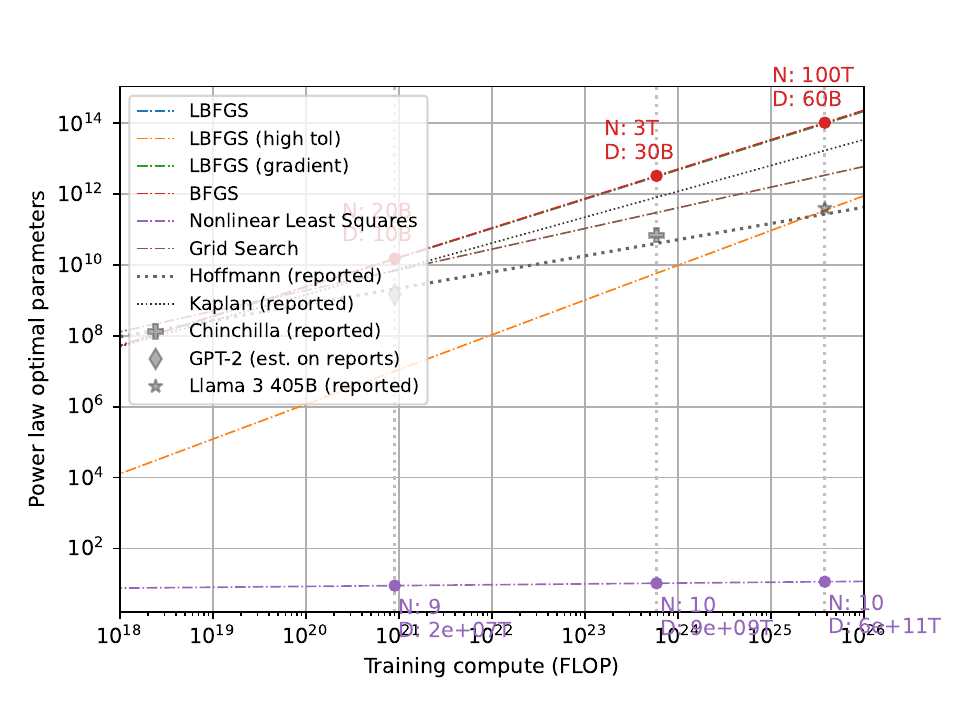}
    \footnotesize{Ours}
    \label{fig:analysis_opt_ours}
\end{subfigure}
\caption{\textbf{\S\ref{sec:opt}, \S\ref{sec:repl-opt}} We fit a power law to data from \citet{besiroglu2024chinchilla} (left) and our data (right) using various optimizers, beginning with the original (1) L-BFGS. L-BFGS and BFGS implementations have an early stopping mechanism, which conditions on the stability of the solution between optimization steps. We set this threshold for L-BFGS to a (2) higher value 1e-4 (stopping earlier).
We found that using any higher or lower values resulted in the same solutions or instability for all three datasets. L-BFGS and BFGS also have the option to use the true gradient of the loss, instead of an estimate, which is the default. In this figure, we include this setting, (3) using the true gradient for L-BFGS. We compare these L-BFGS settings to (4) BFGS. We test the same tolerance value and gradient settings, and find that none of these options change the outcome of BFGS in our analysis, and omit them from this figure for legibility. Finally, we compare to (5) non-linear least squares and (6) pure grid search, using a grid 5 times more dense along each axis as we used initialization with other optimizers. This density is chosen to approximately match the runtime of L-BFGS.
Many of these optimizers do converge to similar solutions, but this depends on the data, and the settings which diverge from this majority do not follow any immediately evident pattern.} 
\label{fig:analysis_opt}

\end{subfigure}

\end{figure}
\begin{figure}[]
\ContinuedFloat
\centering 

\caption{(\textbf{\S\ref{sec:own-repl}}) We study the effects of various decisions in the fitting of a power law, as outlined in our checklist (Appendix~\ref{sec:app_checklist}) and detailed in \S\ref{sec:power-law-form}-\S\ref{sec:opt}. For each set of analyses, we the scaling laws found by \citep{kaplan2020scaling} and \citep{hoffmann2022training} for comparison. We also include markers indicating 3 existing models for comparison purposes: Llama 3 405B \citep{dubey2024llama}, the Chinchilla model \citep{hoffmann2022training}, and an estimate of the 1.5B GPT-2 model \citep{radford2019language}, for which we know details of the dataset storage size and word count, but not an exact count of data BPE tokens, which we estimate at 100B. We additionally annotate, at the compute budget $C$ for each of these 3 reference points, the maximum and minimum \textit{predicted} (i.e. extrapolated) optimal model parameter count $N_{opt}$ and data budget $D_{opt}$ from the fitted power laws. We use a thicker, solid line for the method in each plot which achieves the lowest optimization loss, with the exception of the plots comparing power law form, those comparing loss functions and those comparing optimizers, for which this would be nonsensical.
We find overall, throughout our analyses, that all of the decisions we explore have an impact on the final fit of the power law, supporting our conclusion that more thorough reporting of these decisions is critical for scaling law reproducibility.
}
\label{fig:overall}
\end{figure}

\subsection{Form (\S\ref{sec:power-law-form})} \label{sec:repl-power-law-form}

We consider the (1) baseline \citet{hoffmann2022training} form (Approach 3) and then (2) apply the trick employed by \citet{muennighoff2024scaling} of setting $\alpha = \beta$ in the \citet{hoffmann2022training} form, which assumes the optimal $D/N$ ratio stays roughly constant -- $L(N, D) = E + \frac{A}{N^\alpha} + \frac{B}{D^\alpha}$. We also compare with (3) an ISOFlop approach (Approach 3 of \citet{hoffmann2022training}), in which we fit an optimal $N$ for each compute budget $C$, $C(N)$, which can then be used to fit the predicted loss $L(N, D)$. This approach usually necessitates the usage of mid-training checkpoints (discussed further in \S\ref{sec:repl-data}), as it is infeasible to train a large enough number of models for each FLOP budget considered. However, we apply it here without using only final model checkpoints, and extend to mid-training checkpoints in \S\ref{sec:repl-data}). We adapt the implementation from \citet{porian2024resolving}, which contains more details about interpolation of data points and specific hyperparameters. In all data sets, (2) approaches the law reported by \citet{hoffmann2022training}, but (1) only does so for the data from \citet{porian2024resolving}(Figure~\ref{fig:analysis_form}).

We experiment with the power law form reported by \citet{kaplan2020scaling}, but this consistently yields a law which suggests that the optimal number of data points is 1, even when varying many aspects of the power law fitting procedure. The difficulty of fitting to this form might be partially a result of severe under-reporting in \citet{kaplan2020scaling} with regard to procedural details, including hyperparameters for both model training and fitting. 



\subsection{Training (\S\ref{sec:model_training})} \label{sec:repl-model_training}

\citet{hu2024minicpm} study the effects of several model training decisions, including batch size, learning rate schedule, and model width. Their analysis focuses on optimizing hyperparameters, not on the ways hyperparameter and architecture choices affect the reliability of scaling law fitting. Observed variations between settings suggest that suboptimal performance could skew the scaling law fit. 

To substantiate this further, we simulate the effects of not sweeping the learning rate in our models. As a baseline, (1) we sweep at each ($N$, $D$) pair for the optimal learning rate over a range of values, at most a multiple of 2 apart. Next, (2) we use a learning rate of 1e-3 for all $N$, the optimal for our 1 billion parameter models, and do the same for (3) 2e-3 and (4) 4e-3, which is optimal for our 12 million parameter models. Lastly, we use all models across all learning rates at the same $N$ and $D$. Results vary dramatically across these settings. Somewhat surprisingly, using all learning rates results in a very similar power law to sweeping the learning rate, whereas using a fixed learning rate of 1e-3 or 4e-3 yields the lowest optimization loss or closest match to the \citet{hoffmann2022training} power laws, respectively (Figure~\ref{fig:analysis_lr_ours}).

We also study the effects of limiting model scale range, data scale range (implicitly), and data-to-parameters range by filtering all 3 datasets: we compare (1) using all $N, D$ scales, (2) only models with $N$ up to about 100 million or (3) 400 million parameters, and including (4) only models with $D/N \leq 18$ or (5) $D/N \geq 22$. These ranges are designed to exclude $D/N = 20$, the rule of thumb based on \citet{hoffmann2022training}. The minimum or maximum $D/N$ ratio tested does skew results; above $10^{22}$ FLOPs, (4) and (5) fit to optimal ratios $D/N < 18$ and $D/N > 22$, respectively. Removing our largest models in (2) also creates a major shift in the predicted optimal $D/N$ (Figure~\ref{fig:analysis_nd_ours}).

Related to choice of $N, D$ and $C$ values, we investigate different ways of counting $N$ and $C$, specifically whether to include embedding parameters and FLOPs. We compare (1) our baseline, which includes embeddings in both $N$ and $C$, with (2) excluding embeddings only in $N$, (3) excluding embeddings only in $C$, (4) excluding embeddings in both $N$ and $C$. We also compare to (5) using the $C=6ND$ approximation, including embedding parameters. In all other settings, we calculate the FLOPs in a manner similar to \citet{hoffmann2022training}, and we open source the code for these calculations.
With both datasets, the exclusion of embeddings in FLOPs has very little impact on the final fit. Similarly, using the $C=6ND$ approximation has no visible impact. For the \citet{porian2024resolving} models, the exclusion of embedding parameters in the calculation of $N$ results in scaling laws which differ substantially, and with increasing divergences at large scales (Figure~\ref{fig:analysis_counting_ours}). 

\subsection{Data Collection (\S\ref{sec:data})} \label{sec:repl-data}


For our own models and those trained in \citet{porian2024resolving}, we compare using (1) only the final checkpoints of each model with (2) using all collected mid-training checkpoints. We also consider removing checkpoints collected during the (3) first 10\%, (4) first 20\%, and (5) first 50\% of training,
Using mid-training checkpoints sometimes results in more stable fits which are similar to the \citet{hoffmann2022training} scaling laws, but the effect is unreliable and is dependent on other decisions.  Following this finding, we also re-run all other analyses using setting (2), and find that these power law fits are often more consistent when varying other aspects of the power law fitting procedure (Figures~\ref{fig:analysis_checkpoint_ours}~and~\ref{fig:analysis_checkpoint_rsld}).

For ease of comparison, we replicate all graphs, reorganizing those in Figures~\ref{fig:analysis_checkpoint_ours}~and~\ref{fig:analysis_checkpoint_rsld} with their checklist topics, to allow for easier comparison between results using mid-training checkpoints and only final checkpoints. See Appendix~\S\ref{sec:app_megafigure} Figure~\ref{fig:app_mega}.




\subsection{Fitting (\S\ref{sec:opt})} \label{sec:repl-opt}
We vary the initialization method for our power law fitting procedure: (1) our baseline replication of \citet{hoffmann2022training} Approach 3, which conducts the full optimization process on a grid of 6x6x5x5x5=4500 initializations \citep{hoffmann2022training}, (2) searching for the lowest loss initialization point \citep{caballero2022broken} and optimizing only from that initialization, (3) optimizing from the 1000 lowest loss initializations, (4) randomly sampling $k$(=100) points \citep{frantar2023scaling,tao2024scaling}, and (5) initializing with the coefficients found in \citet{hoffmann2022training}, as \citet{besiroglu2024chinchilla} does.
With the \citet{besiroglu2024chinchilla} data, (5) yields a fit nearly identical to that reported by \citet{hoffmann2022training}, although (1) results in the lowest fitting loss. With the \citet{porian2024resolving} data, all approaches except (2) yield a very similar fit, which gives a recommended $D/N$ ratio similar to that of\citet{hoffmann2022training}. However, using our data, (2) optimizing over only the most optimal initialization yields the best match to the \citet{hoffmann2022training} power laws, followed by (5) initialization from the reported \citet{hoffmann2022training} scaling law parameters. Optimizing over the full grid yields the power law that diverges most from the \citet{hoffmann2022training} law, suggesting the difficulty of optimizing over this space, and the presence of many local minima (Figure~\ref{fig:analysis_init}).

Then, we analyze the choice of loss objectives, including (1) the baseline log-Huber loss, (2) MSE, (3) MAE, and (4) the Huber loss. We find that there is somewhat less variance in the resulting power laws than we have seen in other power law fitting decisions, but the recommended optimal data budgets still span a wide range (Figure~\ref{fig:analysis_loss}).

Finally, we consider the choice of optimizer. We first fit a power law to all 3 datasets using the original (1) L-BFGS, with settings matching those described by \citet{hoffmann2022training}. L-BFGS and BFGS implementations have an early stopping mechanism, which conditions on the stability of the solution between optimization steps. We experiment with setting this threshold for L-BFGS to a (2) higher value 1e-4 (stopping earlier).
We find that using any higher or lower values results in the same solutions or instability for all three datasets. L-BFGS and BFGS also have the option to use the true gradient of the loss, instead of an estimate, which is the default. In this figure, we include this setting, (3) using the true gradient for L-BFGS. We compare these L-BFGS settings to (4) BFGS. We test the same tolerance value and gradient settings, and find that none of these options change the outcome of BFGS in our analysis. Finally, we compare to (5) non-linear least squares and (6) pure grid search, using a grid 5 times more dense along each axis as that which we used for initialization with other optimizers. This density is chosen to approximately match the runtime of L-BFGS.
Many of these optimizer settings converge to similar solutions, but this depends on the data, and the settings which diverge from this majority do not follow any immediately evident pattern (Figure~\ref{fig:analysis_opt}).

\section{Conclusion}\label{sec:conclusion}

We survey over 50 papers on scaling laws, and discuss differences in form, training setup, evaluation, and curve fitting, which may lead to significantly different conclusions. We also discuss significant under-reporting of details crucial to replicating the conclusion of these studies, and provide guidelines in the form of a checklist aid researchers in reporting more complete details. In addition to discussing several prior replication studies in literature, we empirically demonstrate the fragility of this process, by systematically varying these choices on available checkpoints and models that we train from scratch.
We choose to avoid overly prescriptive recommendations, because there is no known set of actions which can guarantee a good scaling law fit, but we make some suggestions based on patterns in our findings (Appendix \S\ref{sec:app_recs}).
Despite our preliminary investigations, our understanding of which decisions may skew the results of a scaling law study is sparse, and defines the path for future work.

\paragraph{Ethics Statement} This work discusses how a lack of reproducibility and open-sourcing may be harmful for scaling laws research, given that the factors in a study setup that may change research conclusions vary widely.

\paragraph{Reproducibility Statement} The model checkpoints and analysis code required to reproduce the results discussed in Section \ref{sec:own-repl} are at \url{https://github.com/hadasah/scaling_laws}.



\subsubsection*{Acknowledgments}

We thank Aaron Defazio, Divyansh Pareek, Aditya Kusupati and Tim Althoff for their valuable feedback. We also acknowledge the computing resources and support from the Hyak supercomputer system at the
University of Washington.

\bibliography{iclr2024_conference}

\begin{thebibliography}{80}
\providecommand{\natexlab}[1]{#1}
\providecommand{\url}[1]{\texttt{#1}}
\expandafter\ifx\csname urlstyle\endcsname\relax
  \providecommand{\doi}[1]{doi: #1}\else
  \providecommand{\doi}{doi: \begingroup \urlstyle{rm}\Url}\fi

\bibitem[Aghajanyan et~al.(2023)Aghajanyan, Yu, Conneau, Hsu, Hambardzumyan, Zhang, Roller, Goyal, Levy, and Zettlemoyer]{aghajanyan2023scaling}
Armen Aghajanyan, Lili Yu, Alexis Conneau, Wei-Ning Hsu, Karen Hambardzumyan, Susan Zhang, Stephen Roller, Naman Goyal, Omer Levy, and Luke Zettlemoyer.
\newblock Scaling laws for generative mixed-modal language models.
\newblock In \emph{International Conference on Machine Learning}, pp.\  265--279. PMLR, 2023.

\bibitem[Alabdulmohsin et~al.(2022)Alabdulmohsin, Neyshabur, and Zhai]{alabdulmohsin2022revisiting}
Ibrahim~M Alabdulmohsin, Behnam Neyshabur, and Xiaohua Zhai.
\newblock Revisiting neural scaling laws in language and vision.
\newblock \emph{Advances in Neural Information Processing Systems}, 35:\penalty0 22300--22312, 2022.

\bibitem[Amodei et~al.(2016)Amodei, Ananthanarayanan, Anubhai, Bai, Battenberg, Case, Casper, Catanzaro, Cheng, Chen, et~al.]{amodei2016deep}
Dario Amodei, Sundaram Ananthanarayanan, Rishita Anubhai, Jingliang Bai, Eric Battenberg, Carl Case, Jared Casper, Bryan Catanzaro, Qiang Cheng, Guoliang Chen, et~al.
\newblock Deep speech 2: End-to-end speech recognition in english and mandarin.
\newblock In \emph{International conference on machine learning}, pp.\  173--182. PMLR, 2016.

\bibitem[Anil et~al.(2023)Anil, Dai, Firat, Johnson, Lepikhin, Passos, Shakeri, Taropa, Bailey, Chen, et~al.]{anil2023palm}
Rohan Anil, Andrew~M Dai, Orhan Firat, Melvin Johnson, Dmitry Lepikhin, Alexandre Passos, Siamak Shakeri, Emanuel Taropa, Paige Bailey, Zhifeng Chen, et~al.
\newblock Palm 2 technical report.
\newblock \emph{arXiv preprint arXiv:2305.10403}, 2023.

\bibitem[Ardalani et~al.(2022)Ardalani, Wu, Chen, Bhushanam, and Aziz]{ardalani2022understanding}
Newsha Ardalani, Carole-Jean Wu, Zeliang Chen, Bhargav Bhushanam, and Adnan Aziz.
\newblock Understanding scaling laws for recommendation models.
\newblock \emph{arXiv preprint arXiv:2208.08489}, 2022.

\bibitem[Bahri et~al.(2024)Bahri, Dyer, Kaplan, Lee, and Sharma]{bahri2021explaining}
Yasaman Bahri, Ethan Dyer, Jared Kaplan, Jaehoon Lee, and Utkarsh Sharma.
\newblock Explaining neural scaling laws.
\newblock \emph{Proceedings of the National Academy of Sciences}, 121\penalty0 (27), June 2024.
\newblock ISSN 1091-6490.
\newblock \doi{10.1073/pnas.2311878121}.
\newblock URL \url{http://dx.doi.org/10.1073/pnas.2311878121}.

\bibitem[Banko \& Brill(2001)Banko and Brill]{banko2001scaling}
Michele Banko and Eric Brill.
\newblock Scaling to very very large corpora for natural language disambiguation.
\newblock In \emph{Proceedings of the 39th annual meeting of the Association for Computational Linguistics}, pp.\  26--33, 2001.

\bibitem[Bansal et~al.(2022)Bansal, Ghorbani, Garg, Zhang, Cherry, Neyshabur, and Firat]{bansal2022data}
Yamini Bansal, Behrooz Ghorbani, Ankush Garg, Biao Zhang, Colin Cherry, Behnam Neyshabur, and Orhan Firat.
\newblock Data scaling laws in nmt: The effect of noise and architecture.
\newblock In \emph{International Conference on Machine Learning}, pp.\  1466--1482. PMLR, 2022.

\bibitem[Besiroglu et~al.(2024)Besiroglu, Erdil, Barnett, and You]{besiroglu2024chinchilla}
Tamay Besiroglu, Ege Erdil, Matthew Barnett, and Josh You.
\newblock Chinchilla scaling: A replication attempt.
\newblock \emph{arXiv preprint arXiv:2404.10102}, 2024.

\bibitem[Bi et~al.(2024)Bi, Chen, Chen, Chen, Dai, Deng, Ding, Dong, Du, Fu, et~al.]{bi2024deepseek}
Xiao Bi, Deli Chen, Guanting Chen, Shanhuang Chen, Damai Dai, Chengqi Deng, Honghui Ding, Kai Dong, Qiushi Du, Zhe Fu, et~al.
\newblock Deepseek llm: Scaling open-source language models with longtermism.
\newblock \emph{arXiv preprint arXiv:2401.02954}, 2024.

\bibitem[Black et~al.(2022)Black, Biderman, Hallahan, Anthony, Gao, Golding, He, Leahy, McDonell, Phang, Pieler, Prashanth, Purohit, Reynolds, Tow, Wang, and Weinbach]{black2022gptneox20bopensourceautoregressivelanguage}
Sid Black, Stella Biderman, Eric Hallahan, Quentin Anthony, Leo Gao, Laurence Golding, Horace He, Connor Leahy, Kyle McDonell, Jason Phang, Michael Pieler, USVSN~Sai Prashanth, Shivanshu Purohit, Laria Reynolds, Jonathan Tow, Ben Wang, and Samuel Weinbach.
\newblock Gpt-neox-20b: An open-source autoregressive language model, 2022.
\newblock URL \url{https://arxiv.org/abs/2204.06745}.

\bibitem[Caballero et~al.(2022)Caballero, Gupta, Rish, and Krueger]{caballero2022broken}
Ethan Caballero, Kshitij Gupta, Irina Rish, and David Krueger.
\newblock Broken neural scaling laws.
\newblock \emph{arXiv preprint arXiv:2210.14891}, 2022.

\bibitem[Cherti et~al.(2023)Cherti, Beaumont, Wightman, Wortsman, Ilharco, Gordon, Schuhmann, Schmidt, and Jitsev]{cherti2023reproducible}
Mehdi Cherti, Romain Beaumont, Ross Wightman, Mitchell Wortsman, Gabriel Ilharco, Cade Gordon, Christoph Schuhmann, Ludwig Schmidt, and Jenia Jitsev.
\newblock Reproducible scaling laws for contrastive language-image learning.
\newblock In \emph{Proceedings of the IEEE/CVF Conference on Computer Vision and Pattern Recognition}, pp.\  2818--2829, 2023.

\bibitem[Clark et~al.(2022)Clark, de~Las~Casas, Guy, Mensch, Paganini, Hoffmann, Damoc, Hechtman, Cai, Borgeaud, et~al.]{clark2022unified}
Aidan Clark, Diego de~Las~Casas, Aurelia Guy, Arthur Mensch, Michela Paganini, Jordan Hoffmann, Bogdan Damoc, Blake Hechtman, Trevor Cai, Sebastian Borgeaud, et~al.
\newblock Unified scaling laws for routed language models.
\newblock In \emph{International conference on machine learning}, pp.\  4057--4086. PMLR, 2022.

\bibitem[Covert et~al.(2024)Covert, Ji, Hashimoto, and Zou]{covert2024scaling}
Ian Covert, Wenlong Ji, Tatsunori Hashimoto, and James Zou.
\newblock Scaling laws for the value of individual data points in machine learning.
\newblock \emph{arXiv preprint arXiv:2405.20456}, 2024.

\bibitem[Dettmers \& Zettlemoyer(2023)Dettmers and Zettlemoyer]{dettmers2023case}
Tim Dettmers and Luke Zettlemoyer.
\newblock The case for 4-bit precision: k-bit inference scaling laws.
\newblock In \emph{International Conference on Machine Learning}, pp.\  7750--7774. PMLR, 2023.

\bibitem[Dettmers et~al.(2022)Dettmers, Lewis, Belkada, and Zettlemoyer]{dettmers2022llm}
Tim Dettmers, Mike Lewis, Younes Belkada, and Luke Zettlemoyer.
\newblock Llm. int8 (): 8-bit matrix multiplication for transformers at scale.
\newblock \emph{arXiv preprint arXiv:2208.07339}, 2022.

\bibitem[Droppo \& Elibol(2021)Droppo and Elibol]{droppo2021scaling}
Jasha Droppo and Oguz Elibol.
\newblock Scaling laws for acoustic models.
\newblock \emph{arXiv preprint arXiv:2106.09488}, 2021.

\bibitem[Dubey et~al.(2024)Dubey, Jauhri, Pandey, Kadian, Al-Dahle, Letman, Mathur, Schelten, Yang, Fan, et~al.]{dubey2024llama}
Abhimanyu Dubey, Abhinav Jauhri, Abhinav Pandey, Abhishek Kadian, Ahmad Al-Dahle, Aiesha Letman, Akhil Mathur, Alan Schelten, Amy Yang, Angela Fan, et~al.
\newblock The llama 3 herd of models.
\newblock \emph{arXiv preprint arXiv:2407.21783}, 2024.

\bibitem[Fedus et~al.(2022)Fedus, Zoph, and Shazeer]{fedus2022switch}
William Fedus, Barret Zoph, and Noam Shazeer.
\newblock Switch transformers: Scaling to trillion parameter models with simple and efficient sparsity.
\newblock \emph{The Journal of Machine Learning Research}, 23\penalty0 (1):\penalty0 5232--5270, 2022.

\bibitem[Fernandes et~al.(2023)Fernandes, Ghorbani, Garcia, Freitag, and Firat]{fernandes2023scaling}
Patrick Fernandes, Behrooz Ghorbani, Xavier Garcia, Markus Freitag, and Orhan Firat.
\newblock Scaling laws for multilingual neural machine translation.
\newblock \emph{arXiv preprint arXiv:2302.09650}, 2023.

\bibitem[Filipovich et~al.(2022)Filipovich, Cappelli, Hesslow, and Launay]{filipovich2022scaling}
Matthew~J Filipovich, Alessandro Cappelli, Daniel Hesslow, and Julien Launay.
\newblock Scaling laws beyond backpropagation.
\newblock \emph{arXiv preprint arXiv:2210.14593}, 2022.

\bibitem[Frantar et~al.(2023)Frantar, Riquelme, Houlsby, Alistarh, and Evci]{frantar2023scaling}
Elias Frantar, Carlos Riquelme, Neil Houlsby, Dan Alistarh, and Utku Evci.
\newblock Scaling laws for sparsely-connected foundation models.
\newblock \emph{arXiv preprint arXiv:2309.08520}, 2023.

\bibitem[Gao et~al.(2020)Gao, Biderman, Black, Golding, Hoppe, Foster, Phang, He, Thite, Nabeshima, Presser, and Leahy]{gao2020pile800gbdatasetdiverse}
Leo Gao, Stella Biderman, Sid Black, Laurence Golding, Travis Hoppe, Charles Foster, Jason Phang, Horace He, Anish Thite, Noa Nabeshima, Shawn Presser, and Connor Leahy.
\newblock The pile: An 800gb dataset of diverse text for language modeling, 2020.
\newblock URL \url{https://arxiv.org/abs/2101.00027}.

\bibitem[Gao et~al.(2023)Gao, Schulman, and Hilton]{gao2023scaling}
Leo Gao, John Schulman, and Jacob Hilton.
\newblock Scaling laws for reward model overoptimization.
\newblock In \emph{International Conference on Machine Learning}, pp.\  10835--10866. PMLR, 2023.

\bibitem[Gao et~al.(2024)Gao, la~Tour, Tillman, Goh, Troll, Radford, Sutskever, Leike, and Wu]{gao2024scalingevaluatingsparseautoencoders}
Leo Gao, Tom~Dupré la~Tour, Henk Tillman, Gabriel Goh, Rajan Troll, Alec Radford, Ilya Sutskever, Jan Leike, and Jeffrey Wu.
\newblock Scaling and evaluating sparse autoencoders, 2024.
\newblock URL \url{https://arxiv.org/abs/2406.04093}.

\bibitem[Geiping et~al.(2022)Geiping, Goldblum, Somepalli, Shwartz-Ziv, Goldstein, and Wilson]{geiping2022much}
Jonas Geiping, Micah Goldblum, Gowthami Somepalli, Ravid Shwartz-Ziv, Tom Goldstein, and Andrew~Gordon Wilson.
\newblock How much data are augmentations worth? an investigation into scaling laws, invariance, and implicit regularization.
\newblock \emph{arXiv preprint arXiv:2210.06441}, 2022.

\bibitem[Ghorbani et~al.(2021)Ghorbani, Firat, Freitag, Bapna, Krikun, Garcia, Chelba, and Cherry]{ghorbani2021scaling}
Behrooz Ghorbani, Orhan Firat, Markus Freitag, Ankur Bapna, Maxim Krikun, Xavier Garcia, Ciprian Chelba, and Colin Cherry.
\newblock Scaling laws for neural machine translation.
\newblock \emph{arXiv preprint arXiv:2109.07740}, 2021.

\bibitem[Goldstein et~al.(2004)Goldstein, Morris, and Yen]{goldstein2004problems}
Michel~L Goldstein, Steven~A Morris, and Gary~G Yen.
\newblock Problems with fitting to the power-law distribution.
\newblock \emph{The European Physical Journal B-Condensed Matter and Complex Systems}, 41:\penalty0 255--258, 2004.

\bibitem[Gordon et~al.(2021)Gordon, Duh, and Kaplan]{gordon2021data}
Mitchell~A Gordon, Kevin Duh, and Jared Kaplan.
\newblock Data and parameter scaling laws for neural machine translation.
\newblock In \emph{Proceedings of the 2021 Conference on Empirical Methods in Natural Language Processing}, pp.\  5915--5922, 2021.

\bibitem[Goyal et~al.(2024)Goyal, Maini, Lipton, Raghunathan, and Kolter]{goyal2024scaling}
Sachin Goyal, Pratyush Maini, Zachary~C Lipton, Aditi Raghunathan, and J~Zico Kolter.
\newblock Scaling laws for data filtering--data curation cannot be compute agnostic.
\newblock In \emph{Proceedings of the IEEE/CVF Conference on Computer Vision and Pattern Recognition}, pp.\  22702--22711, 2024.

\bibitem[Gu \& Dao(2023)Gu and Dao]{gu2023mamba}
Albert Gu and Tri Dao.
\newblock Mamba: Linear-time sequence modeling with selective state spaces.
\newblock \emph{arXiv preprint arXiv:2312.00752}, 2023.

\bibitem[H{\"a}gele et~al.(2024)H{\"a}gele, Bakouch, Kosson, Allal, Von~Werra, and Jaggi]{hagele2024scaling}
Alexander H{\"a}gele, Elie Bakouch, Atli Kosson, Loubna~Ben Allal, Leandro Von~Werra, and Martin Jaggi.
\newblock Scaling laws and compute-optimal training beyond fixed training durations.
\newblock \emph{arXiv preprint arXiv:2405.18392}, 2024.

\bibitem[Hashimoto(2021)]{hashimoto2021model}
Tatsunori Hashimoto.
\newblock Model performance scaling with multiple data sources.
\newblock In \emph{International Conference on Machine Learning}, pp.\  4107--4116. PMLR, 2021.

\bibitem[Henighan et~al.(2020)Henighan, Kaplan, Katz, Chen, Hesse, Jackson, Jun, Brown, Dhariwal, Gray, et~al.]{henighan2020scaling}
Tom Henighan, Jared Kaplan, Mor Katz, Mark Chen, Christopher Hesse, Jacob Jackson, Heewoo Jun, Tom~B Brown, Prafulla Dhariwal, Scott Gray, et~al.
\newblock Scaling laws for autoregressive generative modeling.
\newblock \emph{arXiv preprint arXiv:2010.14701}, 2020.

\bibitem[Hernandez et~al.(2021)Hernandez, Kaplan, Henighan, and McCandlish]{hernandez2021scaling}
Danny Hernandez, Jared Kaplan, Tom Henighan, and Sam McCandlish.
\newblock Scaling laws for transfer.
\newblock \emph{arXiv preprint arXiv:2102.01293}, 2021.

\bibitem[Hernandez et~al.(2022)Hernandez, Brown, Conerly, DasSarma, Drain, El-Showk, Elhage, Hatfield-Dodds, Henighan, Hume, et~al.]{hernandez2022scaling}
Danny Hernandez, Tom Brown, Tom Conerly, Nova DasSarma, Dawn Drain, Sheer El-Showk, Nelson Elhage, Zac Hatfield-Dodds, Tom Henighan, Tristan Hume, et~al.
\newblock Scaling laws and interpretability of learning from repeated data.
\newblock \emph{arXiv preprint arXiv:2205.10487}, 2022.

\bibitem[Hestness et~al.(2017)Hestness, Narang, Ardalani, Diamos, Jun, Kianinejad, Patwary, Yang, and Zhou]{hestness2017deep}
Joel Hestness, Sharan Narang, Newsha Ardalani, Gregory Diamos, Heewoo Jun, Hassan Kianinejad, Md~Mostofa~Ali Patwary, Yang Yang, and Yanqi Zhou.
\newblock Deep learning scaling is predictable, empirically.
\newblock \emph{arXiv preprint arXiv:1712.00409}, 2017.

\bibitem[Hilton et~al.(2023)Hilton, Tang, and Schulman]{hilton2023scaling}
Jacob Hilton, Jie Tang, and John Schulman.
\newblock Scaling laws for single-agent reinforcement learning.
\newblock \emph{arXiv preprint arXiv:2301.13442}, 2023.

\bibitem[Hoffmann et~al.(2022)Hoffmann, Borgeaud, Mensch, Buchatskaya, Cai, Rutherford, Casas, Hendricks, Welbl, Clark, et~al.]{hoffmann2022training}
Jordan Hoffmann, Sebastian Borgeaud, Arthur Mensch, Elena Buchatskaya, Trevor Cai, Eliza Rutherford, Diego de~Las Casas, Lisa~Anne Hendricks, Johannes Welbl, Aidan Clark, et~al.
\newblock Training compute-optimal large language models.
\newblock \emph{arXiv preprint arXiv:2203.15556}, 2022.

\bibitem[Hu et~al.(2024)Hu, Tu, Han, He, Cui, Long, Zheng, Fang, Huang, Zhao, et~al.]{hu2024minicpm}
Shengding Hu, Yuge Tu, Xu~Han, Chaoqun He, Ganqu Cui, Xiang Long, Zhi Zheng, Yewei Fang, Yuxiang Huang, Weilin Zhao, et~al.
\newblock Minicpm: Unveiling the potential of small language models with scalable training strategies.
\newblock \emph{arXiv preprint arXiv:2404.06395}, 2024.

\bibitem[Huber(1992)]{huber1992robust}
Peter~J Huber.
\newblock Robust estimation of a location parameter.
\newblock In \emph{Breakthroughs in statistics: Methodology and distribution}, pp.\  492--518. Springer, 1992.

\bibitem[Ivgi et~al.(2022)Ivgi, Carmon, and Berant]{ivgi2022scaling}
Maor Ivgi, Yair Carmon, and Jonathan Berant.
\newblock Scaling laws under the microscope: Predicting transformer performance from small scale experiments.
\newblock \emph{arXiv preprint arXiv:2202.06387}, 2022.

\bibitem[Jones(2021)]{jones2021scaling}
Andy~L Jones.
\newblock Scaling scaling laws with board games.
\newblock \emph{arXiv preprint arXiv:2104.03113}, 2021.

\bibitem[Kaplan et~al.(2020)Kaplan, McCandlish, Henighan, Brown, Chess, Child, Gray, Radford, Wu, and Amodei]{kaplan2020scaling}
Jared Kaplan, Sam McCandlish, Tom Henighan, Tom~B. Brown, Benjamin Chess, Rewon Child, Scott Gray, Alec Radford, Jeffrey Wu, and Dario Amodei.
\newblock Scaling laws for neural language models.
\newblock 2020.

\bibitem[Kudo(2018)]{kudo2018sentencepiece}
T~Kudo.
\newblock Sentencepiece: A simple and language independent subword tokenizer and detokenizer for neural text processing.
\newblock \emph{arXiv preprint arXiv:1808.06226}, 2018.

\bibitem[Liu \& Nocedal(1989)Liu and Nocedal]{liu1989limited}
Dong~C Liu and Jorge Nocedal.
\newblock On the limited memory bfgs method for large scale optimization.
\newblock \emph{Mathematical programming}, 45\penalty0 (1):\penalty0 503--528, 1989.

\bibitem[McCandlish et~al.(2018)McCandlish, Kaplan, Amodei, and Team]{mccandlish2018empirical}
Sam McCandlish, Jared Kaplan, Dario Amodei, and OpenAI~Dota Team.
\newblock An empirical model of large-batch training.
\newblock \emph{arXiv preprint arXiv:1812.06162}, 2018.

\bibitem[Mikami et~al.()Mikami, Fukumizu, Murai, Suzuki, Kikuchi, Suzuki, Maeda, and Hayashi]{mikamiscaling}
Hiroaki Mikami, Kenji Fukumizu, Shogo Murai, Shuji Suzuki, Yuta Kikuchi, Taiji Suzuki, Shin-ichi Maeda, and Kohei Hayashi.
\newblock A scaling law for syn-to-real transfer: How much is your pre-training effective?

\bibitem[Muennighoff et~al.(2024)Muennighoff, Rush, Barak, Le~Scao, Tazi, Piktus, Pyysalo, Wolf, and Raffel]{muennighoff2024scaling}
Niklas Muennighoff, Alexander Rush, Boaz Barak, Teven Le~Scao, Nouamane Tazi, Aleksandra Piktus, Sampo Pyysalo, Thomas Wolf, and Colin~A Raffel.
\newblock Scaling data-constrained language models.
\newblock \emph{Advances in Neural Information Processing Systems}, 36, 2024.

\bibitem[Neumann \& Gros(2022)Neumann and Gros]{neumann2022scaling}
Oren Neumann and Claudius Gros.
\newblock Scaling laws for a multi-agent reinforcement learning model.
\newblock \emph{arXiv preprint arXiv:2210.00849}, 2022.

\bibitem[OpenAI(2023)]{openai2023gpt}
R~OpenAI.
\newblock Gpt-4 technical report.
\newblock \emph{arXiv}, pp.\  2303--08774, 2023.

\bibitem[Pearce \& Song(2024{\natexlab{a}})Pearce and Song]{pearce2024reconciling}
Tim Pearce and Jinyeop Song.
\newblock Reconciling kaplan and chinchilla scaling laws.
\newblock \emph{arXiv preprint arXiv:2406.12907}, 2024{\natexlab{a}}.

\bibitem[Pearce \& Song(2024{\natexlab{b}})Pearce and Song]{pearce2024reconcilingkaplanchinchillascaling}
Tim Pearce and Jinyeop Song.
\newblock Reconciling kaplan and chinchilla scaling laws, 2024{\natexlab{b}}.
\newblock URL \url{https://arxiv.org/abs/2406.12907}.

\bibitem[Penedo et~al.(2023)Penedo, Malartic, Hesslow, Cojocaru, Cappelli, Alobeidli, Pannier, Almazrouei, and Launay]{penedo2023refinedwebdatasetfalconllm}
Guilherme Penedo, Quentin Malartic, Daniel Hesslow, Ruxandra Cojocaru, Alessandro Cappelli, Hamza Alobeidli, Baptiste Pannier, Ebtesam Almazrouei, and Julien Launay.
\newblock The refinedweb dataset for falcon llm: Outperforming curated corpora with web data, and web data only, 2023.
\newblock URL \url{https://arxiv.org/abs/2306.01116}.

\bibitem[Penedo et~al.(2024)Penedo, Kydlíček, allal, Lozhkov, Mitchell, Raffel, Werra, and Wolf]{penedo2024finewebdatasetsdecantingweb}
Guilherme Penedo, Hynek Kydlíček, Loubna~Ben allal, Anton Lozhkov, Margaret Mitchell, Colin Raffel, Leandro~Von Werra, and Thomas Wolf.
\newblock The fineweb datasets: Decanting the web for the finest text data at scale, 2024.
\newblock URL \url{https://arxiv.org/abs/2406.17557}.

\bibitem[Poli et~al.(2024)Poli, Thomas, Nguyen, Ponnusamy, Deiseroth, Kersting, Suzuki, Hie, Ermon, R{\'e}, et~al.]{poli2024mechanistic}
Michael Poli, Armin~W Thomas, Eric Nguyen, Pragaash Ponnusamy, Bj{\"o}rn Deiseroth, Kristian Kersting, Taiji Suzuki, Brian Hie, Stefano Ermon, Christopher R{\'e}, et~al.
\newblock Mechanistic design and scaling of hybrid architectures.
\newblock \emph{arXiv preprint arXiv:2403.17844}, 2024.

\bibitem[Porian et~al.(2024)Porian, Wortsman, Jitsev, Schmidt, and Carmon]{porian2024resolving}
Tomer Porian, Mitchell Wortsman, Jenia Jitsev, Ludwig Schmidt, and Yair Carmon.
\newblock Resolving discrepancies in compute-optimal scaling of language models.
\newblock \emph{arXiv preprint arXiv:2406.19146}, 2024.

\bibitem[Prato et~al.(2021)Prato, Guiroy, Caballero, Rish, and Chandar]{prato2021scaling}
Gabriele Prato, Simon Guiroy, Ethan Caballero, Irina Rish, and Sarath Chandar.
\newblock Scaling laws for the few-shot adaptation of pre-trained image classifiers.
\newblock \emph{arXiv preprint arXiv:2110.06990}, 2021.

\bibitem[Radford et~al.(2019)Radford, Wu, Child, Luan, Amodei, Sutskever, et~al.]{radford2019language}
Alec Radford, Jeffrey Wu, Rewon Child, David Luan, Dario Amodei, Ilya Sutskever, et~al.
\newblock Language models are unsupervised multitask learners.
\newblock \emph{OpenAI blog}, 1\penalty0 (8):\penalty0 9, 2019.

\bibitem[Rae et~al.(2021)Rae, Borgeaud, Cai, Millican, Hoffmann, Song, Aslanides, Henderson, Ring, Young, et~al.]{rae2021scaling}
Jack~W Rae, Sebastian Borgeaud, Trevor Cai, Katie Millican, Jordan Hoffmann, Francis Song, John Aslanides, Sarah Henderson, Roman Ring, Susannah Young, et~al.
\newblock Scaling language models: Methods, analysis \& insights from training gopher.
\newblock \emph{arXiv preprint arXiv:2112.11446}, 2021.

\bibitem[Raffel et~al.(2020)Raffel, Shazeer, Roberts, Lee, Narang, Matena, Zhou, Li, and Liu]{raffel2020exploring}
Colin Raffel, Noam Shazeer, Adam Roberts, Katherine Lee, Sharan Narang, Michael Matena, Yanqi Zhou, Wei Li, and Peter~J Liu.
\newblock Exploring the limits of transfer learning with a unified text-to-text transformer.
\newblock \emph{Journal of machine learning research}, 21\penalty0 (140):\penalty0 1--67, 2020.

\bibitem[Reid et~al.(2024)Reid, Savinov, Teplyashin, Lepikhin, Lillicrap, Alayrac, Soricut, Lazaridou, Firat, Schrittwieser, et~al.]{reid2024gemini}
Machel Reid, Nikolay Savinov, Denis Teplyashin, Dmitry Lepikhin, Timothy Lillicrap, Jean-baptiste Alayrac, Radu Soricut, Angeliki Lazaridou, Orhan Firat, Julian Schrittwieser, et~al.
\newblock Gemini 1.5: Unlocking multimodal understanding across millions of tokens of context.
\newblock \emph{arXiv preprint arXiv:2403.05530}, 2024.

\bibitem[Rosenfeld et~al.(2019)Rosenfeld, Rosenfeld, Belinkov, and Shavit]{rosenfeld2019constructive}
Jonathan~S Rosenfeld, Amir Rosenfeld, Yonatan Belinkov, and Nir Shavit.
\newblock A constructive prediction of the generalization error across scales.
\newblock \emph{arXiv preprint arXiv:1909.12673}, 2019.

\bibitem[Ruan et~al.(2024)Ruan, Maddison, and Hashimoto]{ruan2024observational}
Yangjun Ruan, Chris~J Maddison, and Tatsunori Hashimoto.
\newblock Observational scaling laws and the predictability of language model performance.
\newblock \emph{arXiv preprint arXiv:2405.10938}, 2024.

\bibitem[Sardana \& Frankle(2023)Sardana and Frankle]{sardana2023beyond}
Nikhil Sardana and Jonathan Frankle.
\newblock Beyond chinchilla-optimal: Accounting for inference in language model scaling laws.
\newblock \emph{arXiv preprint arXiv:2401.00448}, 2023.

\bibitem[Schaeffer et~al.(2023)Schaeffer, Miranda, and Koyejo]{schaeffer2023emergent}
Rylan Schaeffer, Brando Miranda, and Sanmi Koyejo.
\newblock Are emergent abilities of large language models a mirage?
\newblock \emph{arXiv preprint arXiv:2304.15004}, 2023.

\bibitem[Shazeer(2020)]{shazeer2020gluvariantsimprovetransformer}
Noam Shazeer.
\newblock Glu variants improve transformer, 2020.
\newblock URL \url{https://arxiv.org/abs/2002.05202}.

\bibitem[Shazeer et~al.(2017)Shazeer, Mirhoseini, Maziarz, Davis, Le, Hinton, and Dean]{shazeer2017outrageously}
Noam Shazeer, Azalia Mirhoseini, Krzysztof Maziarz, Andy Davis, Quoc Le, Geoffrey Hinton, and Jeff Dean.
\newblock Outrageously large neural networks: The sparsely-gated mixture-of-experts layer.
\newblock \emph{arXiv preprint arXiv:1701.06538}, 2017.

\bibitem[Shin et~al.(2023)Shin, Kwak, Kim, Ramstr{\"o}m, Jeong, Ha, and Kim]{shin2023scaling}
Kyuyong Shin, Hanock Kwak, Su~Young Kim, Max~Nihl{\'e}n Ramstr{\"o}m, Jisu Jeong, Jung-Woo Ha, and Kyung-Min Kim.
\newblock Scaling law for recommendation models: Towards general-purpose user representations.
\newblock In \emph{Proceedings of the AAAI Conference on Artificial Intelligence}, volume~37, pp.\  4596--4604, 2023.

\bibitem[Sorscher et~al.(2022)Sorscher, Geirhos, Shekhar, Ganguli, and Morcos]{sorscher2022beyond}
Ben Sorscher, Robert Geirhos, Shashank Shekhar, Surya Ganguli, and Ari Morcos.
\newblock Beyond neural scaling laws: beating power law scaling via data pruning.
\newblock \emph{Advances in Neural Information Processing Systems}, 35:\penalty0 19523--19536, 2022.

\bibitem[Su et~al.(2024)Su, Ahmed, Lu, Pan, Bo, and Liu]{su2024roformer}
Jianlin Su, Murtadha Ahmed, Yu~Lu, Shengfeng Pan, Wen Bo, and Yunfeng Liu.
\newblock Roformer: Enhanced transformer with rotary position embedding.
\newblock \emph{Neurocomputing}, 568:\penalty0 127063, 2024.

\bibitem[Tao et~al.(2024)Tao, Liu, Dou, Muennighoff, Wan, Luo, Lin, and Wong]{tao2024scaling}
Chaofan Tao, Qian Liu, Longxu Dou, Niklas Muennighoff, Zhongwei Wan, Ping Luo, Min Lin, and Ngai Wong.
\newblock Scaling laws with vocabulary: Larger models deserve larger vocabularies.
\newblock \emph{arXiv preprint arXiv:2407.13623}, 2024.

\bibitem[Tay et~al.(2022)Tay, Dehghani, Abnar, Chung, Fedus, Rao, Narang, Tran, Yogatama, and Metzler]{tay2022scaling}
Yi~Tay, Mostafa Dehghani, Samira Abnar, Hyung~Won Chung, William Fedus, Jinfeng Rao, Sharan Narang, Vinh~Q Tran, Dani Yogatama, and Donald Metzler.
\newblock Scaling laws vs model architectures: How does inductive bias influence scaling?
\newblock \emph{arXiv preprint arXiv:2207.10551}, 2022.

\bibitem[Vaswani(2017)]{vaswani2017attention}
A~Vaswani.
\newblock Attention is all you need.
\newblock \emph{Advances in Neural Information Processing Systems}, 2017.

\bibitem[Wei et~al.(2022)Wei, Tay, Bommasani, Raffel, Zoph, Borgeaud, Yogatama, Bosma, Zhou, Metzler, et~al.]{wei2022emergent}
Jason Wei, Yi~Tay, Rishi Bommasani, Colin Raffel, Barret Zoph, Sebastian Borgeaud, Dani Yogatama, Maarten Bosma, Denny Zhou, Donald Metzler, et~al.
\newblock Emergent abilities of large language models.
\newblock \emph{arXiv preprint arXiv:2206.07682}, 2022.

\bibitem[Yang et~al.(2022)Yang, Hu, Babuschkin, Sidor, Liu, Farhi, Ryder, Pachocki, Chen, and Gao]{yang2022tensor}
Greg Yang, Edward~J Hu, Igor Babuschkin, Szymon Sidor, Xiaodong Liu, David Farhi, Nick Ryder, Jakub Pachocki, Weizhu Chen, and Jianfeng Gao.
\newblock Tensor programs v: Tuning large neural networks via zero-shot hyperparameter transfer.
\newblock \emph{arXiv preprint arXiv:2203.03466}, 2022.

\bibitem[Zhai et~al.(2022)Zhai, Kolesnikov, Houlsby, and Beyer]{zhai2022scaling}
Xiaohua Zhai, Alexander Kolesnikov, Neil Houlsby, and Lucas Beyer.
\newblock Scaling vision transformers.
\newblock In \emph{Proceedings of the IEEE/CVF conference on computer vision and pattern recognition}, pp.\  12104--12113, 2022.

\bibitem[Zhang et~al.(2022)Zhang, Ghorbani, Bapna, Cheng, Garcia, Shen, and Firat]{zhang2022examining}
Biao Zhang, Behrooz Ghorbani, Ankur Bapna, Yong Cheng, Xavier Garcia, Jonathan Shen, and Orhan Firat.
\newblock Examining scaling and transfer of language model architectures for machine translation.
\newblock In \emph{International Conference on Machine Learning}, pp.\  26176--26192. PMLR, 2022.

\bibitem[Zhu \& Gupta(2017)Zhu and Gupta]{zhu2017prune}
Michael Zhu and Suyog Gupta.
\newblock To prune, or not to prune: exploring the efficacy of pruning for model compression.
\newblock \emph{arXiv preprint arXiv:1710.01878}, 2017.

\end{thebibliography}
\bibliographystyle{iclr2024_conference}

\appendix

\newpage


\section{Our model training (\S\ref{sec:own-repl})}\label{app:our_models}

We train a variety of Transformer LMs, ranging in size from 12 million to 1 billion parameters, on FineWeb \citep{penedo2024finewebdatasetsdecantingweb}, tokenized with the GPT-NeoX-20B tokenizer \citep{black2022gptneox20bopensourceautoregressivelanguage}, which is a BPE tokenizer with a vocabulary size of 50257, trained on the Pile \citep{gao2020pile800gbdatasetdiverse}. All models were trained on a combination of NVIDIA GeForce RTX 2080 Ti, Quadro RTX 6000, NVIDIA A40, and NVIDIA L40 GPUs. These transformers follow the standard architecture, and use pre-layer RMSNorm \citep{kudo2018sentencepiece}, the SwiGLU activation function \citep{shazeer2020gluvariantsimprovetransformer}, and rotary positional embeddings \citep{su2024roformer}. We use a batch size of 512 with a sequence length of 2048. The learning rate is warmed up linearly over 50 steps to the peak learning rate and then follows a cosine decay to 10\% of the peak. We use an Adam optimizer with $\beta_1 =
0.9, \beta_2 = 0.95$. We sweep over learning rates and data budgets at each model scale. For our evaluation metric, we use perplexity on a validation set of C4 \citep{raffel2020exploring}. See Table~\ref{tab:hparams} and Table~\ref{tab:arch} for additional architecture details and hyperparameters, including data budget and learning rate.

\begin{table}[H]
    \centering
\begin{tabular}{lc}
\toprule
 Hyperparameter & Value \\ \midrule
Vocabulary size & 50K \\
Batch Size &    512	\\
Sequence Length & 2048	\\
Attention Head Size & 64 \\
Learning Rate & Swept $2^{\{0, 1, 2, 3\}} \cdot 10^{\{-3, -4 \}}$ \\
Feedforward Dimension & 4 x hidden dimension \\
LR Schedule & Cosine Decay \\
LR Warmup & 50 steps \\
End LR & 0.1 x Peak LR \\
Optimizer & Adam $(\beta_1, \beta_2 = 0.9, 0.95)$ \\
\bottomrule
\end{tabular}%
    \caption{\small{Hyperparameter details for models trained in \S\ref{sec:own-repl}}}
\label{tab:hparams} 
\end{table}
\begin{table}[H]
    \centering
    \resizebox{\columnwidth}{!}{%
\begin{tabular}{l|c|c|c|c}
\toprule
Model size & \# layers & Hidden Dim & \# Attn Heads & \# Steps \\ \midrule
12M & 5	    & 448	& 7	 & \{100, 200, 250, 360, 500, 750, \\
& & & &  1,000, 4,000\} \\ \midrule
17M & 7	    & 448	& 7	& \{100, 200, 250, 500, 750, 1,000,\\
& & & &   1,250, 1,500\} \\ \midrule
25M & 8	& 512	& 8		& \{250, 360, 500, 750, 1,000, 1,500, \\
& & & &  2,000, 8,000, 16,000\} \\ \midrule
35M & 9	    & 576	& 9		& \{200, 250, 360, 500, 750, 1,000, \\
& & & &  1,250, 1,500, 4,000, 16,000\} \\ \midrule
50M & 10	& 640	& 10 	& \{250, 500, 750, 1,000, 1,250, 1,500, \\
& & & &  1,800, 2,000, 2,500, 4,000, 16,000\} \\ \midrule
70M & 12	& 704	& 11 	& \{250, 500, 750, 1,000, 1,250, 1,500, \\
& & & &  2,000, 2,500, 3,000, 5,000\} \\ \midrule
100M & 14	& 768	& 12 	& \{250, 500, 1,000, 1,500, 2,000, \\
& & & &  2,500, 3,000, 4,000, 6,000, 12,000\} \\ \midrule
200M & 18	& 960	& 15 	& \{500, 1,500, 6,000, 9,000\} \\ \midrule
300M & 19	& 1152	& 18 	& \{4,000\} \\ \midrule
400M & 20	& 1280	& 20	& \{3,000, 6,000\} \\ \midrule
1B & 26	    & 1792	& 28	& \{20,000\} \\ \bottomrule
\end{tabular}%
}
    \caption{\small{Architecture and Data budget details for models trained in \S\ref{sec:own-repl}}}
\label{tab:arch} 
\end{table}

\section{Full Checklist} \label{sec:app_checklist}

We define each category as follows:

\begin{itemize}
    \item {
    \textbf{Scaling Law Hypothesis:} This specifies the form of the scaling law, that of the variables and parameters, and the relation between each.}
    \item {\textbf{Training Setup:} This specifies the exact training setup of each of the models trained to test the scaling law hypothesis.}
    \item {\textbf{Data Collection:} Evaluating various checkpoints of our trained models to collect data points that will be used to fit a scaling law in the next stage.}
    \item {\textbf{Fitting Algorithm:} Using the data points collected in the previous stage to optimize the scaling law hypothesis.}
\end{itemize}

\fbox{\begin{minipage}{38em}

\subsubsection*{Scaling Law Reproducilibility Checklist}

\small

\begin{minipage}[t]{0.9\textwidth}
\raggedright
\paragraph{Scaling Law Hypothesis (\S\ref{sec:power-law-form})}

\begin{itemize}[leftmargin=*]
    \item What is the form of the power law?
    \item What are the variables related by (included in) the power law?
    \item What are the parameters to fit?
    \item On what principles is this form derived?
    \item Does this form make assumptions about how the variables are related?
\end{itemize}

\paragraph{Training Setup (\S\ref{sec:model_training})}
\begin{itemize}[leftmargin=*]
    \item How many models are trained?
    \item At which sizes?
    \item On how much data each? On what data? Is any data repeated within the training for a model?
    \item How are model size, dataset size, and compute budget size counted? For example, how are parameters of the model counted? Are any parameters excluded (e.g., embedding layers)? How is compute cost calculated?
    \item Are code/code snippets provided for calculating these variables if applicable?
    \item How are hyperparameters chosen (e.g., optimizer, learning rate schedule, batch size)? Do they change with scale?
    \item What other settings must be decided (e.g., model width vs. depth)? Do they change with scale?
    \item Is the training code open source?
\end{itemize}

\raggedright

\paragraph{Data Collection(\S\ref{sec:data})}
\begin{itemize}[leftmargin=*]
    \item Are the model checkpoints provided openly?
    \item How many checkpoints per model are evaluated to fit each scaling law? Which ones, if so?
    \item What evaluation metric is used? On what dataset?
    \item Are the raw evaluation metrics modified? Some examples include loss interpolation, centering around a mean or scaling logarithmically.
    \item If the above is done, is code for modifying the metric provided? 
\end{itemize}

\paragraph{Fitting Algorithm (\S\ref{sec:opt})}
\begin{itemize}[leftmargin=*]
    \item What objective (loss) is used?
    \item What algorithm is used to fit the equation?
    \item What hyperparameters are used for this algorithm?
    \item How is this algorithm initialized?
    \item Are all datapoints collected used to fit the equations? For example, are any outliers dropped? Are portions of the datapoints used to fit different equations?
    \item How is the correctness of the scaling law considered? Extrapolation, Confidence Intervals, Goodness of Fit?
\end{itemize}

\end{minipage}


\end{minipage}}

\pagebreak

\section{Full Sheet}\label{app:full-details}

We provide an overview of all the papers surveyed in Tables \ref{tab:full-basic},\ref{tab:full-powerlaw}, \ref{tab:full-setup}, \ref{tab:full-eval} and \ref{tab:full-opt}.


\begin{table}[!htp]
\centering
\resizebox{\textwidth}{!}{%
\begin{tabular}{lllllllll}
\toprule
Paper & Domain & Training Code? & Analysis Code? & Checkpoints? & Metric Scores? \\
\midrule
\cite{rosenfeld2019constructive} & Vision, LM & N & N & N & N \\
\cite{mikamiscaling} & Vision & N & Y & Y & Y \\
\cite{schaeffer2023emergent} & LM & N & N & N & N \\
\cite{sardana2023beyond} & LM & N & N & N & N \\
\cite{sorscher2022beyond} & Vision & N & N & N & Y \\
\cite{caballero2022broken} & LM & N & Y & N & Y \\
\cite{besiroglu2024chinchilla} & LM &  & Y & N & Y \\
\cite{gordon2021data} & NMT & Y & Y & Y & Y \\
\cite{bansal2022data} & NMT & N & N & N & N \\
\cite{hestness2017deep} & NMT, LM, Vision, Speech & N & N & N & N \\
\cite{bi2024deepseek} & LM & N & N & N & N \\
\cite{bahri2021explaining} & Vision & N & N & N & N \\
\cite{geiping2022much} & Vision & Y & Y & N & N \\
\cite{poli2024mechanistic} & LM & N & N & N & N \\
\cite{hu2024minicpm} & LM & Y & N & N & N \\
\cite{hashimoto2021model} & NLP & N & N & N & N \\
\cite{ruan2024observational} & LM & Y & Y & N & Y \\
\cite{anil2023palm} & LM & N & N & N & N \\
\cite{pearce2024reconciling} & LM & N & Y & N & N \\
\cite{cherti2023reproducible} & VLM & Y & Y & Y & Y \\
\cite{porian2024resolving} & LM & Y & Y & N & Y \\
\cite{alabdulmohsin2022revisiting} & LM, Vision & N & Y & Y & Y \\
\cite{gao2024scalingevaluatingsparseautoencoders} & NLP & Y & Y & Y & N \\
\cite{muennighoff2024scaling} & LM & Y & Y & Y & N \\
\cite{rae2021scaling} & LM & N & N & N & N \\
\cite{shin2023scaling} & RecSys & N & N & N & N \\
\cite{hernandez2022scaling} & LM & N & N & N & N \\
\cite{filipovich2022scaling} & LM & N & N & N & N \\
\cite{neumann2022scaling} & RL & Y & Y* & Y & N \\
\cite{droppo2021scaling} & Speech & N & N & N & N \\
\cite{henighan2020scaling} & LM, Vision, Video, VLM & N & N & N & N \\
\cite{goyal2024scaling} & LM, Vision, VLM & N & Y & N & Y \\
\cite{aghajanyan2023scaling} & Multimodal LM & N & N & N & N \\
\cite{kaplan2020scaling} & LM & N & N & N & N \\
\cite{ghorbani2021scaling} & NMT & N & Y & N & N \\
\cite{gao2023scaling} & RL/LM & N & N & N & N \\
\cite{hilton2023scaling} & RL & N & N & N & N \\
\cite{frantar2023scaling} & LM, Vision & N & N & N & N \\
\cite{prato2021scaling} & Vision & Y* & Y & Y & Y \\
\cite{covert2024scaling} & LM & Y & Y & N & N \\
\cite{hernandez2021scaling} & LM & N & N & N & N \\
\cite{ivgi2022scaling} & NLP & N & N & N & N \\
\cite{tay2022scaling} & LM & N & N & N & N \\
\cite{tao2024scaling} & LM & N & Y & N & Y \\
\cite{jones2021scaling} & RL & Y & Y & N & Y \\
\cite{zhai2022scaling} & Vision & Y & N & N & N \\
\cite{dettmers2023case} & LM & N & N & N & N \\
\cite{dubey2024llama} & LM & N & N & N & N \\
\cite{hoffmann2022training} & LM & N & N & N & N \\
\cite{ardalani2022understanding} & RecSys & N & N & N & N \\
\cite{clark2022unified} & LM & N & Y & N & Y \\
\bottomrule
\end{tabular}
}

\caption{Details on domain of experiments and availability of code by category for each paper surveyed.}
\label{tab:full-basic}
\end{table}

\begin{table}[]
\centering
\resizebox{\textwidth}{!}{%
\begin{tabular}{lllll}
\toprule
Paper & Power Law Form & Purpose Of Power Law (E.G., Performance Prediction, Optimal Ratio) & \# Power Law Parameters & \# Of Scaling Laws \\
\midrule
\cite{rosenfeld2019constructive} & $\tilde{\epsilon}(m, n)=a n^{-\alpha}+b m^{-\beta}+c_{\infty}$ & Performance Prediction & 5-6 & 8 \\
\cite{mikamiscaling} & $L(n, s)=\delta\left(\gamma+n^{-\alpha}\right) s^{-\beta}$ & Performance Prediction & 4 & 3 \\
\cite{schaeffer2023emergent} & None & N/A & NA & NA \\
\cite{sardana2023beyond} & $L(N, D) = E + \frac{A}{N^\alpha} + \frac{B}{D^\beta} $; $N^*\left(\ell, D_{\text {inf }}\right), D_{\text {tr }}^*\left(\ell, D_{\text {inf }}\right)={\arg \min } _{N, D_{\mathrm{tr}} \mid L\left(N, D_{\mathrm{tr}}\right)=\ell}$ & Performance Prediction & 5 & 4 \\
\cite{sorscher2022beyond} & $c \cdot \alpha^{-\beta} ,  c \cdot \exp (-b \alpha)$ & Performance Prediction & 2 & 34 \\
\cite{caballero2022broken} & $y=a+\left(b x^{-c_0}\right) \prod_{i=1}^n\left(1+\left(\frac{x}{d_i}\right)^{1 / f_i}\right)^{-c_i * f_i}$ & Performance Prediction & 5+ & 100+ \\
\cite{besiroglu2024chinchilla} & $L(N, D) = E + \frac{A}{N^\alpha} + \frac{B}{D^\beta} $ & Performance Prediction & 5 & 1 \\
\cite{gordon2021data} & $L(N, D) = \left[ \left( \frac{N}{N_c}\right)^{\frac{\alpha_N}{\alpha_D}} + \frac{D}{D_c} \right]^{\alpha_D} $ & Performance Prediction & 4 & 3 \\
\cite{bansal2022data} & $L(D)=\alpha\left(D^{-1}+C\right)^p$ & Performance Prediction & 3 & 20 \\
\cite{hestness2017deep} & $\varepsilon(m) \sim \alpha m^{\beta_g}+\gamma$ & Performance Prediction & 3 & 17 \\
\cite{bi2024deepseek} & $\begin{aligned} M_{\mathrm{opt}} & =M_{\mathrm{base}} \cdot C^a \\ D_{\mathrm{opt}} & =D_{\mathrm{base}} \cdot C^b\end{aligned}$, $\begin{aligned} & \eta_{\mathrm{opt}}=0.3118 \cdot C^{-0.1250} \\ & B_{\mathrm{opt}}=0.2920 \cdot C^{0.3271}\end{aligned}$ & Optimal Ratio, Performance Prediction & 2 & 5 \\
\cite{bahri2021explaining} & $L(D) \propto D^{-\alpha_K}, \quad L(P) \propto P^{-\alpha_K}$ & Performance Prediction & 2 & 35 \\
\cite{geiping2022much} & $f(x)=a x^{-c}+b$, $v_{\text {Effective Extra Samples from Augmentations }}(x)=f_{\text {ref }}^{-1}\left(f_{\text {aug }}(x)\right)-x$ & Performance Prediction & 3 & ~50 \\
\cite{poli2024mechanistic} & $\log N^* \propto a \log C$ and $\log D^* \propto b \log C$ & Performance Prediction & 2 &  \\
\cite{hu2024minicpm} & $L(N, D)=C_N N^{-\alpha}+C_D D^{-\beta}+L_0$ & Performance Prediction & 5 & 6 \\
\cite{hashimoto2021model} & $\min _{\lambda, \alpha} \mathbb{E}_{\hat{q}, \hat{n}}\left[\left(\log (R(\hat{n}, \hat{q})-\epsilon)-\alpha \log (\hat{n})+\log \left(C_\lambda(\hat{q})\right)\right)^2\right]$ $R(\hat{n}, \hat{q})=\mathbb{E}\left[\ell\left(\hat{\theta}\left(p_{\hat{n}, \hat{q}}\right) ; x, y\right)\right]$ & Performance Prediction & 2+n(data mixes) & 4 \\
\cite{ruan2024observational} & $E_m \approx h \sigma\left(\beta^{\top} S_m+\alpha\right)$ & Performance Prediction & 3 &  \\
\cite{anil2023palm} & $N^{\star}(C) \approx N_0^{\star} \cdot C^a$ & Performance Prediction & 2 & 1 \\
\cite{pearce2024reconciling} & $N^*_{{\setminus E}} = b C_{{\setminus E}}^m$ $L = bC^m$ & Optimal Ratio, Performance Prediction & 2 & 1 \\
\cite{cherti2023reproducible} & $E=\beta C^{\alpha}$ & Performance Prediction & 2 & 8 \\
\cite{porian2024resolving} & $N^{\star}(C) \approx N_0^{\star} \cdot C^a$ & Optimal Ratio & 2 & 6 \\
\cite{alabdulmohsin2022revisiting} & $\varepsilon_x=\beta x^c$; $\varepsilon_x - \varepsilon_\infty=\beta x^c$; $\varepsilon_x=\beta (x^{-1} + \gamma)^{-c}$;   $\varepsilon_x=\gamma(x)(1+\gamma(x))^{-1} \varepsilon_0+(1+\gamma(x))^{-1} \varepsilon_{\infty}$ & Performance Prediction & 2-4 & ~600 \\
\cite{gao2024scalingevaluatingsparseautoencoders} & $L(n, k)=\exp \left(\alpha+\beta_k \log (k)+\beta_n \log (n)+\gamma \log (k) \log (n)\right)+\exp (\zeta+\eta \log (k))$ & Performance Prediction & 2-6 & 1 \\
\cite{muennighoff2024scaling} & $L\left(U_N, U_D, R_N, R_D\right)=\frac{A}{\left(U_N+U_N R_N^*\left(1-e^{\frac{-R_N}{R_N^*}}\right)\right)^\alpha}+\frac{B}{\left(U_D+U_D R_D^*\left(1-e^{\frac{-R_D}{R_D^*}}\right)\right)^\beta}+E$ & Performance Prediction & 2 (+4) & 1 \\
\cite{rae2021scaling} & None & Performance Prediction & N/A & N/A \\
\cite{shin2023scaling} & None & Scaling trend & NA & NA \\
\cite{hernandez2022scaling} & $E=k * N^\alpha$ & Optimal Ratio & 2 & 1 \\
\cite{filipovich2022scaling} & $\mathcal{L}(C)=\left(C_c C\right)^{\alpha_C}$ & Performance Prediction & 2 & 3 \\
\cite{neumann2022scaling} & $N_{\text {opt }}(C)=\left(\frac{C}{C_0}\right)^{\alpha_C^{o p t}}$, $E_i=\frac{1}{1+\left(N_j / N_i\right)^{\alpha_N}}$ & Performance Prediction & 2 & 3 * 2 \\
\cite{droppo2021scaling} & $L(N, D)=\left[\left(L_{\infty}\right)^{\frac{1}{\alpha}}+\left(\frac{N_C}{N}\right)^{\frac{\alpha_N}{\alpha}}+\left(\frac{D_C}{D}\right)^{\frac{\alpha_D}{\alpha}}\right]^\alpha$ & Performance Prediction & 6 & 3 \\
\cite{henighan2020scaling} & $L(x)=L_{\infty}+\left(\frac{x_0}{x}\right)^{\alpha_x}$ & Performance Prediction & 3 & 36 \\
\cite{goyal2024scaling} & $y_k=a \cdot n_1^{b_1} \prod_{j=2}^k\left(\frac{n_j}{n_{j-1}}\right)^{b_j}+d$ & Performance Prediction & 2+ 2*n(data mixes) & 1 \\
\cite{aghajanyan2023scaling} & $L(N, D_j)=E_j + \frac{A_j}{N^{\alpha_j}} + \frac{B_j}{|D_j|^{\beta_j}}$, $L(N, D_i, D_j) = [\frac{L(N, D_i) + L(N, D_j)}{2}] - C_{i,j} + \frac{A_{i,j}}{N^{\alpha_{i,j}}} + \frac{B_{i,j}}{|D_i|+|D_j|^{\beta_{i,j}}}$ & Performance Prediction & 5 & 14 \\
\cite{kaplan2020scaling} & $L(N, D) = \left[ \left( \frac{N}{N_c}\right)^{\frac{\alpha_N}{\alpha_D}} + \frac{D}{D_c} \right]^{\alpha_D}$     & Performance Prediction & 4 & ~7 \\
\cite{ghorbani2021scaling} & $\mathrm{BLEU}=c_B L^{-p_B}$, $\hat{L}_{o p t}(B)=\alpha^* B^{-\left(p_d+p_e\right)}+L_{\infty}, \quad \alpha^* \equiv \alpha\left(\frac{\bar{N}_e\left(p_e+p_d\right)}{p_e}\right)^{p_e}\left(\frac{\bar{N}_d\left(p_e+p_d\right)}{p_d}\right)^{p_d}$ & Optimal Ratio, Performance Prediction & 6 & ~8 \\
\cite{gao2023scaling} & $\begin{aligned} & R_{\mathrm{bo} n}(d)=d\left(\alpha_{\mathrm{bo} n}-\beta_{\mathrm{bo} n} d\right), \\ & R_{\mathrm{RL}}(d)=d\left(\alpha_{\mathrm{RL}}-\beta_{\mathrm{RL}} \log d\right)\end{aligned}$ & Performance Prediction & 2 & 2 \\
\cite{hilton2023scaling} & $I^{-\beta}=\left(\frac{N_c}{N}\right)^{\alpha_N}+\left(\frac{E_c}{E}\right)^{\alpha_E}$ & Optimal Ratio, Performance Prediction & 5 & 3 \\
\cite{frantar2023scaling} & $L(S, N, D)=\left(a_S(1-S)^{b_S}+c_S\right) \cdot\left(\frac{1}{N}\right)^{b_N}+\left(\frac{a_D}{D}\right)^{b_D}+c$ & Optimal Ratio, Performance Prediction & 7 & 2 \\
\cite{prato2021scaling} & $\begin{aligned} & \operatorname{Err}(N)=\operatorname{Err}_{\infty}+k N^\alpha, \\ & \operatorname{Err}(C)=\operatorname{Err}_{\infty}+k C^\alpha,\end{aligned}$ & Performance Prediction & 3 & 12 \\
\cite{covert2024scaling} & $\log \left|\psi_k(z)\right| \approx \log |c(z)|-\alpha(z) \log (k)$ & Performance Prediction & 2 & Many \\
\cite{hernandez2021scaling} & $L \approx\left[\left(\frac{N_C}{N}\right)^{\frac{\alpha_N}{\alpha_D}}+\frac{D_C}{k\left(D_F\right)^\alpha(N)^\beta}\right]^{\alpha_D}$ & Performance Prediction & 3 & 1 \\
\cite{ivgi2022scaling} & NS & Performance Prediction & NA & NA \\
\cite{tay2022scaling} & None & Scaling trend & NA & NA \\
\cite{tao2024scaling} & $N_{\mathrm{v}}^{\mathrm{opt}}=N_{\mathrm{v}}^0 *\left(\frac{N_{\mathrm{nv}}}{N_{\mathrm{nv}}^0}\right)^\gamma$, $\mathcal{L}_u=-E+\frac{A_1}{N_{\mathrm{nv}}^{\alpha_1}}+\frac{A_2}{N_{\mathrm{v}}^{\alpha_2}}+\frac{B}{D^\beta}$ & Optimal Ratio, Performance Prediction & 7 & 2 \\
\cite{jones2021scaling} & $\begin{aligned} \text { plateau } & =m_{\text {boardsize }}^{\text {plateau }} \cdot \text { boardsize }+c^{\text {plateau }} \\ \text { incline } & =m_{\text {boardsize }}^{\text {incline }} \cdot \text { boardsize }+m_{\text {flops }}^{\text {incline }} \cdot \log \text { flop }+c^{\text {incline }} \\ \text { elo } & =\text { incline.clamp }(\text { plateau }, 0)\end{aligned}$ & Performance Prediction & 5 & 1 \\
\cite{zhai2022scaling} & $E=\alpha+\beta(C+\gamma)^{-\mu}$ & Performance Prediction & 4 & 3 \\
\cite{dettmers2023case} & None & Scaling trend & NA & NA \\
\cite{dubey2024llama} & $N^{\star}(C)=A C^\alpha$. & Optimal Ratio  & 2 & 2 \\
\cite{hoffmann2022training} & A3: $L(N, D) = E + \frac{A}{N^\alpha} + \frac{B}{D^\beta} $ & Optimal Ratio, Performance Prediction & 5 & 3 \\
\cite{ardalani2022understanding} & $\left(\alpha x^{-\beta}+\gamma\right)$ & Performance Prediction & 3 & 3 \\
\cite{clark2022unified} & $\log L(N, E) \triangleq+a \log N+b \log E+c \log N \log E+d$ & Performance Prediction & 4 & 3 \\
\bottomrule
\end{tabular}
}

\caption{Details on power law for each paper surveyed.}
\label{tab:full-powerlaw}
\end{table}

\begin{table}[]
\centering
\resizebox{\textwidth}{!}{%
\begin{tabular}{llllllll}
\toprule
Paper & Training Runs / Law & Max. Training Flops & Max. Training Params & Max. Training Data & Data Described? & Hyperparameters Described? & How Are Model Params Counted \\
&&&&&&& (E.G., W/ Or W/Out Embeddings) \\
\midrule
\cite{rosenfeld2019constructive} & 42-49 &  & 0.7M-70M & 100M words / 1.2M images & Y & Y & Non-embedding \\
\cite{mikamiscaling} & 7 &  & ResNet-101 & 64k-1.28M images & Y & Y & NA \\
\cite{schaeffer2023emergent} & 4 &  & $10^{11}$ & NA & Y & NA & Non-embedding \\
\cite{sardana2023beyond} & 47 &  & 150M-6B & 1.5B-1.25T tokens & N & Y & NA \\
\cite{sorscher2022beyond} & ~60 &  & 86M (ViT) & 200 epochs & Y & Y & NA \\
\cite{caballero2022broken} & 3-40 &  & NS & NS & N & N & NS \\
\cite{besiroglu2024chinchilla} & NA & NA & NA & NA & Y & NA & Non-embedding \\
\cite{gordon2021data} & 45-55 &  & 56M & 28.3M-51.1M examples & Y & Y & Non-embedding \\
\cite{bansal2022data} & 10 &  & 170M-800M & 500K-512M sentences (28B tokens) & Y & Y & NS \\
\cite{hestness2017deep} & ~9 &  & upto 193M  & $2^{19}-2^{28}$ tokens, upto $2^9$ images, $2k$ audio hours & Y & Y & NS \\
\cite{bi2024deepseek} & 80 & $1e17-3e20$ &  &  & Y & Y & Non-embedding \\
\cite{bahri2021explaining} & 8-27 &  & 36.5M & upto 78k steps; 100 epochs & Y & Y & NS \\
\cite{geiping2022much} & 13 &  & ResNet-18 & upto 7.6M images & Y & Y & NS \\
\cite{poli2024mechanistic} & 500 total & 8.00E+19 & 70M-7B &  & Y & Y & Non-embedding  \\
\cite{hu2024minicpm} & 36 &  & 40M-2B & 400M-120B tokens & Y & Y & Non-embedding  \\
\cite{hashimoto2021model} &  &  &  & upto 600k sentences & Y & Y & NA \\
\cite{ruan2024observational} & 27* -77* &  & 70B-180B & 3T-6T tokens & N/A & N/A & N.S. \\
\cite{anil2023palm} & 12 & 1.00E+22 & 15B & 4.00E+11 & N & N & Non-embedding \\
\cite{pearce2024reconciling} & 20 (simulated), 25 (real) &  & 1.5B (simulated), 4.6M (real) & 23B (simulated), 500M (real) tokens & Y & Y & w/ Embedding and Non-embedding considered separately \\
\cite{cherti2023reproducible} & 3* - 29 &  & 214M & 34B (pretrain), 2B (finetune) examples  & Y & Y & N.S \\
\cite{porian2024resolving} & 16 & 2.00E+19 & 901M &  & Y & Y & w/ Embedding and Non-embedding considered separately \\
\cite{alabdulmohsin2022revisiting} & 1* &  & 110M-1B & 1e6-1e10 ex / 3e11 tokens & Mixed & N & N/A \\
\cite{gao2024scalingevaluatingsparseautoencoders} & N.S & N.S & N.S & N.S & N & N & N.S \\
\cite{muennighoff2024scaling} & 142 &  & 8,7B & 900B tokens & Y & Y & w/ embedding \\
\cite{rae2021scaling} & 4 & 6.31E+23 & 280B &  & Y & Y & Non-embedding \\
\cite{shin2023scaling} & 17 & ~0.1 PF Days & 160M & 500M-50B tokens & Y & Y & NA \\
\cite{hernandez2022scaling} & 56 &  & 1.5M-800M & 100B tokens & N & N & NS \\
\cite{filipovich2022scaling} & 4 &  & 57-509M & 30B token & Y & N & NS \\
\cite{neumann2022scaling} & 14 &  & ~$5*10^5$ & $10^4$ steps & Y & Y & NS \\
\cite{droppo2021scaling} & 5-21 &  & ~$10^7$ & 134-23k hrs speech & Y & Y & NS \\
\cite{henighan2020scaling} & 6-10 &  & ~$10^11$ & ~$10^12$ tokens & Y & Y & Non-embedding \\
\cite{goyal2024scaling} & 5 &  & CLIP L/14 - ~300M +63M & 32-640M samples & Y & Y & Embedding \\
\cite{aghajanyan2023scaling} & 21 &  & 8M-6.7B & 5-100B tokens & Y & Y & Non-embedding \\
\cite{kaplan2020scaling} & ~40-150 &  & 1.5B & 23B tokens & Y & Y & Non-embedding \\
\cite{ghorbani2021scaling} & 12-14 &  & 191-3B & NS & Y & Y & Non-embedding \\
\cite{gao2023scaling} & 9 &  & 3B & 120-90k & N & Y & NS \\
\cite{hilton2023scaling} & NS & $10^{20}$ &  &  & Y & Y & NS \\
\cite{frantar2023scaling} & 48 and 112 &  & 0.66M-85M  & 1.8B images, 65B tokens & Y & Y & Non-embedding \\
\cite{prato2021scaling} & 5 &  &  & $10^6$ samples & Y & N & NA \\
\cite{covert2024scaling} & 10 &  & NA & 1000 samples for IMDB & Y & Y & NA \\
\cite{hernandez2021scaling} & NS & $10^{21}$ & $10^8$ &  & Y & N & Non-embedding \\
\cite{ivgi2022scaling} & 5-8 &  & $10^4-10^8$ & varies; 500k steps PT & Y & Y & Non-embedding \\
\cite{tay2022scaling} &  &  & 16-30B & $2^19$ & Y & Y & NA \\
\cite{tao2024scaling} & 60 &  & 33M-1.13B NV + 4-96k V & 4.3B-509B Characters & Y & Y & Embedding and Non-embedding considered separately \\
\cite{jones2021scaling} & 200 & 1E+12-1E+17 &  & 4E+08-2E+09 & Y & Y & NA \\
\cite{zhai2022scaling} & 44 &  & 5.4M-1.8B & 1-13M images & Y & Y & NA \\
\cite{dettmers2023case} & 4 &  & 19M-176B & NA & NA & Y & NA \\
\cite{dubey2024llama} & NS & $6*10^{18}-10^22$ & 40M-16B &  & Y* & Y & NS \\
\cite{hoffmann2022training} & ~200-450 & $6*10^{18}-3*10^{21}$ & 16B & 5B-400B tokens & Y & Y & Non-embedding \\
\cite{ardalani2022understanding} & NS & $10^2$-$10^6$ TFlops &  & ~5M-5B samples & N & N & All are considered \\
\cite{clark2022unified} & 56 &  & 15M-1.3B & 130B tokens & Y & Y & Non-embedding \\
\bottomrule
\end{tabular}

}

\caption{Details on training setup for each paper surveyed.}
\label{tab:full-setup}
\end{table}

\begin{table}[]
\centering
\resizebox{\textwidth}{!}{%

\begin{tabular}{lllll}
\toprule
Paper & Data Points Per Law? & Scaling Law Metric & Modification Of Final Metric? & Subsets Of Data Used \\
\midrule
\cite{rosenfeld2019constructive} & 42-49 & Loss / Top1 Error & N & N \\
\cite{mikamiscaling} & 7 & Error Rate & N & N \\
\cite{schaeffer2023emergent} & NA & Various downstream & NA & NA \\
\cite{sardana2023beyond} & NS & Loss & NS & NS \\
\cite{sorscher2022beyond} & ~60 & Error Rate & NA & NA \\
\cite{caballero2022broken} & 3-40 & FID, Loss, Error Rate, Elo Score & N & NS \\
\cite{besiroglu2024chinchilla} & 245 & Loss & N & N \\
\cite{gordon2021data} & 45-55 & Loss & N & N \\
\cite{bansal2022data} & NS & Loss, BLEU & NS & NS \\
\cite{hestness2017deep} & NS & Token Error, CER, Error Rate, Loss & Median min. validation error across multiple training runs with separate random seeds & NS \\
\cite{bi2024deepseek} & upto 80 & Validation bits-per-byte & NS & NS \\
\cite{bahri2021explaining} & upto 100 & Loss & NS & NS \\
\cite{geiping2022much} & ~50 & Effective Extra Samples & Interpolation & NS \\
\cite{poli2024mechanistic} & NS & Loss & NS & NS \\
\cite{hu2024minicpm} & NS & Loss & NS & NS \\
\cite{hashimoto2021model} & NS & Loss & NS & NS \\
\cite{ruan2024observational} &  & Various downstream & N & N \\
\cite{anil2023palm} & 12 & Loss & N & N \\
\cite{pearce2024reconciling} & 20, 5 & Loss & N & N \\
\cite{cherti2023reproducible} & 3-29 & Error Rate & N & N \\
\cite{porian2024resolving} & 12 & Loss & N & N \\
\cite{alabdulmohsin2022revisiting} & N.S. & Loss / Accuracy & N & N/A \\
\cite{gao2024scalingevaluatingsparseautoencoders} & N.S & MSE & N.S & N.S \\
\cite{muennighoff2024scaling} & 142 & Loss & N & Outliers removed \\
\cite{rae2021scaling} & 4 & Loss & N/A & N/A \\
\cite{shin2023scaling} & NA & Loss & NA & NA \\
\cite{hernandez2022scaling} & NS & Loss & N & N \\
\cite{filipovich2022scaling} & NS & Loss & N & N \\
\cite{neumann2022scaling} & 238 & Elo Score & N & N \\
\cite{droppo2021scaling} & NS & Loss & N & N \\
\cite{henighan2020scaling} & NS & Loss, Error Rate & NS & Drop smaller models \\
\cite{goyal2024scaling} & NS & Error Rate & N & N \\
\cite{aghajanyan2023scaling} & NS & Perplexity & N & N \\
\cite{kaplan2020scaling} & NS & Loss & NS & NS \\
\cite{ghorbani2021scaling} & NS & Loss, BLEU & Median of last 50k steps &  \\
\cite{gao2023scaling} & ~90 & RM Score & NS & NS \\
\cite{hilton2023scaling} & NS & Intrinsic Performance & Smoothing learning curve & Exclude early checkpoints \\
\cite{frantar2023scaling} & 48 and 112 & Loss & NS & NS \\
\cite{prato2021scaling} & 5 & Error Rate & NS & NS \\
\cite{covert2024scaling} & (1000-5000 )*10 & Expectation  & NS & N \\
\cite{hernandez2021scaling} & 40-120 & Loss & NS & NS \\
\cite{ivgi2022scaling} & 5-8 & Loss & N & [2.5, 97.5] percentile \\
\cite{tay2022scaling} & NA & Loss, Accuracy & NA & NA \\
\cite{tao2024scaling} & 20*60 & Loss & Interpolation & NS \\
\cite{jones2021scaling} & 2800 & Elo Score & NS & NS \\
\cite{zhai2022scaling} & NS & Accuracy & NS & NS \\
\cite{dettmers2023case} & NA & Accuracy & NA & NA \\
\cite{dubey2024llama} & ~150 & Loss, Accuracy & NS & NS \\
\cite{hoffmann2022training} & upto 1500 & Loss & N & Lowest loss model of a FLOP count, last 15\% of checkpoints \\
\cite{ardalani2022understanding} & ~130 & Loss & NS & NS \\
\cite{clark2022unified} & ~26*56 & Loss & Log & NS \\
\bottomrule
\end{tabular}

}

\caption{Details on data extraction for each paper surveyed.}
\label{tab:full-eval}
\end{table}

\begin{table}[]
\centering
\resizebox{\textwidth}{!}{%
\begin{tabular}{llllll}
\toprule
Paper & Curve-Fitting Method & Loss Objective & Hyperparameters Reported? & Initialization & Are Scaling Laws Validated? \\
\midrule
\cite{rosenfeld2019constructive} & Least Squares Regression & Custom error term & N/A & Random & Y \\
\cite{mikamiscaling} & Non-linear Least Squares in log-log space &  & N/A & N/A & Y \\
\cite{schaeffer2023emergent} & NA & NA & NA & NA & NA \\
\cite{sardana2023beyond} & L-BFGS & Huber Loss & Y & Grid Search & N \\
\cite{sorscher2022beyond} & NA & NA & NA & NA & NA \\
\cite{caballero2022broken} & Least Squares Regression & MSLE & N/A & Grid Search, optimize one & Y \\
\cite{besiroglu2024chinchilla} & L-BFGS & Huber Loss & Y & Grid Search & Y \\
\cite{gordon2021data} & Least Squares Regression &  & N/A & N.S. & N \\
\cite{bansal2022data} & NS & NS & N & NS & N \\
\cite{hestness2017deep} & NS & RMSE & N & NS & Y \\
\cite{bi2024deepseek} & NS & NS & N & NS & Y \\
\cite{bahri2021explaining} & NS & NS & N & NS & N \\
\cite{geiping2022much} & Non-linear Least Squares &  & NA & Non-augmented parameters & Y \\
\cite{poli2024mechanistic} & NS & NS & N & NS & N \\
\cite{hu2024minicpm} & scipy curvefit & NS & N & NS & N \\
\cite{hashimoto2021model} & Adagrad & Custom Loss & Y & Xavier & Y \\
\cite{ruan2024observational} & Linear Least Squares & Various & N/A & N/A & Y \\
\cite{anil2023palm} & Polynomial Regression (Quadratic) & N.S. & N & N.S. & Y \\
\cite{pearce2024reconciling} & Polynomial Least Squares & MSE on Log-loss & N/A & N/A & N \\
\cite{cherti2023reproducible} & Linear Least Squares & MSE & N/A & N/A & N \\
\cite{porian2024resolving} & Weighted Linear Regression & weighted SE on Log-loss & N/A & N/A & Y \\
\cite{alabdulmohsin2022revisiting} & Least Squares Regression & MSE & Y & N.S. & Y \\
\cite{gao2024scalingevaluatingsparseautoencoders} & N.S & N.S & N.S & N.S & N.S \\
\cite{muennighoff2024scaling} & L-BFGS & Huber on Log-loss & Y & Grid Search, optimize all & Y \\
\cite{rae2021scaling} & None & None & N/A & N/A & N \\
\cite{shin2023scaling} & NA & NA & NA & NA & NA \\
\cite{hernandez2022scaling} & NS & NS & NS & NS & NS \\
\cite{filipovich2022scaling} & NS & NS & NS & NS & NS \\
\cite{neumann2022scaling} & NS & NS & NS & NS & NS \\
\cite{droppo2021scaling} & NS & NS & NS & NS & NS \\
\cite{henighan2020scaling} & NS & NS & NS & NS & NS \\
\cite{goyal2024scaling} & Grid Search & L2 error & Y & NA & Y \\
\cite{aghajanyan2023scaling} & L-BFGS & Huber on Log-loss & Y & Grid Search, optimize all & Y \\
\cite{kaplan2020scaling} & NS & NS & NS & NS & N \\
\cite{ghorbani2021scaling} & Trust Region Reflective algorithm, Least Squares & Soft-L1 Loss & Y & Fixed & Y \\
\cite{gao2023scaling} & NS & NS & NS & NS & Y \\
\cite{hilton2023scaling} & CMA-ES+Linear Regression & L2 log loss & Y & Fixed & Y \\
\cite{frantar2023scaling} & BFGS & Huber on Log-loss & Y & N Random Trials & Y \\
\cite{prato2021scaling} & NS & NS & NS & NS & NS \\
\cite{covert2024scaling} & Adam & Custom Loss & Y & NS & Y \\
\cite{hernandez2021scaling} & NS & NS & NS & NS & Y \\
\cite{ivgi2022scaling} & Linear Least Squares in Log-Log space & MSE & NA & NS & Y \\
\cite{tay2022scaling} & NA & NA & NA & NA & NA \\
\cite{tao2024scaling} & L-BFGS, Least Squares & Huber on Log-loss & Y & N Random Trials from Grid & Y \\
\cite{jones2021scaling} & L-BFGS & NS & NS & NS & NS \\
\cite{zhai2022scaling} & NS & NS & NS & NS & NS \\
\cite{dettmers2023case} & NA & NA & NA & NA & NA \\
\cite{dubey2024llama} & NS & NS & NS & NS & Y \\
\cite{hoffmann2022training} & L-BFGS & Huber on Log-loss & Y & Grid Search, optimize all & Y \\
\cite{ardalani2022understanding} & NS & NS & NS & NS & NS \\
\cite{clark2022unified} & L-BFGS-B & L2 Loss & Y & Fixed & NS \\
\bottomrule
\end{tabular}

}

\caption{Details on optimization for each paper surveyed.}
\label{tab:full-opt}
\end{table}

\newpage

\section{Recommendations}\label{sec:app_recs}

As seen in our analyses, many decisions in our checklist have a number of reasonable options, but those reasonable choices lead to a wide range of scaling law fits, and the observed variations do not follow any clear pattern. It is probable that variations would be even harder to predict when varying model architectures or other design decisions, removing the possibility of a universal set of best practices.
However, it is certainly possible to determine that some scaling law fits are plausible or highly implausible, and to observe the stability of the fitting procedure. 
With the caveat that following any recommendations can not guarantee good scaling law fit, we can make some more concrete recommendations based on these observations:
\paragraph{Scaling Law Hypothesis}
\begin{itemize}
    \item Fitting fewer scaling law parameters at a time typically results in greater stability. In some cases, it may be beneficial to decompose the scaling law fitting problem into two separate procedures. Examples of this approach are the IsoFLOP procedure from \citet{hoffmann2022training}, as well as fitting first the relation between $L$ and $C$, then finding the optimal $N$ and $D$ for a $C$, as seen in \citet{porian2024resolving}.
\end{itemize}
\paragraph{Training Setup}
\begin{itemize}
    \item The trained models should include a wide range of input variables settings. For example, when the input variables to the scaling law are $L$, $N$, $D$, the included models should include a wide range of $N$ and $D$ values for each $C$, or equivalently, should include a wide range of $D/N$ ratios. If the included settings do not include the true optimum, the procedure will struggle to fit to the optimum.
    \item Sweeping for the optimal learning rate results in a less stable fit than fixing the learning rate. We hypothesize that this may be because the true optimal learning rate for each model and data budget size is not any of the options we consider, and thus, each model varies in the difference between its true and approximate optimal learning rate. This may introduce additional noise to the data. Due to resource constraints, we are unable to fully test this hypothesis, and it may not hold at significantly larger scale, but we recommend fixing the learning rate, or changing it according to, say, model size, according to a fixed formula.
\end{itemize}
\paragraph{Data Collection}
\begin{itemize}
    \item Results across tasks or datasets should not be mixed. Neither performance predictions nor optimal $D/N$ ratios are fixed across different evaluation settings for the same set of models.
\end{itemize}
\paragraph{Fitting Algorithm}
\begin{itemize}
    \item Scaling law fitting is sensitive to initialization; most known optimization methods for scaling laws are only able, in practice, to the shift parameters near to their initialization values. Thus, a  dense search over initialization values is necessary. If there is a strong hypothesis guiding the choice of one specific initialization, such as a previously fit and validated scaling law, this will also limit the set of possible final scaling law parameter values.
    \item Different losses emphasize the contribution of errors from certain datapoints. The chosen loss should be suited to the distribution of datapoints and sources of noise.
    \item A simple grid search is unlikely to result in a good scaling law fit. Additionally, optimizers designed to fit linear relations may make assumptions about the distribution of errors and should not be used to fit a power law. 
\end{itemize}

\subsection{Example Checklist}

We provide one possible set of responses to our checklist, reflective of some recommendations enumerated above, and loosely based on \citet{hoffmann2022training}. These answers roughly correspond to a subset of the experiments we run in \S\ref{sec:own-repl}.

\newpage

\fbox{\begin{minipage}[!ht]{38em}

\subsubsection*{(Mis)Fitting: Scaling Law Reproducibility Checklist}

\small

\begin{minipage}[!htp]{0.95\textwidth}
\raggedright
\paragraph{Scaling Law Hypothesis (\S\ref{sec:power-law-form})}

\begin{itemize}[leftmargin=*]
    \item What is the form of the power law? \textit{$L(N, D) = E + \frac{A}{N^\alpha} + \frac{B}{D^\beta} $}
    \item What are the variables related by (included in) the power law? \textit{$N$: the number of model parameters, $D$: the number of data tokens, and $L$: the model's validation loss}
    \item What are the parameters to fit? \textit{$A$, $B$, $E$, $\alpha$, $\beta$}
    \item On what principles is this form derived? \textit{This is taken from \citet{hoffmann2022training}, who hypothesize this form on the basis of prior work in risk decomposition.}
    \item Does this form make assumptions about how the variables are related? \textit{This form inherently assumes that $N$ and $D$ do not have any interaction in their effect on the scaling of $L$. For some experiments, we use the assumption $\alpha = \beta$ to simplify optimization.}
\end{itemize}

\paragraph{Training Setup (\S\ref{sec:model_training})}
\begin{itemize}[leftmargin=*]
    \item How many models are trained?
    \textit{Refer to Table 10}
    \item At which sizes?
    \textit{Refer to Table 10}
    \item On how much data each? On what data? Is any data repeated within the training for a model?
    \textit{Refer to Table 10.}
    \item How are model size, dataset size, and compute budget size counted? For example, how are parameters of the model counted? Are any parameters excluded (e.g., embedding layers)? How is compute cost calculated?  \textit{We include the results including and excluding embedding layers for both the total parameter count $N$ and the total FLOP count $C$. We also include, for comparison, results using the estimate $C=6ND$.}
    \item Are code/code snippets provided for calculating these variables if applicable?
    \url{https://github.com/hadasah/scaling_laws}
    \item How are hyperparameters chosen (e.g., optimizer, learning rate schedule, batch size)? Do they change with scale? \textit{Most hyperparameters are chosen based on best practices in current literature; several are taken directly from the settings in \citet{hoffmann2022training}. For learning rate, we conduct an extensive hyperparameter search across 2-3 orders of magnitude, multiplying by 2-2.5, and then conduct training at 3 learning rates, including the optimum, for nearly all ($N$, $D$) configurations.}
    \item What other settings must be decided (e.g., model width vs. depth)? Do they change with scale?  
    \textit{Refer to Table 10}
    \item Is the training code open source?
    \textit{Yes}
\end{itemize}

\begin{table}[H]
    \centering
    \resizebox{\columnwidth}{!}{%
\begin{tabular}{l|c|c|c|c}
\toprule
Model size & \# layers & Hidden Dim & \# Attn Heads & \# Steps \\ \midrule
12M & 5	    & 448	& 7	 & \{100, 200, 250, 360, 500, 750, \\
& & & &  1,000, 4,000\} \\ \midrule
17M & 7	    & 448	& 7	& \{100, 200, 250, 500, 750, 1,000,\\
& & & &   1,250, 1,500\} \\ \midrule
25M & 8	& 512	& 8		& \{250, 360, 500, 750, 1,000, 1,500, \\
& & & &  2,000, 8,000, 16,000\} \\ \midrule
35M & 9	    & 576	& 9		& \{200, 250, 360, 500, 750, 1,000, \\
& & & &  1,250, 1,500, 4,000, 16,000\} \\ \midrule
50M & 10	& 640	& 10 	& \{250, 500, 750, 1,000, 1,250, 1,500, \\
& & & &  1,800, 2,000, 2,500, 4,000, 16,000\} \\ \midrule
70M & 12	& 704	& 11 	& \{250, 500, 750, 1,000, 1,250, 1,500, \\
& & & &  2,000, 2,500, 3,000, 5,000\} \\ \midrule
100M & 14	& 768	& 12 	& \{250, 500, 1,000, 1,500, 2,000, \\
& & & &  2,500, 3,000, 4,000, 6,000, 12,000\} \\ \midrule
200M & 18	& 960	& 15 	& \{500, 1,500, 6,000, 9,000\} \\ \midrule
300M & 19	& 1152	& 18 	& \{4,000\} \\ \midrule
400M & 20	& 1280	& 20	& \{3,000, 6,000\} \\ \midrule
1B & 26	    & 1792	& 28	& \{20,000\} \\ \bottomrule
\end{tabular}%
}
    \caption{\small{Architecture and Data budget details for models trained in \S\ref{sec:own-repl}}}
\label{tab:arch} 
\end{table}

\raggedright

\end{minipage}

\end{minipage}}

\fbox{\begin{minipage}[!ht]{38em}

\small

\begin{minipage}[!ht]{0.9\textwidth}
\raggedright

\paragraph{Data Collection(\S\ref{sec:data})}
\begin{itemize}[leftmargin=*]
    \item Are the model checkpoints provided openly?
    \textit{Yes, at  \url{https://github.com/hadasah/scaling_laws}}
    \item How many checkpoints per model are evaluated to fit each scaling law? Which ones, if so?  \textit{Unless clearly denoted otherwise, one checkpoint per model is evaluated; the last checkpoint. By default, no mid-training checkpoints are used, i.e., from before the termination of the cosine learning rate schedules.}
    \item What evaluation metric is used? On what dataset?  \textit{We use cross-entropy loss, measured on a held-out validation subset of the Common Crawl \citep{raffel2020exploring} dataset.}
    \item Are the raw evaluation metrics modified? Some examples include loss interpolation, centering around a mean or scaling logarithmically.  \textit{No.}
    \item If the above is done, is code for modifying the metric provided? 
    \textit{Yes.}
\end{itemize}

\paragraph{Fitting Algorithm (\S\ref{sec:opt})}
\begin{itemize}[leftmargin=*]
    \item What objective (loss) is used? \textit{We try various loss objectives (1) log-Huber loss, (2) MSE, (3) MAE and (4) Huber loss}
    \item What algorithm is used to fit the equation? \textit{Mainly L-BFGS, but we also experiment with BFGS, non-linear least squares and grid search.}
    \item What hyperparameters are used for this algorithm? \textit{Thresholds of $\{1e-4, 1e-6, 1e-8\}$, exact gradient }
    \item How is this algorithm initialized? \textit{We initialize with 4500 initializations similar to \citet{hoffmann2022training}.}
    \item Are all datapoints collected used to fit the equations? For example, are any outliers dropped? Are portions of the datapoints used to fit different equations? \textit{No outliers are dropped in general, but we do show some results on specific subsets of models. For example, we compare the result of a scaling law fit when using only models trained at a peak learning rate of $1e-4$ or $4e-4$.}
    \item How is the correctness of the scaling law considered? Extrapolation, Confidence Intervals, Goodness of Fit? \textit{Currently, we do not evaluate the correctness beyond comparing to results in literature \citep{hoffmann2022training,kaplan2020scaling}.}
\end{itemize}

\end{minipage}


\end{minipage}}

\newpage

\section{Full Analysis Plots}\label{sec:app_megafigure}

\begin{figure}[!htp]
\centering
\begin{subfigure}{\textwidth}
    \centering
\begin{subfigure}{0.49\textwidth}
    \includegraphics[width=\textwidth]{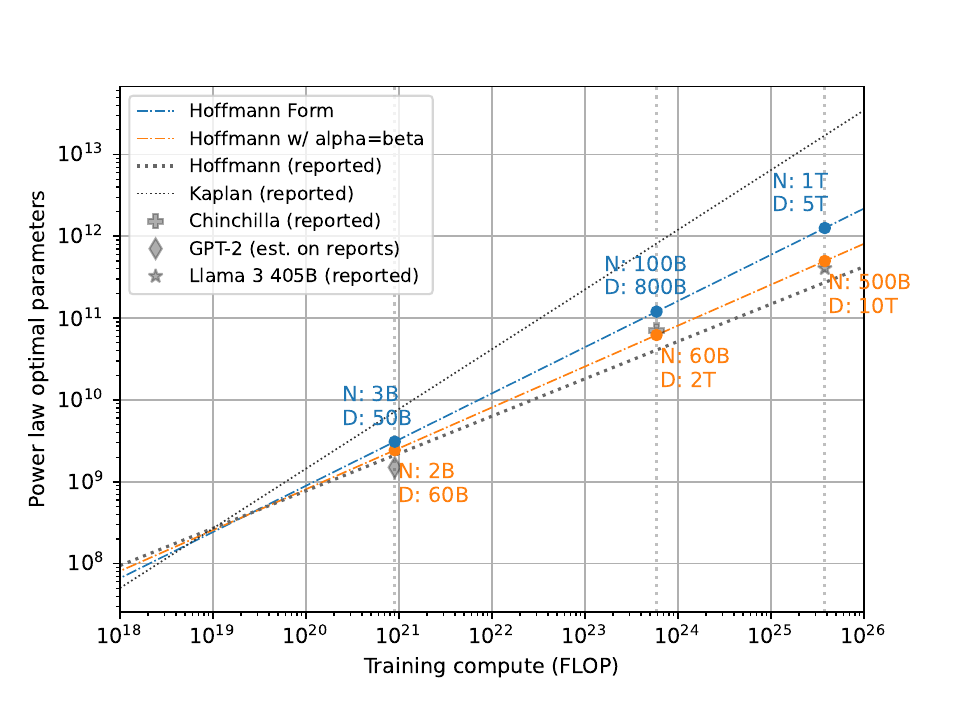}  \footnotesize{\citet{hoffmann2022training,besiroglu2024chinchilla}}
\end{subfigure}
\\ \vspace{1em}
\centering
\begin{subfigure}{0.49\textwidth}
    \centering
    \includegraphics[width=\textwidth]{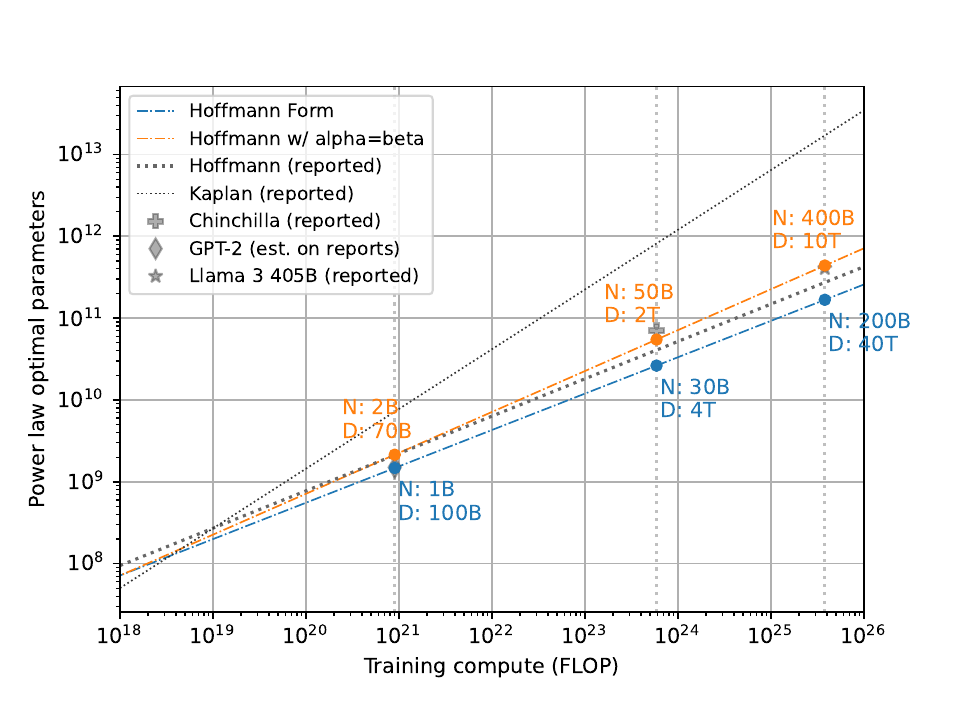}
    \footnotesize{\citet{porian2024resolving}}
\end{subfigure}
\hfill
\begin{subfigure}{0.49\textwidth}
    \centering
    \includegraphics[width=\textwidth]{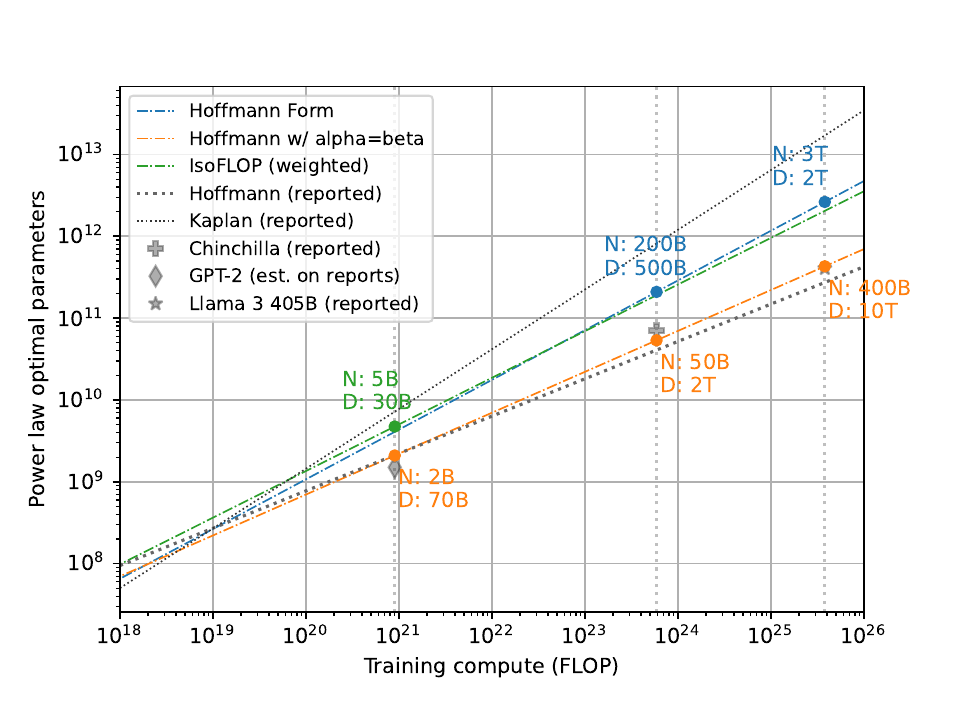}
    \footnotesize{\citet{porian2024resolving} (all checkpoints)}
\end{subfigure}
\\ \vspace{1em}
\centering
\begin{subfigure}{0.49\textwidth}
    \centering
    \includegraphics[width=\textwidth]{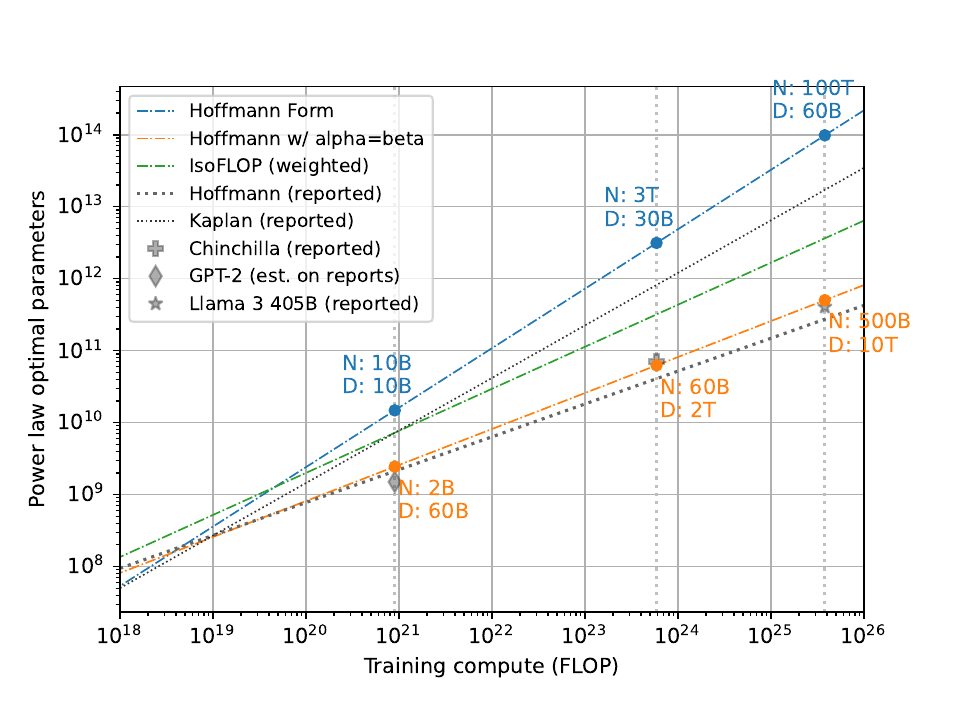}
    \footnotesize{Ours}
\end{subfigure}
    \hfill
\begin{subfigure}{0.49\textwidth}
    \centering
    \includegraphics[width=\textwidth]{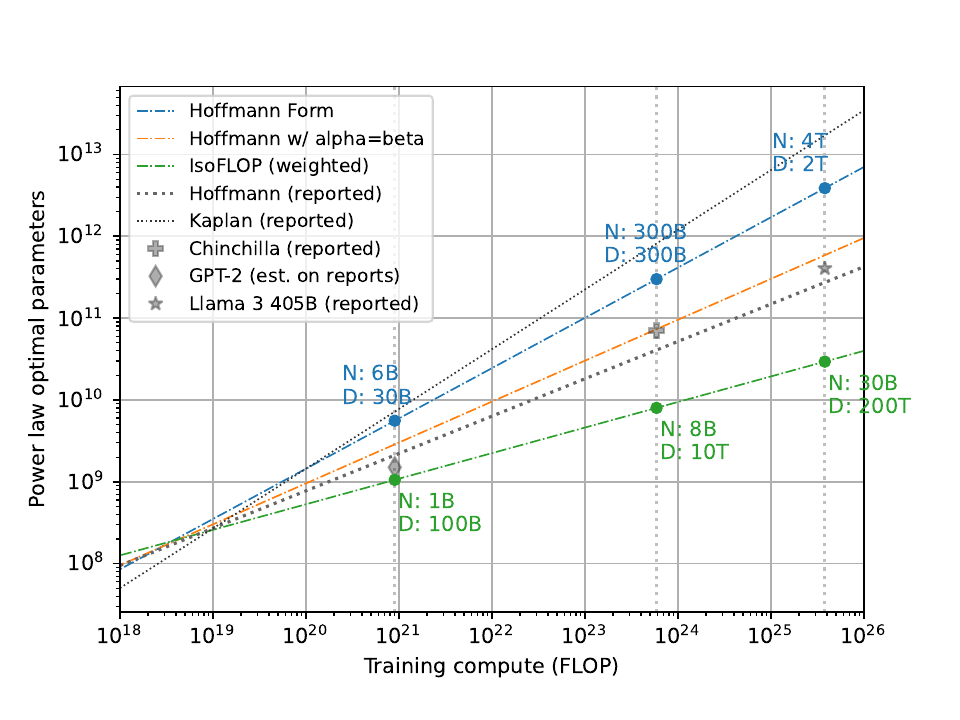}
    \footnotesize{Ours (all checkpoints)}
\end{subfigure}
\\ \vspace{1em}
\caption{\textbf{\S\ref{sec:power-law-form}, \S\ref{sec:repl-power-law-form}} Using data from both \citet{besiroglu2024chinchilla} (left) and our own models (right), we compare the effects of fitting to the power law form used in Approach 3 of \citet{hoffmann2022training} with the variant used by \citet{muennighoff2024scaling}, which assumes that the exponents $\alpha, \beta$ are equal -- equivalently, that $N^*(C)$ and $D^*(C)$ scale about linearly with each other.  When using only the performance of final checkpoints from both \citet{besiroglu2024chinchilla} and our own experiments, taking this assertion results in a law much closer to the one reported by \citet{hoffmann2022training}. On our own models, we also show results when using the IsoFLOP approach from \citet{hoffmann2022training}. As we are using only results from final model checkpoints, the size of the data input to the IsoFLOP approach in this case is reduced.}
\label{fig:analysis_form_app}
\end{subfigure}
\end{figure}

\begin{figure}[]
\ContinuedFloat 
    \centering
\begin{subfigure}{\textwidth}
\begin{subfigure}{0.49\textwidth}
    \centering
    \includegraphics[width=\textwidth]{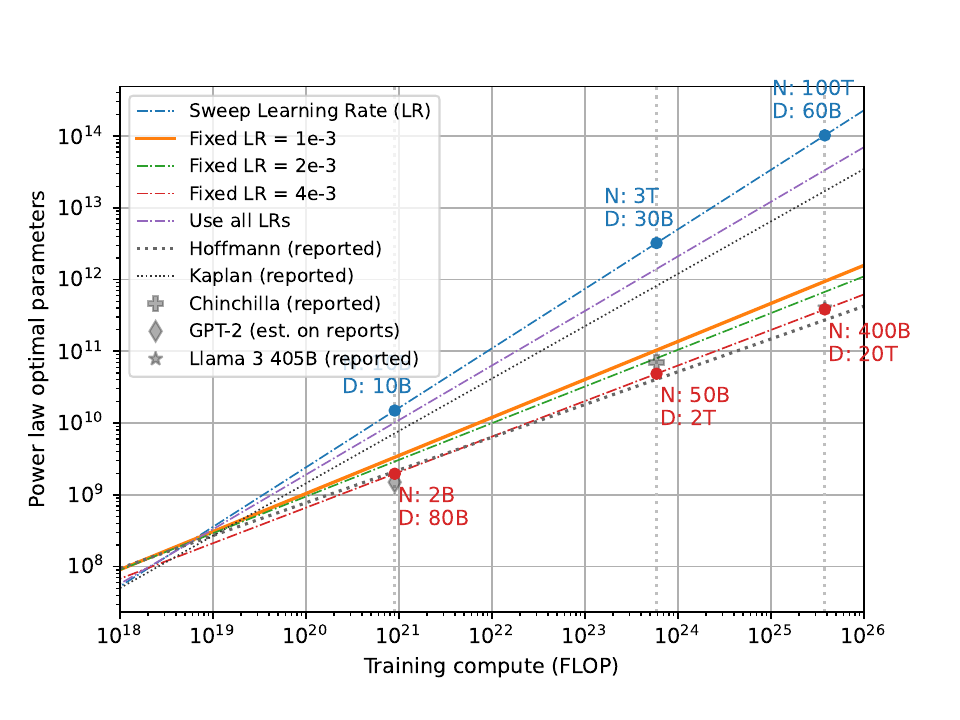}
    \footnotesize{Ours}
\end{subfigure}
    \hfill
\begin{subfigure}{0.49\textwidth}
    \centering
    \includegraphics[width=\textwidth]{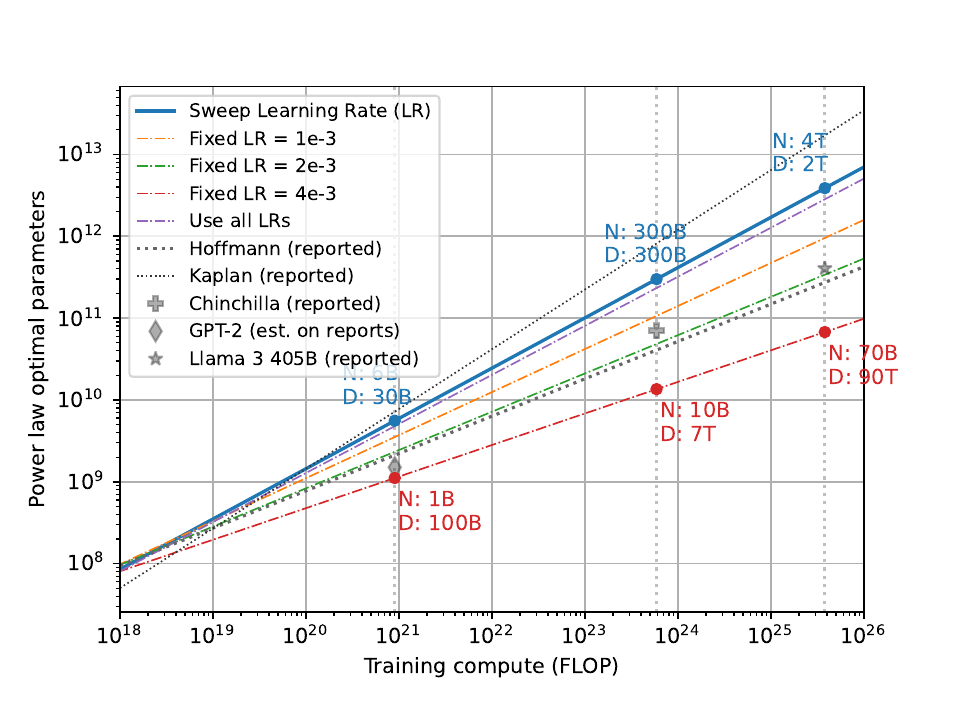}
    \footnotesize{Ours (all checkpoints)}
\end{subfigure}
    \caption{\textbf{\S\ref{sec:model_training}, \S\ref{sec:repl-model_training}} With our models, we simulate the effects of not sweeping the learning rate. As a baseline, (1) we sweep at each ($N$, $D$) pair for the optimal learning rate over a range of values at most a multiple of 2 apart. Next, (2) we use a learning rate of 1e-3 for all $N$, the optimal for our 1 billion parameter models, and do the same for (3) 2e-3 and (4) 4e-3, which is optimal for our 12 million parameter models. Lastly, we use all models across all learning rates at the same $N$ and $D$. Results vary dramatically across these settings. Somewhat surprisingly, using all learning rates results in a very similar power law to sweeping the learning rate, whereas using a fixed learning rate of 1e-3 or 4e-3 yields the lowest optimization loss or closest match to the \citet{hoffmann2022training} power laws, respectively.}
\end{subfigure}
\end{figure}

\begin{figure}[]
\ContinuedFloat
\centering 
\begin{subfigure}{\textwidth}
    \centering
\begin{subfigure}{0.49\textwidth}
    \centering
    \includegraphics[width=\textwidth]{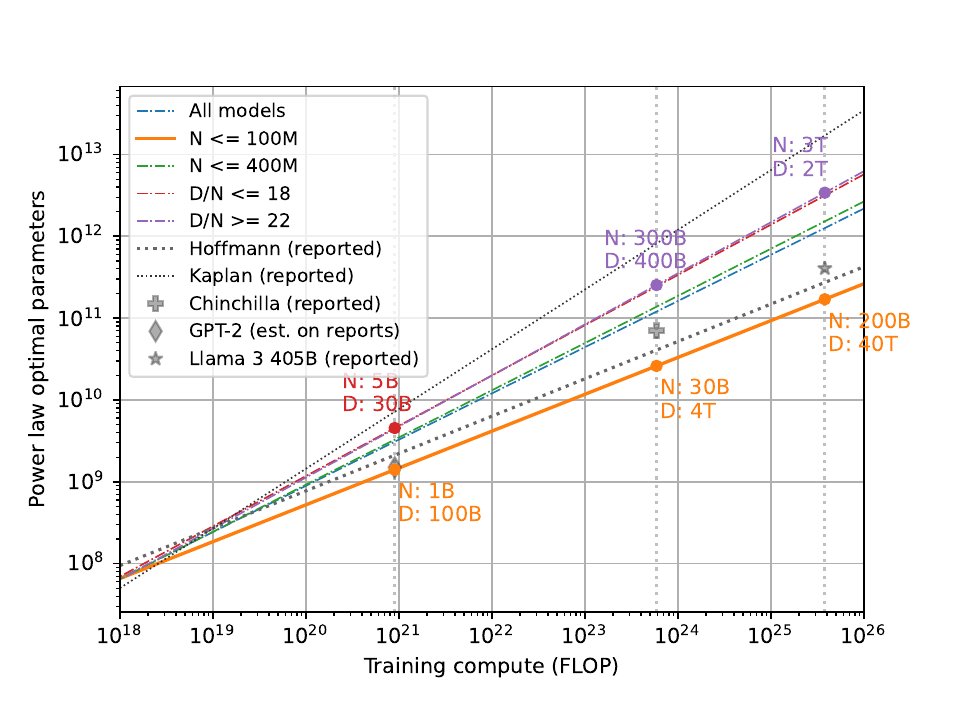}
\footnotesize{\citet{hoffmann2022training,besiroglu2024chinchilla}}
\end{subfigure}
\\ \vspace{1em}
\begin{subfigure}{0.49\textwidth}
    \centering
    \includegraphics[width=\textwidth]{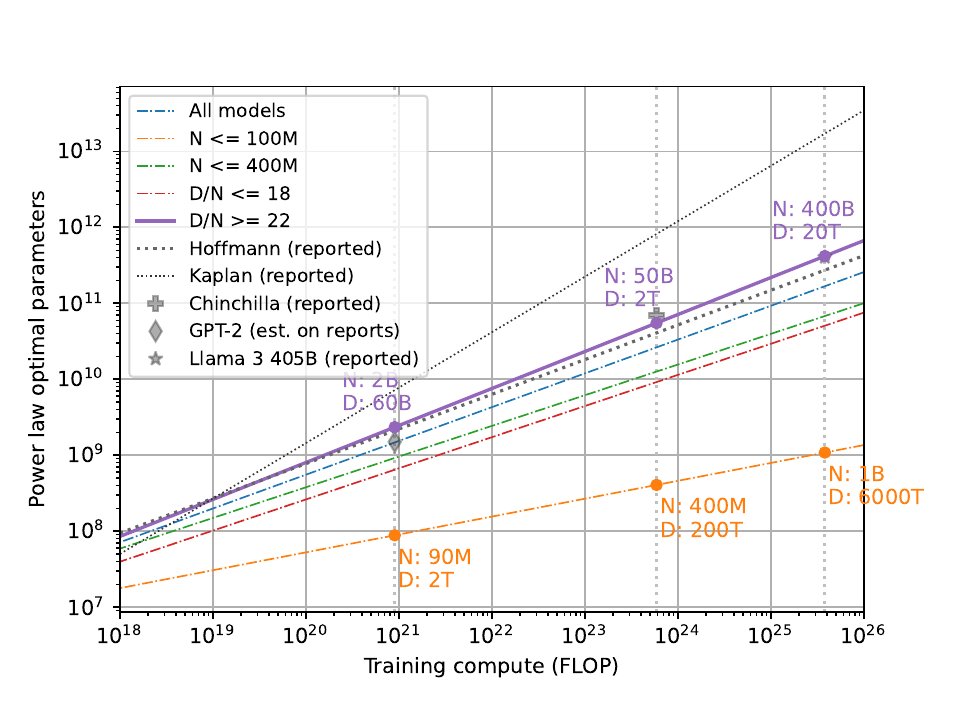}
    \footnotesize{\citet{porian2024resolving}}
\end{subfigure}
\hfill
\begin{subfigure}{0.49\textwidth}
    \centering
    \includegraphics[width=\textwidth]{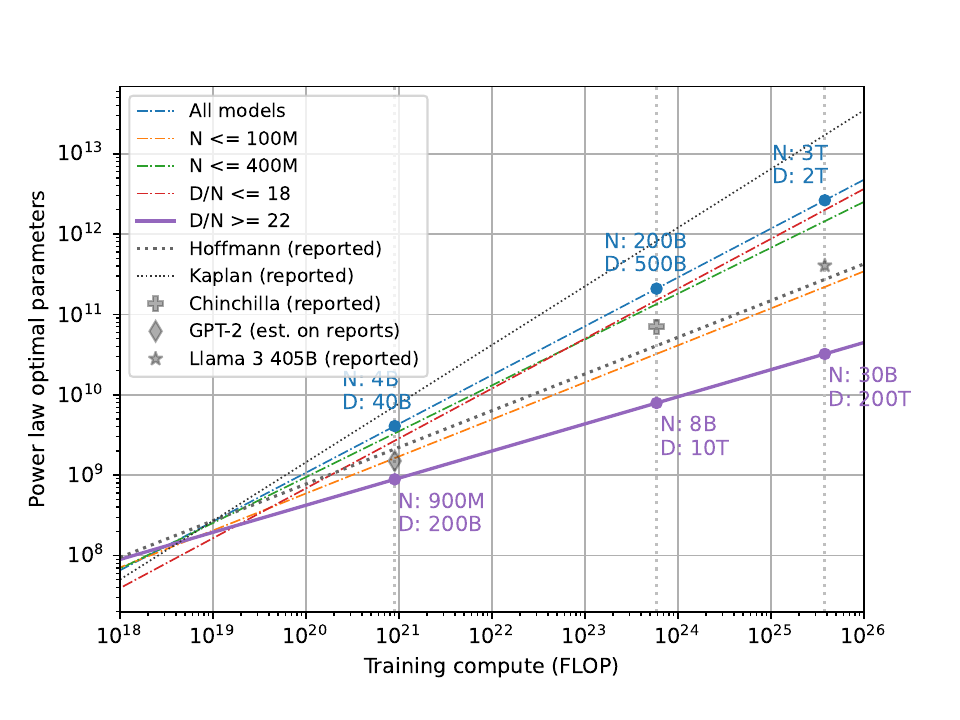}
    \footnotesize{\citet{porian2024resolving} (all checkpoints)}
\end{subfigure}
\\ \vspace{1em}
    \centering
\begin{subfigure}{0.49\textwidth}
    \centering
    \includegraphics[width=\textwidth]{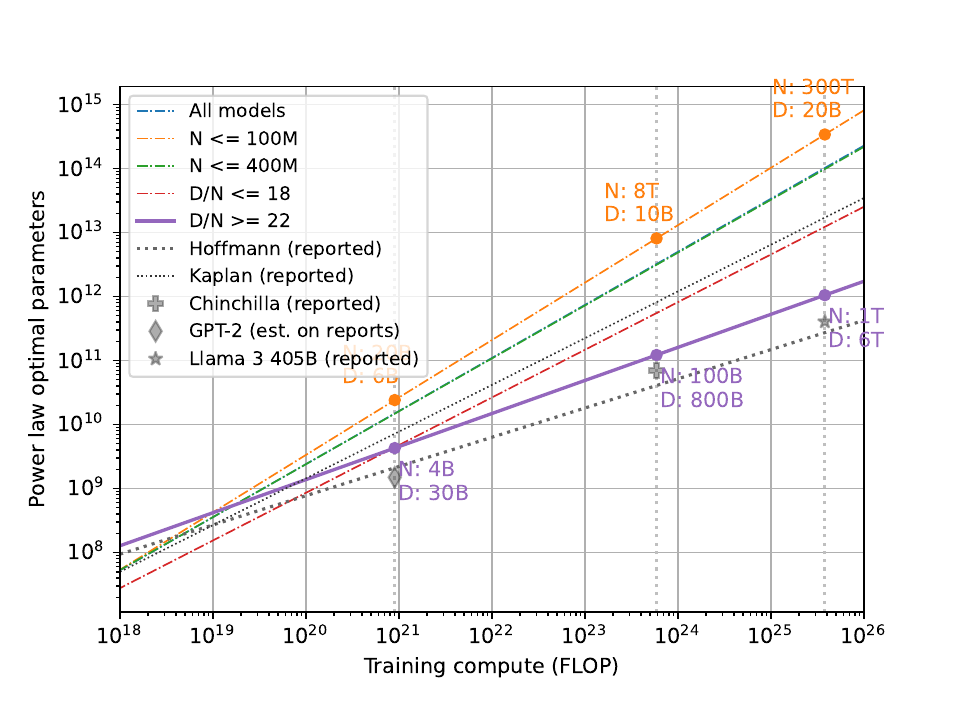}
    \footnotesize{Ours}
\end{subfigure}
\hfill 
\begin{subfigure}{0.49\textwidth}
    \centering
    \includegraphics[width=\textwidth]{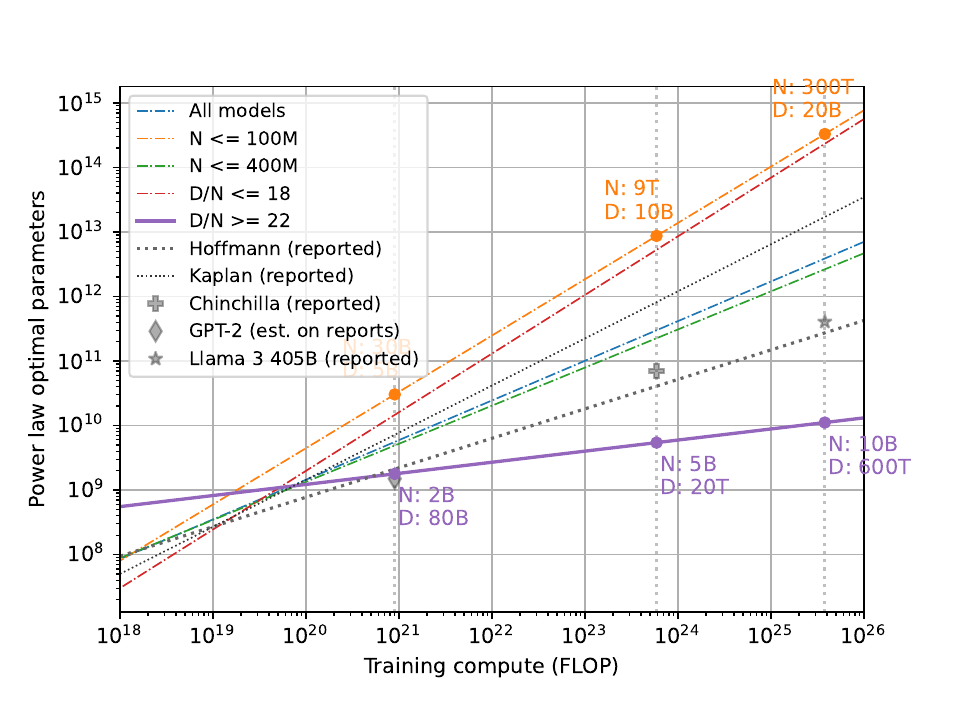}
    \footnotesize{Ours (all checkpoints)}
\end{subfigure}
\\ \vspace{1em}
\caption{\textbf{\S\ref{sec:model_training}, \S\ref{sec:repl-model_training}} From all 3 datasets, we choose subsets with ($N$, $D$) values which fit a particular method one might have of setting up training. We fit with (1) all models, which for our dataset, ranges from 12 million to 1 billion parameters, then with (2) only models of up to about 100 million parameters or (3) up to 400 million parameters. We also compare the effects of a higher or lower hypothesis about the optimal $D/N$ ratio, including (4) only models with $D/N \leq 18$ or (5) $D/N \geq 22$. These ranges are designed to exclude $D/N = 20$, the rule of thumb based on \citet{hoffmann2022training}. The minimum or maximum $D/N$ ratio tested does skew results; above $10^{22}$ FLOPs, (4) and (5) fit to optimal ratios $D/N < 18$ and $D/N > 22$, respectively. Removing our largest models in (2) also creates a major shift in the predicted optimal $D/N$.}
    \label{fig:analysis_nd_app}
\end{subfigure}
\end{figure}

\begin{figure}[]
\ContinuedFloat
\centering 
\begin{subfigure}{\textwidth}
    \centering
\begin{subfigure}{0.49\textwidth}
    \centering
    \includegraphics[width=\textwidth]{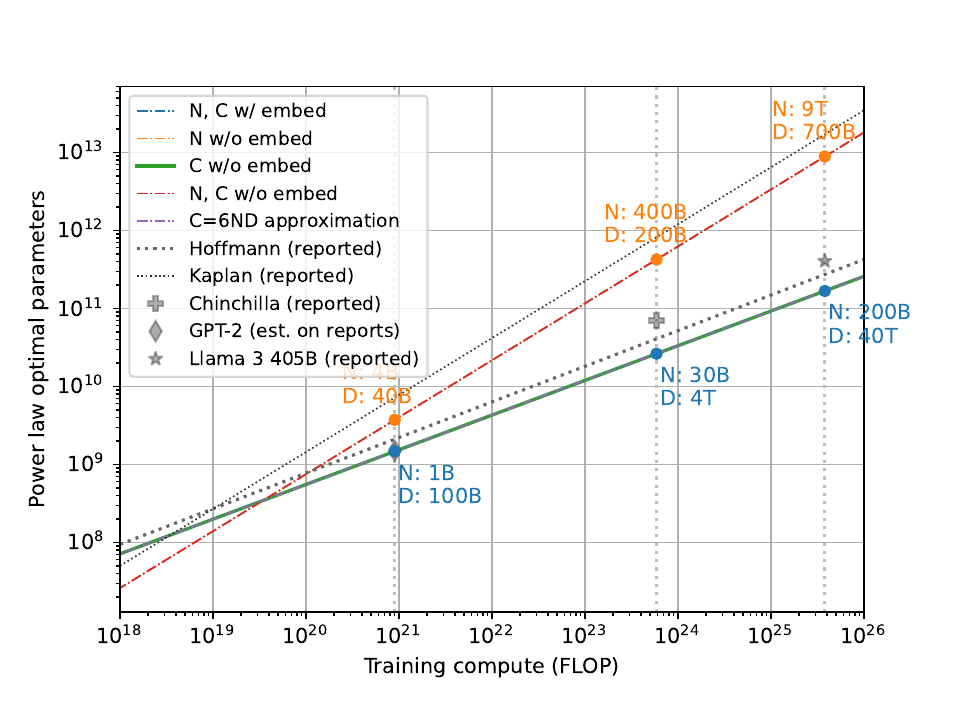}
    \footnotesize{\citet{porian2024resolving}}
\end{subfigure}
\hfill
\begin{subfigure}{0.49\textwidth}
    \centering
    \includegraphics[width=\textwidth]{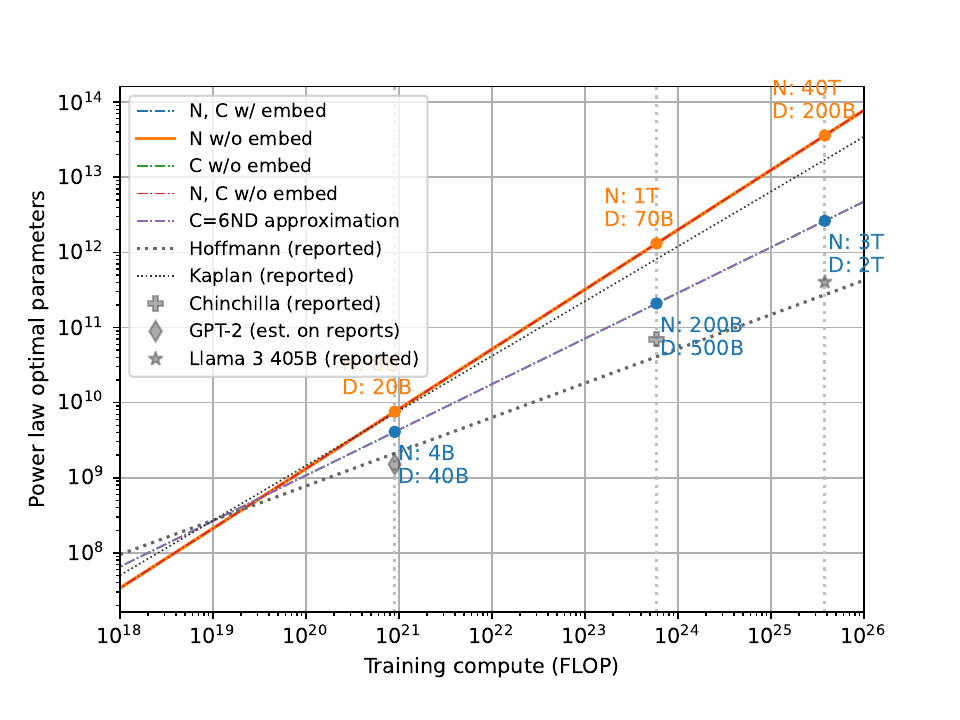}
    \footnotesize{\citet{porian2024resolving} (all checkpoints)}
\end{subfigure}
\\ \vspace{1em}
\begin{subfigure}{0.49\textwidth}
    \centering
    \includegraphics[width=\textwidth]{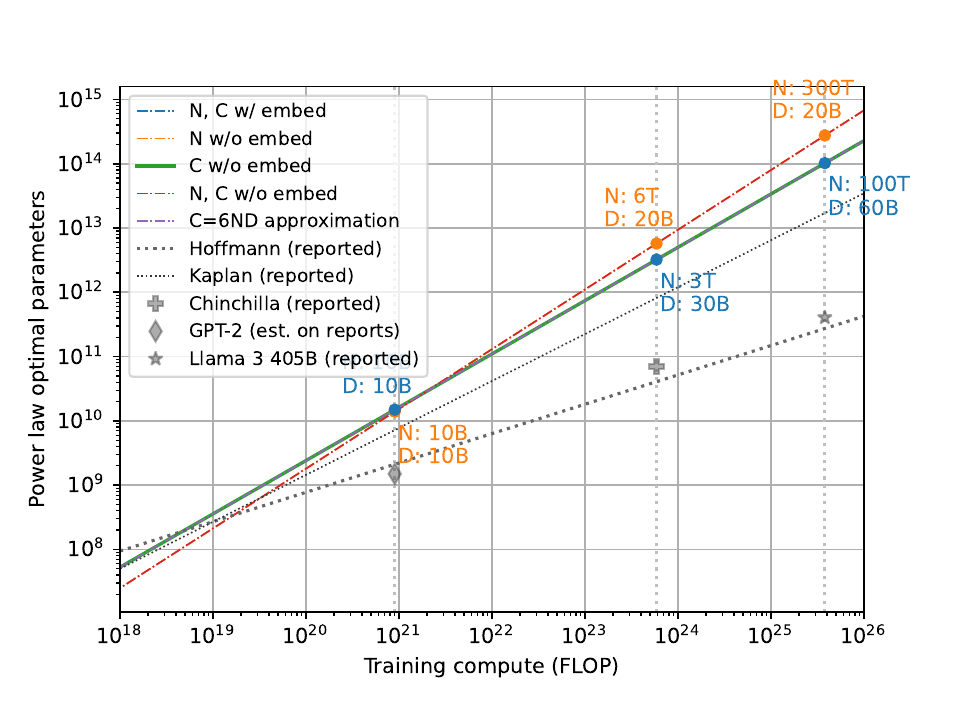}
    \footnotesize{Ours}
\end{subfigure}
\hfill
\begin{subfigure}{0.49\textwidth}
    \centering
    \includegraphics[width=\textwidth]{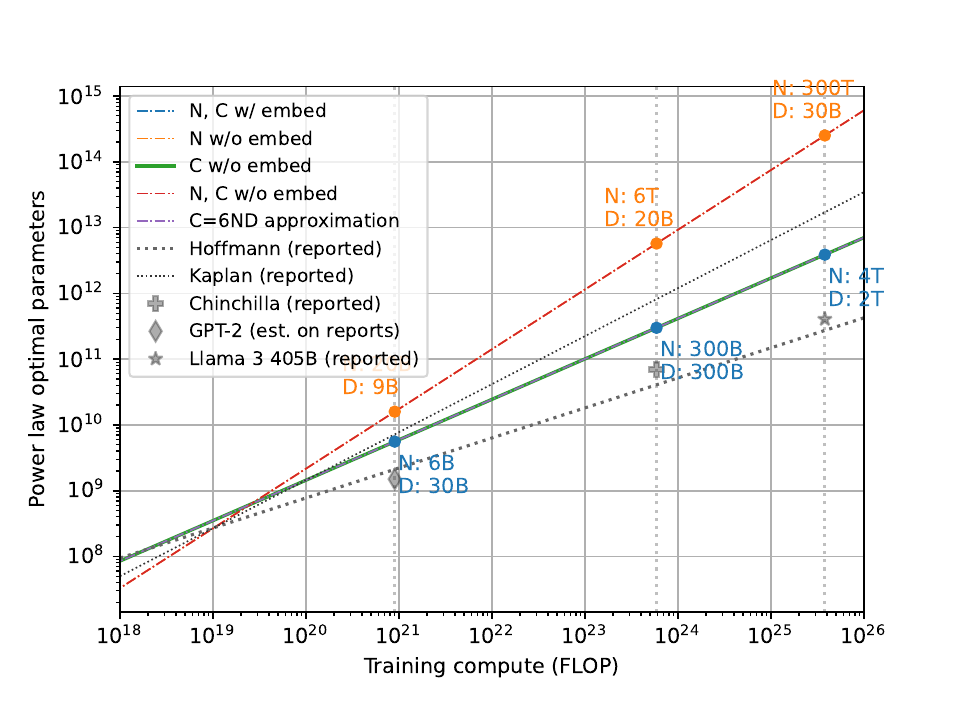}
    \footnotesize{Ours (all checkpoints)}
\end{subfigure}
\\ \vspace{1em}
\caption{\textbf{\S\ref{sec:model_training}, \S\ref{sec:repl-model_training}} With our own data and those from \citet{porian2024resolving}, we fit power laws to the same sets of models, while varying the ways we count $N$ and $C$. We compare (1) including embeddings, which is our baseline, with (2) excluding embeddings only in $N$, (3) excluding embeddings only in $C$, (4) excluding embeddings in both $N$ and $C$. We also compare to using the $C=6ND$ approximation, including embedding parameters. Throughout this work, we calculate the FLOPs in a manner similar to \citet{hoffmann2022training}, and we open source the code for these calculations.
    With both datasets, the exclusion of embeddings in FLOPs has very little impact on the final fit. Similarly, using the $C=6ND$ approximation has no visible impact. For the \citet{porian2024resolving} models, the exclusion of embedding parameters in the calculation of $N$ results in scaling laws which differ substantially, and with increasing divergences at large scales. 
    }
    \label{fig:analysis_counting_app}
\end{subfigure}
\end{figure}

\begin{figure}[]
\ContinuedFloat
\centering 
\begin{subfigure}{\textwidth}
\centering
\begin{subfigure}{0.49\textwidth}
    \centering
    \includegraphics[width=\textwidth]{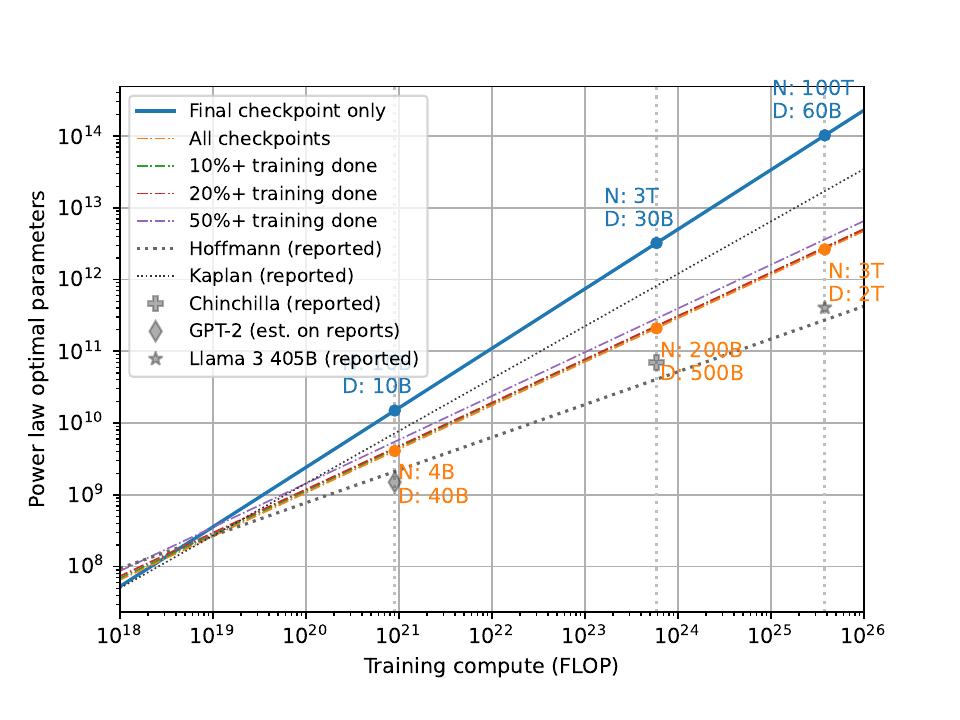}
    \footnotesize{\citet{porian2024resolving} (all checkpoints)}
\end{subfigure}
\\ \vspace{1em}
\begin{subfigure}{0.49\textwidth}
    \centering
    \includegraphics[width=\textwidth]{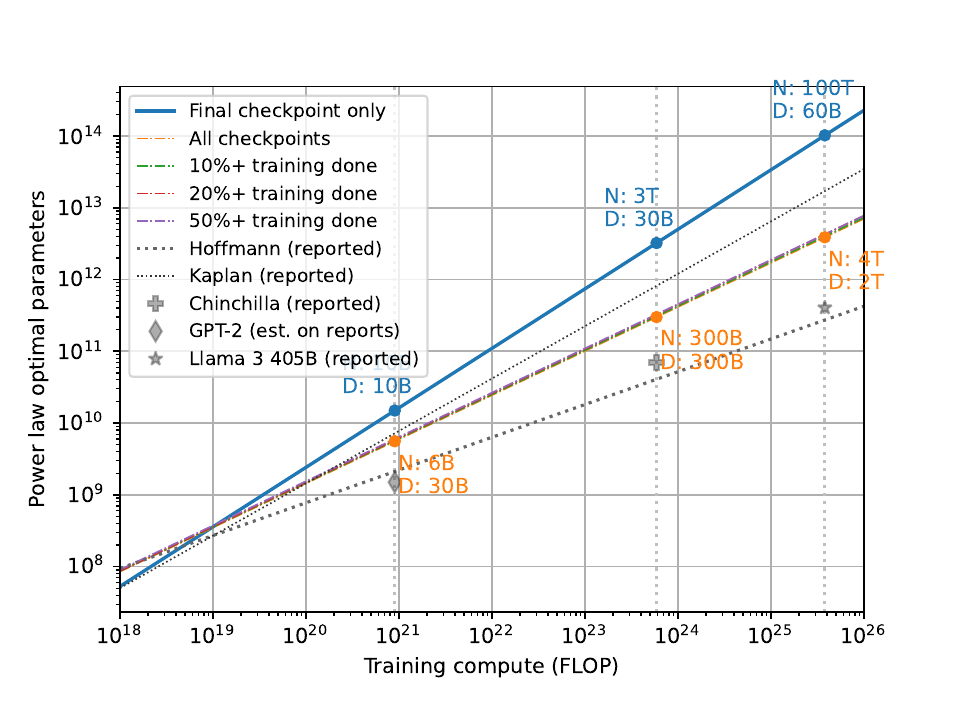}
    \footnotesize{Ours (all checkpoints)}
\end{subfigure}
\\ \vspace{1em}
\caption{\textbf{\S\ref{sec:data}, \S\ref{sec:repl-data}} With models from \citet{porian2024resolving} and our own dataset, we compare (1) fitting a power law using only the final checkpoint with (2) using all mid-training checkpoints (3) using all checkpoints, starting 10\% through training, (4) the same, starting 20\% through training, and (5) the same again, starting 50\% through training. We observe that (2)-(5) consistently
yields power laws more similar to that reported by \citet{hoffmann2022training}, so we also repeat all analyses in Figures~\ref{fig:analysis_form}-\ref{fig:analysis_counting_ours}. Using mid-training checkpoints sometimes results in more stable fits which are similar to the \citet{hoffmann2022training} scaling laws, but the effect is noisy and dependent on other decisions
}
\label{fig:analysis_checkpoint_app}        
\end{subfigure}
\end{figure}

\begin{figure}[]
\ContinuedFloat
\centering 
\begin{subfigure}{\textwidth}
\begin{subfigure}{0.49\textwidth}
    \centering
    \includegraphics[width=\textwidth]{images/analysis_init_epoch_ai_C_vs_N.pdf}
    \footnotesize{\citet{hoffmann2022training,besiroglu2024chinchilla}}
\end{subfigure}
\\ \vspace{1em}
\begin{subfigure}{0.49\textwidth}
    \centering
    \includegraphics[width=\textwidth]{images/analysis_init_rsld_final_C_vs_N.pdf}
    \footnotesize{\citet{porian2024resolving}}
\end{subfigure}
\hfill
\begin{subfigure}{0.49\textwidth}
    \centering
    \includegraphics[width=\textwidth]{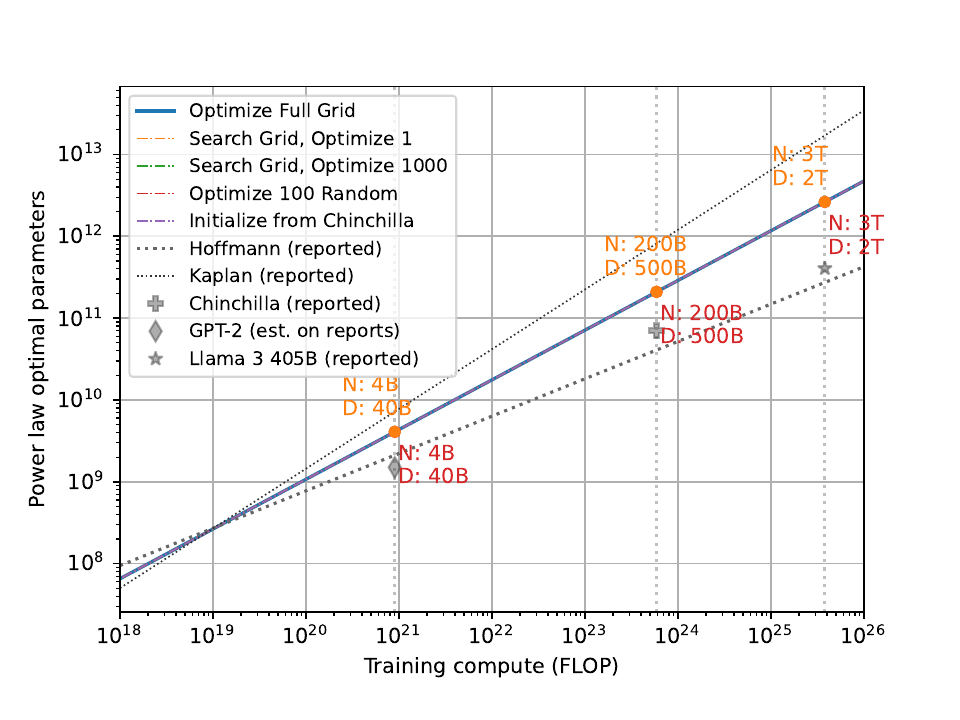}
    \footnotesize{\citet{porian2024resolving} (all checkpoints)}
\end{subfigure}
\\ \vspace{1em}
    \centering
\begin{subfigure}{0.49\textwidth}
    \centering
    \includegraphics[width=\textwidth]{images/analysis_init_misfitting_new_final_C_vs_N.pdf}
    \footnotesize{Ours}
\end{subfigure}
\hfill
\begin{subfigure}{0.49\textwidth}
    \centering
    \includegraphics[width=\textwidth]{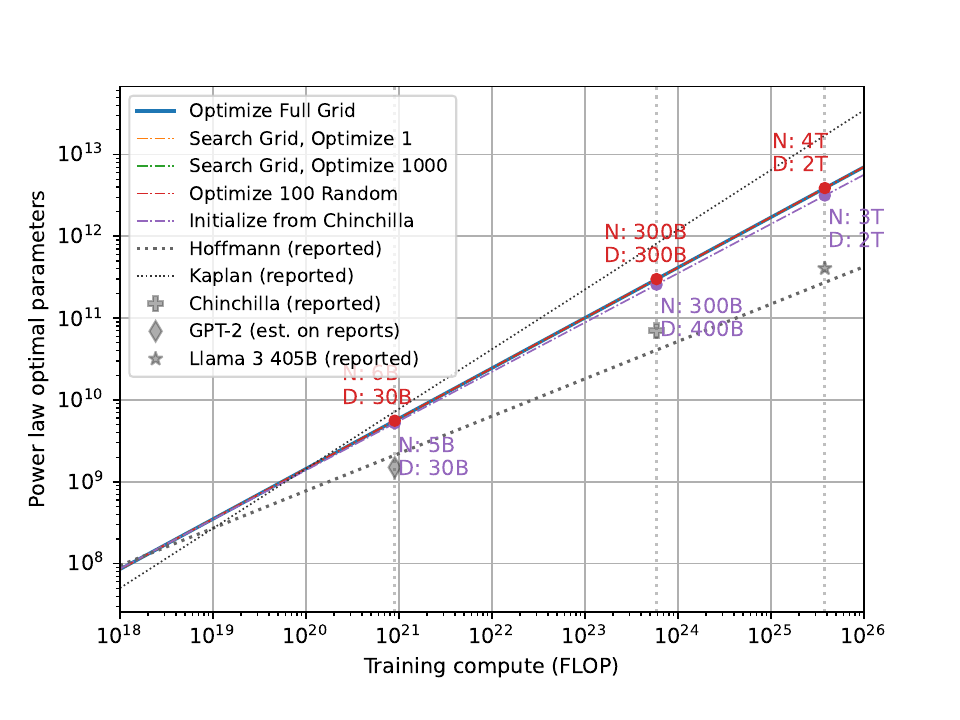}
    \footnotesize{Ours (all checkpoints)}
\end{subfigure}
\\ \vspace{1em}
\caption{\textbf{\S\ref{sec:opt}, \S\ref{sec:repl-opt}} We fit to data from all 3 datasets to experiment with the initialization of parameters in the power law. We start with (1) optimizing every point in a grid search of 6x6x5x5x5=4500 initializations \citep{hoffmann2022training}, (2) randomly sampling from only a single initialization in this grid, (3) searching for the lowest loss initialization point \citep{caballero2022broken}, (4) randomly sampling 100 points, and (5) initializing with the coefficients found in \citet{hoffmann2022training}, as \citet{besiroglu2024chinchilla} does. With the \citet{besiroglu2024chinchilla} data, (5) yields a fit nearly identical to that reported by \citet{hoffmann2022training}, although (1) results in the lowest fitting loss. With the \citet{porian2024resolving} data, all approaches except (2) yield a very similar fit, which gives a recommended $D/N$ ratio similar to that of\citet{hoffmann2022training}. However, using our data, (2) optimizing over only the most optimal initialization yields the best match to the \citet{hoffmann2022training} power laws, followed by (5) initialization from the reported \citet{hoffmann2022training} scaling law parameters. Optimizing over the full grid yields the power law which diverges most from the \citet{hoffmann2022training} law, suggesting the difficulty of optimizing over this space, and the presence of many local minima.
}
\label{fig:analysis_init_app}
\end{subfigure}
\end{figure}

\begin{figure}[]
\ContinuedFloat
\centering 
\begin{subfigure}{\textwidth}
    \centering
\begin{subfigure}{0.49\textwidth}
    \centering
    \includegraphics[width=\textwidth]{images/analysis_loss_epoch_ai_C_vs_N.pdf}
    \footnotesize{\citet{hoffmann2022training,besiroglu2024chinchilla}}
\end{subfigure}
\\ \vspace{1em}
\begin{subfigure}{0.49\textwidth}
    \centering
    \includegraphics[width=\textwidth]{images/analysis_loss_rsld_final_C_vs_N.pdf}
    \footnotesize{\citet{porian2024resolving}}
\end{subfigure}
\hfill
\begin{subfigure}{0.49\textwidth}
    \centering
    \includegraphics[width=\textwidth]{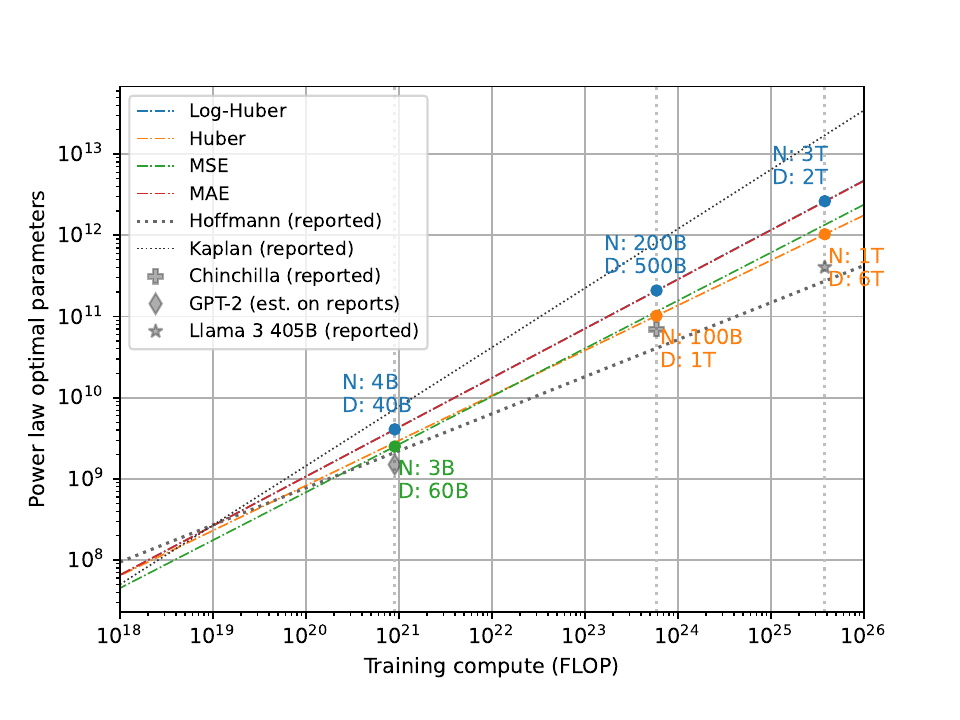}
    \footnotesize{\citet{porian2024resolving} (all checkpoints)}
\end{subfigure}
\\ \vspace{1em}
    \centering
\begin{subfigure}{0.49\textwidth}
    \centering
    \includegraphics[width=\textwidth]{images/analysis_loss_misfitting_new_final_C_vs_N.pdf}
    \footnotesize{Ours}
\end{subfigure}
\hfill
\begin{subfigure}{0.49\textwidth}
    \centering
    \includegraphics[width=\textwidth]{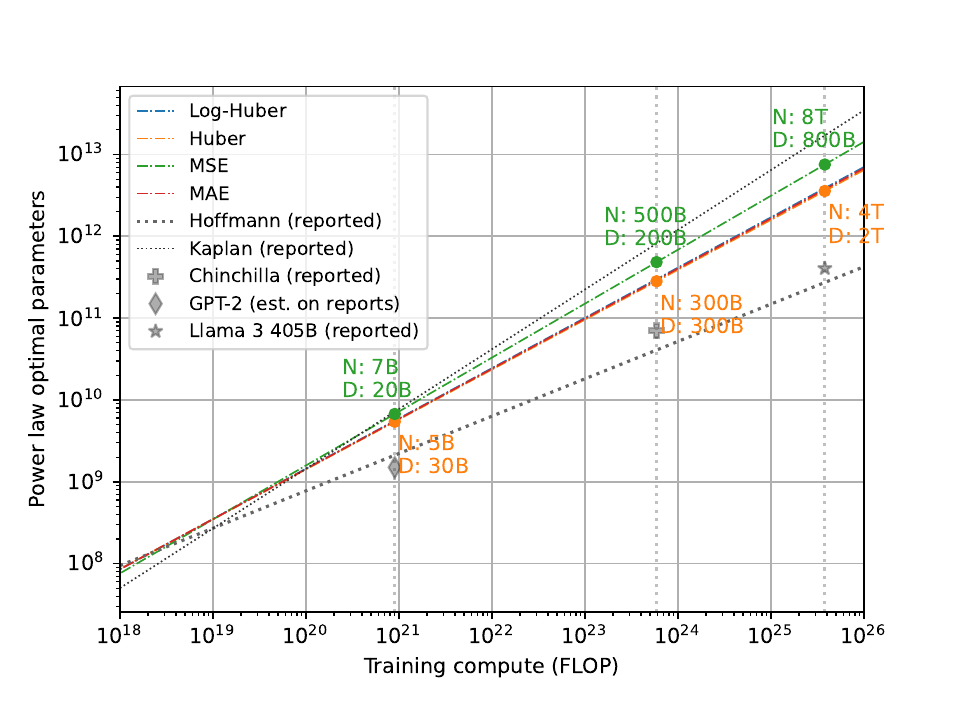}
    \footnotesize{Ours (all checkpoints)}
\end{subfigure}
\\ \vspace{1em}
\caption{\textbf{\S\ref{sec:opt}, \S\ref{sec:repl-opt}} We fit a power law to data from from all 3 datasets, minimizing different objective functions: (1) the baseline log-Huber loss, (2) MSE, (3) MAE, and (4) the Huber loss. 
The resulting power laws are less disparate than when varying many of the other factors discussed above and generally fall near the power law parameters reported by \citet{kaplan2020scaling} and \citet{hoffmann2022training}, but this is still a wide range of recommended optimal parameter counts for each compute budget. Overall, the loss function behavior is not predictable, given the differences between loss functions when looking at the power laws resulting from these three sources of data.}
\label{fig:analysis_loss_app}
\end{subfigure}
\end{figure}

\begin{figure}[]
\ContinuedFloat
\centering 
\begin{subfigure}{\textwidth}
    \centering
\begin{subfigure}{0.49\textwidth}
    \centering
    \includegraphics[width=\textwidth]{images/analysis_opt_epoch_ai_C_vs_N.pdf}
\footnotesize{\citet{hoffmann2022training,besiroglu2024chinchilla}}
\end{subfigure}
\\ \vspace{1em}
\begin{subfigure}{0.49\textwidth}
    \centering
    \includegraphics[width=\textwidth]{images/analysis_opt_rsld_final_C_vs_N.pdf}
    \footnotesize{\citet{porian2024resolving}}
\end{subfigure}
\hfill
\begin{subfigure}{0.49\textwidth}
    \centering
    \includegraphics[width=\textwidth]{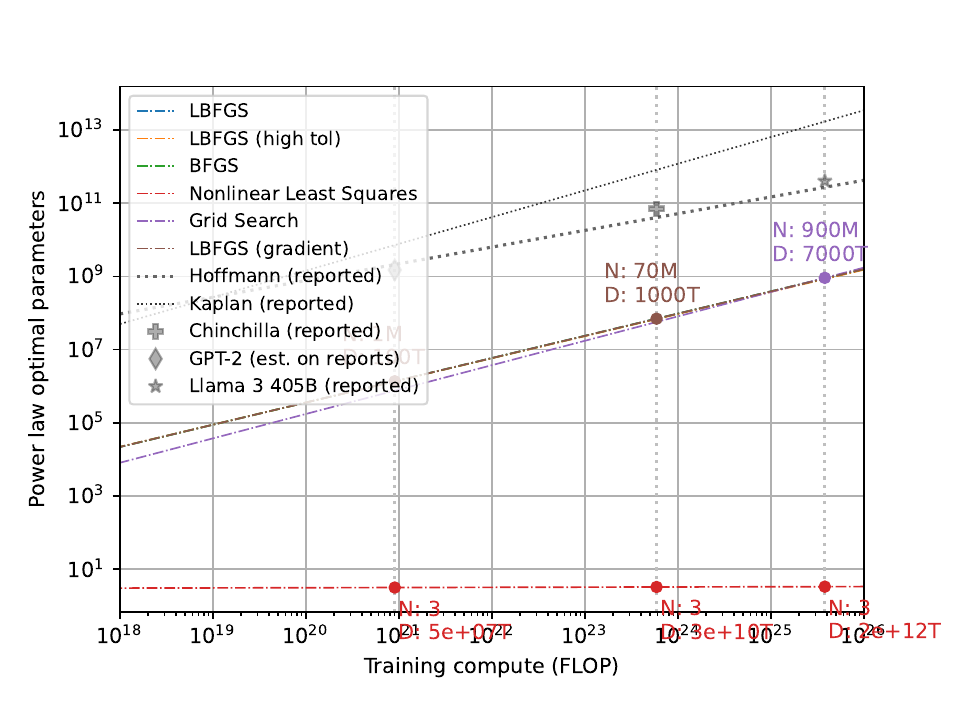}
    \footnotesize{\citet{porian2024resolving} (all checkpoints)}
\end{subfigure}
\\ \vspace{1em}
    \centering
\begin{subfigure}{0.49\textwidth}
    \centering
    \includegraphics[width=\textwidth]{images/analysis_opt_misfitting_new_final_C_vs_N.pdf}
    \footnotesize{Ours}
\end{subfigure}
\hfill
\begin{subfigure}{0.49\textwidth}
    \centering
    \includegraphics[width=\textwidth]{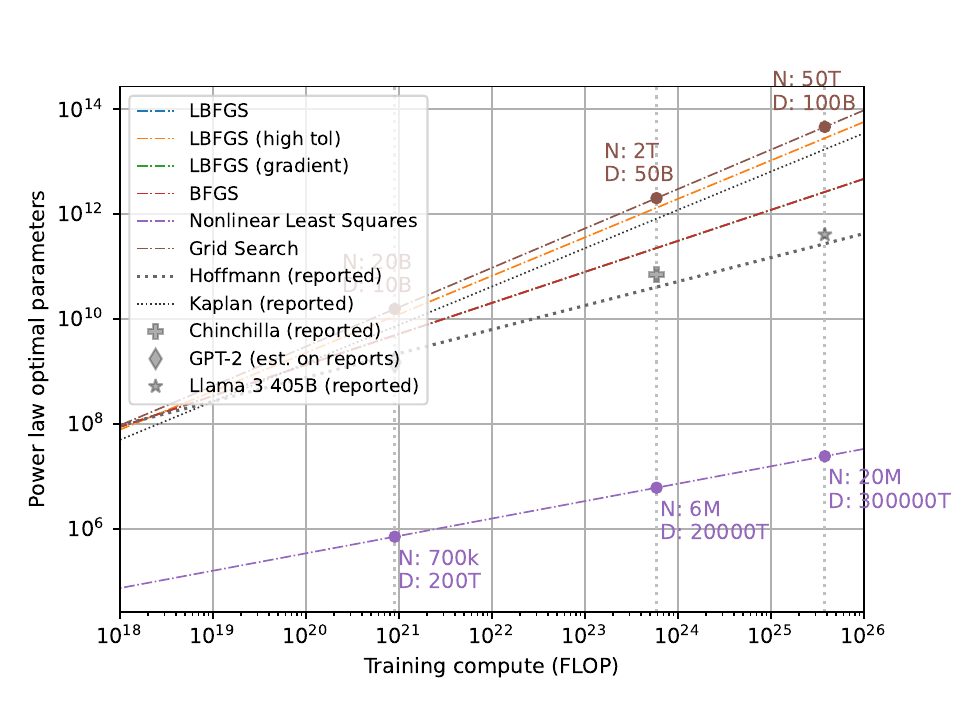}
    \footnotesize{Ours (all checkpoints)}
\end{subfigure}
\caption{\textbf{\S\ref{sec:opt}, \S\ref{sec:repl-opt}} We fit a power law to data from all 3 datasets using various optimizers, beginning with the original (1) L-BFGS. L-BFGS and BFGS implementations have an early stopping mechanism, which conditions on the stability of the solution between optimization steps. We set this threshold for L-BFGS to a (2) higher value 1e-4 (stopping earlier).
We found that using any higher or lower values resulted in the same solutions or instability for all three datasets. L-BFGS and BFGS also have the option to use the true gradient of the loss, instead of an estimate, which is the default. In this figure, we include this setting, (3) using the true gradient for L-BFGS. We compare these L-BFGS settings to (4) BFGS. We test the same tolerance value and gradient settings, and find that none of these options change the outcome of BFGS in our analysis, and omit them from this figure for legibility. Finally, we compare to (5) non-linear least squares and (6) pure grid search, using a grid 5 times more dense along each axis as we used initialization with other optimizers. This density is chosen to approximately match the runtime of L-BFGS.
Many of these optimizers do converge to similar solutions, but this depends on the data, and the settings which diverge from this majority do not follow any immediately evident pattern.} 
\label{fig:analysis_opt_app}
\end{subfigure}
\end{figure}

\begin{figure}[]
\ContinuedFloat
\centering 
\caption{(\textbf{\S\ref{sec:own-repl}}) We replicate the plots in Figure~\ref{fig:overall}, reorganized so that analyses for datasets with mid-training checkpoints appear alongside those for data with final checkpoints only. This side-by-side comparison makes the difference in power law fits apparent, further underscoring the impact of including mid-training datapoints.
We study the effects of various decisions in the fitting of a power law, as outlined in our checklist (Appendix~\ref{sec:app_checklist}) and detailed in \S\ref{sec:power-law-form}-\S\ref{sec:opt}. For each set of analyses, we the scaling laws found by \citep{kaplan2020scaling} and \citep{hoffmann2022training} for comparison. We also include markers indicating 3 existing models for comparison purposes: Llama 3 405B \citep{dubey2024llama}, the Chinchilla model \citep{hoffmann2022training}, and an estimate of the 1.5B GPT-2 model \citep{radford2019language}, for which we know details of the dataset storage size and word count, but not an exact count of data BPE tokens, which we estimate at 100B. We additionally annotate, at the compute budget $C$ for each of these 3 reference points, the maximum and minimum \textit{predicted} (i.e. extrapolated) optimal model parameter count $N_{opt}$ and data budget $D_{opt}$ from the fitted power laws. We use a thicker, solid line for the method in each plot which achieves the lowest optimization loss, with the exception of the plots comparing power law form, those comparing loss functions and those comparing optimizers, for which this would be nonsensical.
We find overall, throughout our analyses, that all of the decisions we explore have an impact on the final fit of the power law, supporting our conclusion that more thorough reporting of these decisions is critical for scaling law reproducibility.
}
\label{fig:app_mega}
\end{figure}

\end{document}